\documentclass{article}

\PassOptionsToPackage{margin=1in}{geometry}
 \usepackage[preprint]{neurips_2026}

\usepackage[utf8]{inputenc} 
\usepackage[T1]{fontenc}    
\usepackage{hyperref}       
\usepackage{url}            
\usepackage{booktabs}       
\usepackage{amsfonts}       
\usepackage{nicefrac}       
\usepackage{microtype}      
\usepackage{xcolor}         
\usepackage{graphicx}
\usepackage{subcaption}
\usepackage{pifont}
\usepackage{wrapfig}
\usepackage{pgf}
\usepackage{multirow}
\usepackage{caption}
\usepackage{algorithm}
\usepackage{algorithmic}
\usepackage{amsmath}
\usepackage{amssymb}
\usepackage{mathtools}
\usepackage{amsthm}
\usepackage{adjustbox}
\usepackage{rotating}
\usepackage{makecell}
\usepackage{longtable}
\usepackage{tikz}
\usetikzlibrary{positioning,arrows.meta,fit,backgrounds,calc,matrix,shapes}
\usepackage{placeins}

\newcommand{\xmark}{\ding{55}}

\newlength{\remainingpageheight}
\newlength{\promptimageheight}

\title{A Foundation Model for Instruction-Conditioned In-Context Time Series Tasks}

\author{%
  Anish Saha \\
  Walmart\\
  Sunnyvale, USA \\
  \texttt{anish.saha@walmart.com} \\
  \And
  Konstantin Shmakov \\
  Walmart\\
  Sunnyvale, USA \\
  \texttt{konstantin.shmakov@walmart.com} \\
}

\begin{document}

\maketitle

\begin{abstract}
In-context learning (ICL) enables task adaptation at inference time by conditioning on demonstrations rather than updating model parameters. 
Although recent time-series foundation models incorporate contextual conditioning, retrieval, or example-based prompting, they typically rely on implicit positional structure or task-specific objectives rather than explicit instruction-conditioned input-output demonstrations. 
We introduce iAmTime, a time-series foundation model trained with instruction-conditioned amortized meta-learning to infer tasks directly from example demonstrations. 
iAmTime represents each episode as a structured prompt over historical context and future-known variables using specialized semantic tokens that attend to designated time-series regions, exchange information across demonstrations, and inject task information into the query representation. 
The model combines a \emph{Hierarchical Multi-Scope Transformer Encoder}, which captures temporal and covariate dynamics while inferring latent task structure from demonstrated input-output mappings, with a \emph{Task-Conditioned Patch Decoder}, which adapts decoding through expert-based routing. 
We train iAmTime on large-scale real and synthetic corpora using supervised and self-supervised instruction-conditioned tasks, including forecasting, imputation, reconstruction, classification, anomaly detection, and source de-mixing. 
Across diverse domains, frequencies, and horizons, iAmTime improves zero-shot adaptation over strong time-series foundation baselines on probabilistic and point forecasting benchmarks, while achieving competitive or superior performance on four non-forecasting tasks.
\end{abstract}

\section{Introduction}
\label{sec:introduction}

Time-series modeling supports decision-making across domains such as retail, energy, finance, transportation, and industrial operations, where applications often require not only forecasting but also anomaly detection \citep{audibert2020usad}, regime classification \citep{fawaz2020deep}, imputation \citep{cao2018brits}, and probabilistic scenario analysis \citep{gneiting2014probabilistic}. The field has progressed from classical models \citep{box1968some} to deep global forecasting methods \citep{salinas2020deepar} and large-scale time-series foundation models \citep{woo2024unified, das2024decoder, ansari2024chronos} trained across diverse domains for zero-shot generalization.


\begin{figure}[t]
    \centering
    \captionsetup[subfigure]{font=small,skip=2pt}

    \begin{subfigure}[t]{0.36\linewidth}
        \vspace{0pt}
        \centering
        \includegraphics[width=\linewidth]
            {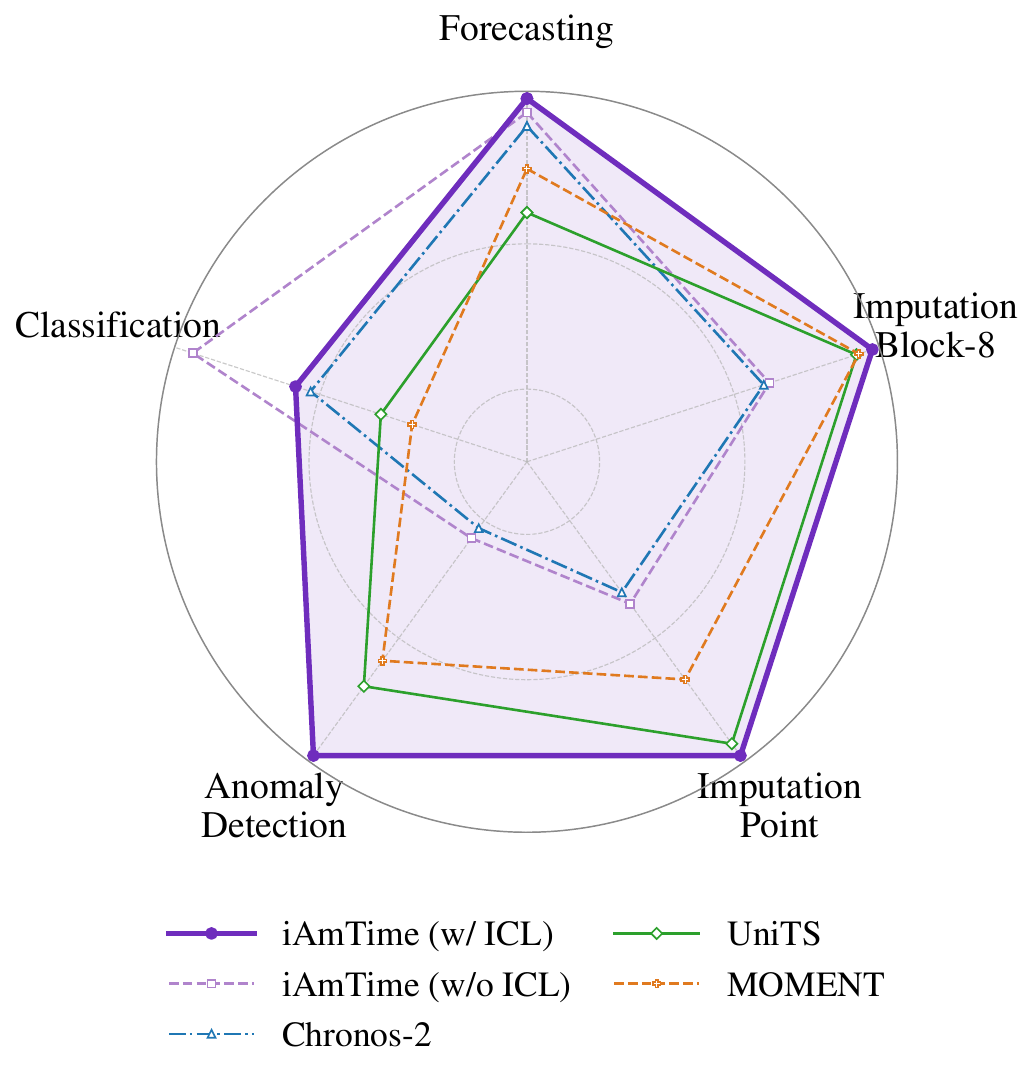}
        \caption{Normalized multi-task performance.}
        \label{fig:task_radar}
    \end{subfigure}
    \hfill
    \begin{subfigure}[t]{0.62\linewidth}
        \vspace{0pt}
        \centering
        \includegraphics[width=\linewidth]
            {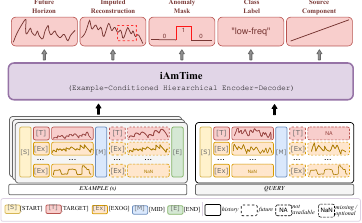}
        \caption{Shared \textsc{iAmTime} architecture.}
        \label{fig:icl_structure}
    \end{subfigure}



    \caption{
        Overview of the instruction-conditioned input, and multi-task performance of
        \textsc{iAmTime}.
    }
    \label{fig:iamtime_overview}
\end{figure}

In parallel, in-context learning (ICL) in Large Language Models enables inference-time adaptation through demonstrations rather than parameter updates \citep{brown2020language}. This capability has been strengthened through few-shot prompting \citep{brown2020language}, instruction tuning \citep{wei2021finetuned}, and chain-of-thought reasoning \citep{wei2022chain}; Transformers trained on input-output episodes can exhibit linear prediction, gradient-descent-like updates, and task selection through attention \citep{garg2022can}, while meta-training on demonstration-query episodes improves ICL ability \citep{min2022metaicl}. Recent time-series models such as CITRAS-FM \citep{yamaguchi2026citras}, TimesFM-2.5 \citep{das2024decoder, das2024context}, Chronos-2 \citep{ansari2025chronos}, TOTO 1.0/2.0 \citep{cohen2025time, khwaja2026toto}, and TiRex 2.0 \citep{podest2026tirex} use long histories, retrieval, or example-based context. However, this typically improves a fixed forecasting objective rather than defining an input-output task mapping at inference time. Multi-task models such as MOMENT \citep{goswami2024moment}, UniTS \citep{gao2024units}, and GPT4TS \citep{zhou2023one} span forecasting and other tasks, but rely on task-specific heads or adaptation rather than a single demonstration-conditioned interface.

We propose iAmTime, a time-series foundation model for instruction-conditioned amortized meta-learning. iAmTime integrates established mechanisms into a unified formulation where input-output demonstrations specify the task at inference time. Examples and queries are represented as structured prompts over targets, covariates, and demonstrated outputs. We use role-typed tokens with masked read/write pathways to summarize permitted regions, exchange mappings across examples, and inject inferred task information into query patches via FiLM-style conditioning; a task-conditioned decoder adapts outputs through expert routing. We train on large-scale real and synthetic corpora across forecasting, imputation, reconstruction, classification, anomaly detection, and source de-mixing, with episode-local remapping and cross-task ambiguity to discourage shortcuts. Although no explicit inner-loop optimization is used as in classical meta-learning \citep{finn2017model}, adaptation is amortized during pretraining and executed in a single example-conditioned forward pass. Across diverse domains and horizons, including fev-bench \citep{shchur2025fev} and GIFT-Eval \citep{aksu2024gift}, this formulation improves zero-shot point and probabilistic forecasting while enabling non-forecasting adaptation through the same interface (see Figure~\ref{fig:iamtime_overview}).
Our contributions are:
\begin{itemize}
    \item We formulate time-series tasks as \emph{demonstration-specified mappings}, enabling inference-time task switching from structured examples without task-specific fine-tuning or heads.
    \item We introduce a leakage-controlled mechanism combining role-typed semantic tokens, region-specific read masks, cross-example mapping retrieval, and FiLM-based task injection to transfer demonstrated mappings to the query.
    \item We develop an instruction-conditioned amortized meta-training regime across forecasting and four other self-supervised temporal tasks, with synthetic multivariate augmentation, episode-local remapping, and cross-task ambiguity to enforce demonstration use.
    \item We show through forecasting, classification, imputation, anomaly detection, mechanism-transfer experiments, and fine-grained ablations that the resulting formulation supports both strong forecasting and inference-time task adaptation within one architecture.
\end{itemize}

\section{Related Work}

\begin{table}[t]
    \centering
    \caption{
    Comparison of modeling capabilities and in-context learning in recent time-series FMs.
    Among these, only \textsc{iAmTime} enables instruction-conditioned in-context learning through
    explicit input-output demonstrations, allowing adaptation to multiple time-series
    tasks at inference time.
    }
    \label{tab:icl_comparison}
    \includegraphics[width=\columnwidth]{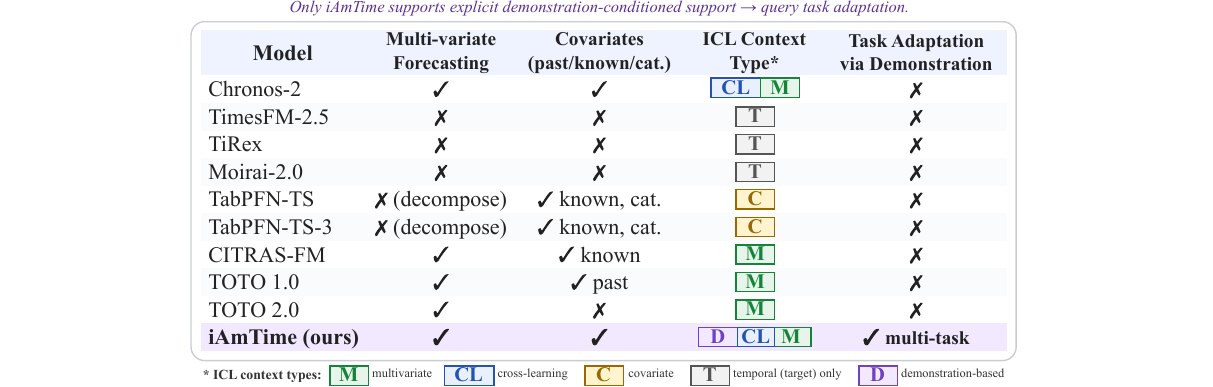}
\end{table}

\textbf{In-context learning and meta-training.}
ICL enables task adaptation from input-output demonstrations without parameter updates \citep{brown2020language}. Its effectiveness depends on prompt structure and calibration and improves with demonstration-query training \citep{zhao2021calibrate, holtzman2021surface}. ICTSP \citep{lu2024context} shows that example-conditioned Transformers improve time-series forecasting through contextual examples. More broadly, meta-learning \citep{vilalta2002perspective, finn2017model} and multi-task learning \citep{evgeniou2004regularized, ruder2017overview} study generalization across task distributions, while diverse-task training improves zero-shot transfer \citep{zhong2021adapting, mishra2022cross, wei2021finetuned}. MetaICL \citep{min2022metaicl} shows that meta-training on demonstration-query episodes improves ICL.
ICTSP \citep{xu2025context} also argues that time-series foundation models must be explicitly pre-trained on multiple tasks to acquire ICL capabilities, however limiting to imputation and backtracing tasks.
We extend this further through instruction-conditioned meta-learning over forecasting and four other self-supervised temporal tasks, and through specialized architectural design and token-based representations coupled with curriculum learning.

\textbf{Time-series foundation and multi-task models.}
Classical methods such as ARIMA \citep{box1968some} and exponential smoothing \citep{hyndman2018forecasting} fit models per series, whereas global models such as DeepState \citep{rangapuram2018deep}, N-BEATS \citep{oreshkin2019n}, N-HITS \citep{challu2023nhits}, TFT \citep{lim2021temporal}, and PatchTST \citep{nie2022time} learn shared representations. Recent foundation models scale this through heterogeneous pretraining for strong zero-shot performance \citep{ansari2025chronos, cohen2025time, auer2025tirex}. Multi-task models such as MOMENT \citep{goswami2024moment}, UniTS \citep{gao2024units}, and GPT4TS \citep{zhou2023one} additionally span forecasting, classification, imputation, and anomaly detection, but typically use task-specific heads or adaptation rather than demonstration-conditioned task switching. Patching, popularized by \citet{nie2022time}, enables efficient long-context modeling; our model augments patch representations with semantic tokens and cross-example task inference.

\textbf{Contextual conditioning and hierarchical structure.}
Richer context, covariates, retrieval, and example selection improve forecasting \citep{das2024context, ansari2025chronos, auer2025tirex}. However, prior contextual conditioning generally supports a fixed forecasting task, with task semantics encoded in parameters or objectives. Our work instead treats input-output \textbf{demonstrations} as instructions defining the task mapping at inference. The hierarchy models temporal, target-covariate interactions, summarizes regions with semantic tokens, and uses cross-example attention to infer demonstrated mappings. Table~\ref{tab:icl_comparison} compares our model with existing time-series foundation models.

\section{Methodology}
\label{sec:methodology}

\subsection{Instruction-Conditioned ICL Formulation}

We consider a time-series learning setting in which the task is not specified by a
task identifier or task-specific head, but inferred from example demonstrations.
Let
$
X=\{x^{(1)},\ldots,x^{(d_x)}\}
$
denote target components and
$
Z=\{z^{(1)},\ldots,z^{(d_z)}\}
$
denote covariates, where each component $x^{(j)} \in \mathbb{R}^{T}$ is 
a one-dimensional time series with a
binary observation mask $m^{(j)} \in \{0,1\}^{T}$. 
The framework supports univariate and multivariate
targets, past-only covariates, future-known covariates, and no-covariate settings. 
Categorical covariates are ordinal-encoded and repeated across time as
constant one-dimensional covariate series.

An in-context example is
\(
\mathcal{E}_i =
(X_i^{\mathrm{hist}}, Z_i^{\mathrm{hist}},
 X_i^{\mathrm{fut}}, Z_i^{\mathrm{fut}}),
\)
while a query is
\(
\mathcal{Q} =
(X_q^{\mathrm{hist}}, Z_q^{\mathrm{hist}}, Z_q^{\mathrm{fut}}),
\)
where query future targets are withheld. Given examples
$\mathcal{S}=\{\mathcal{E}_i\}_{i=1}^{N}$, the model predicts
\(
f_\theta(\mathcal{Q}\mid \mathcal{S}) \rightarrow Y_q^{\mathrm{fut}}
\)
without parameter updates. Thus, examples act as implicit instructions that
define the task through demonstrated input-output mappings.

\begin{figure}[t]
    \centering
    \includegraphics[width=\linewidth]{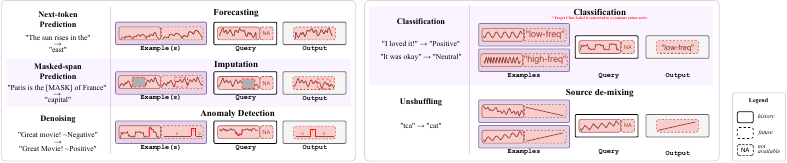}
    \caption{
        NLP tasks and analogous \textsc{iAmTime} prompt structures.
        Only target components of examples and query are shown, 
        tokens and covariates are omitted.
        Examples in Fig.~\ref{fig:icl_structure} are altered accordingly.
    }
    \label{fig:iclTasksandAnalogy}
\end{figure}

\paragraph{Multi-task Support} For classification, a class is encoded as a constant target time-series having the same length across 
\(
X_i^{\mathrm{fut}}
\). 
Class codes are evenly spaced and freshly permuted per instance, so the demonstrations $\{\mathcal{E}_i\}_{i=1}^{N}$ define their meanings.
Other tasks, like source separation, anomaly detection, and reconstruction, follow the same construction as forecasting detailed in
Fig.~\ref{fig:iclTasksandAnalogy} and Appendix~\ref{app:training_details}.

\subsection{Structured Tokenization and Input \& Patch Encoding}

To preserve semantic structure, we represent each example and query using explicit role tokens:
\[
[\texttt{START}], [\texttt{TARGET}], [\texttt{EXOG}],
[\texttt{MID}], [\texttt{FUTURE\_EXOG}], [\texttt{END}].
\]
The semantic roles of the tokens, including their corresponding series, temporal regions, and 
functional interpretation of the tokens are summarized in 
Table ~\ref{tab:token_semantics}. 
An example is represented as:
\[
\mathsf{Ser}(\mathcal{E}_i)=
[\texttt{START}]
[\texttt{TARGET}]X_i^{\mathrm{hist}}
[\texttt{EXOG}]Z_i^{\mathrm{hist}}
[\texttt{MID}]
[\texttt{TARGET}]X_i^{\mathrm{fut}}
[\texttt{FUTURE\_EXOG}]Z_i^{\mathrm{fut}}
[\texttt{END}],
\]
while the query $\mathcal{Q}$ omits future targets. The full ICL \textbf{prompt} is:
\[
\mathcal{P}=
\mathsf{Ser}(\mathcal{E}_1)\oplus \cdots \oplus
\mathsf{Ser}(\mathcal{E}_N)\oplus \mathsf{Ser}(\mathcal{Q}).
\]
These tokens provide discrete anchors for building per-example representations,
aligning historical inputs with demonstrated outputs, and conditioning query
decoding on the inferred mapping. Full token semantics are given in
Appendix~\ref{app:input_and_token_details}.

Each time-series component $x^{(j)}$ (or $z^{(j)}$) is normalized independently using 
standardization followed by the $\sinh^{-1}$ transformation \citep{ansari2025chronos}, where statistics are computed from the historical segment. 
During classification, to compensate for normalization, the class code $c_i$ is preprocessed as $\tilde c_i=c_i\sigma_i+\mu_i$, which normalizes back to $c_i$ (Appendix~\ref{app:icl_classification_protocol}).

Following
\citet{ansari2025chronos}, each component $(j)$ is augmented with a relative time
index and observation mask. The normalized values $\tilde{x}$, time indices $r$, and masks $m$ are
partitioned into patches $(k)$ of length $p$, concatenated, and encoded using $f_\phi$:
\(
h_k^{(j)}
=
f_\phi\!\left(
[\tilde{x}^{(j)}_{(k)},\, r_{(k)},\, m^{(j)}_{(k)}]
\right),
f_\phi:\mathbb{R}^{3p}\rightarrow\mathbb{R}^{D}.
\)
Stacking over patches yields $H^{(j)}\in\mathbb{R}^{P\times D}$.
History and future segments are encoded independently and concatenated along the
patch dimension. Further detail is listed in Appendix~\ref{sec:patch_encoder}.

\subsection{Hierarchical Multi-Scope Transformer Encoder}

The encoder consists of $L$ layers that combine temporal, variate, token,
and cross-example interactions. Each consists of six attention mechanisms divided into 
\emph{patch stream} \& \emph{token stream} 
(Appendix~\ref{app:encoder_details}).

First, \textsc{Temporal Self-Attention} is applied
independently along the patch dimension of each component using T5-style
self-attention with rotary position embeddings (RoPE) \citep{su2024roformer}. 
This captures temporal dynamics
while remaining agnostic to the component's semantic role. Second, 
\textsc{Per-example Fusion Attention} is applied across the series dimension within each example or
query, modeling interactions between targets and covariates within the block.

The central \emph{token stream} consists of Token Read, Token Self-Attention,
Cross-Example Attention, and Token Write. 
In \textsc{Token Read}, structural tokens embeddings $T$ act as
queries over patch representations $H$ (serving as keys/values).
Additionally, to enforce the semantic roles of tokens 
(Table \ref{tab:token_semantics}) and 
prevent information leakage across unrelated regions 
(e.g., preventing a history token from accessing future target values), 
attention is explicitly constrained using \emph{allow masks} that define which subsets of the input each token can access.
The complete read/write region matrix is shown in
Figure~\ref{fig:token_region_matrix}. 
Tokens then interact through \textsc{Token Self-Attention}
within each example/query.

\textsc{Cross-Example Attention} is the core mechanism enabling ICL operation. Let
$T_{\mathrm{ex}}\in\mathbb{R}^{N\times R\times D}$ denote $N$ example-token
representations, and $T_q\in\mathbb{R}^{R\times D}$ the query-token
representations, with $R$ token slots. We restrict cross-example retrieval to the
\texttt{START} and \texttt{MID} tokens:
\[
[T_q^{\texttt{START}},T_q^{\texttt{MID}}]
\leftarrow
\mathrm{Attn}\!\left(
[T_q^{\texttt{START}},T_q^{\texttt{MID}}],
[T_{\mathrm{ex}}^{\texttt{START}},T_{\mathrm{ex}}^{\texttt{MID}}],
[T_{\mathrm{ex}}^{\texttt{START}},T_{\mathrm{ex}}^{\texttt{MID}}]
\right).
\]
This allows the query to retrieve demonstrated mappings between historical
inputs and future outputs, forming a latent task representation conditioned on
the example set.

Finally, \textsc{Token Write} injects the inferred task representation into
query patches using FiLM conditioning \citep{perez2018film}
enabling explicit conditioning of the prediction process.
The modulation is applied in two stages:
first, the updated \texttt{START} token generates global modulation parameters applied
to all query patches $H_q \in \mathbb{R}^{P_q \times D}$,
while the updated \texttt{MID} token generatesfuture-specific modulation 
applied only to forecast-horizon patches $H_q^{(\mathcal{I}_{\mathrm{fut}})}$ 
where, $\mathcal{I}_{\mathrm{fut}}$ denotes the indices of the horizon patches.
FiLM modulates using 
\(
H\odot(1+\gamma)+\beta.
\) where
$(\gamma, \beta)$ modulation parameters are obtained from the
\texttt{START} and \texttt{MID} tokens via learned projections,
detailed in Appendix~\ref{sec:token_stream}.

\subsection{Task-Conditioned Patch Decoder}

Following token write, the task-conditioned query representation $H_q$ 
is decoded using a lightweight mixture-of-experts (MoE) patch decoder. 
We first form a context vector
\(
c_q =
\mathrm{concat}\left(
h_{\texttt{START}}, h_{\texttt{MID}}, \mathrm{mean}(H_q^{\text{fut}})
\right)
\in \mathbb{R}^{3D},
\)
where $h_{\texttt{START}}$ and $h_{\texttt{MID}}$ are the 
$\texttt{START}$ and $\texttt{MID}$ query token embeddings. 
The context is projected using $W_q \in \mathbb{R}^{3D \times D}$ and routed over $E$ learned expert keys $K_{\text{exp}} \in \mathbb{R}^{E \times D}$, via scaled dot-product attention:
\( 
\alpha = \mathrm{softmax}(
     (c_q W_q K_{\text{exp}}^\top) / \sqrt{D} 
    ) \in \mathbb{R}^{E} 
\).
Each expert applies a patch decoder to the query representation,
$
Y_q^{(e)} = \mathrm{Dec}_e(H_q),
$
and the final prediction is the routed mixture:
\(
\hat{Y}_q^{\text{fut}}
=
\sum_{e=1}^{E} \alpha_e Y_q^{(e)}
\in \mathbb{R}^{|Q| \times H},
\)
where $|Q|$ is the number of quantiles and $H$ is the prediction horizon. 
This enables task-dependent interpolation among decoding behaviors while sharing the same encoder representation across experts.
During classification, the decoded output sequence is averaged along length dimension and mapped to the nearest episode-local code. 

\paragraph{Direct multi-horizon prediction.}
The decoder predicts the entire horizon jointly (up to the largest horizon limit fixed while post-training), improving efficiency and avoiding rollout error accumulation, which is important for long-horizon forecasting. Prior work has shown that direct multi-step prediction can outperform autoregressive decoding in time-series settings \citep{zeng2023transformers}.

\section{Training}
\label{sec:training_data}

Training data scale, diversity, and task structure are critical for time-series foundation models. 
We construct a large heterogeneous corpus from the Chronos pretraining corpus \citep{ansari2024chronos} and the GIFT-Eval pretraining corpus \citep{aksu2024gift}, while excluding datasets overlapping with downstream benchmarks such as fev-bench \citep{shchur2025fev}; dataset details are provided in Appendix~\ref{sec:datasets}. 
The resulting real-data pool contains approximately 30M univariate series from Chronos and 2.5M additional series from GIFT-Eval, spanning diverse domains, frequencies, and granularities.
In addition, we collect 200K univariate and multivariate time series from 
UCR classification datasets \citep{dau2019ucr, bagnall2018uea} 
with known labels, and spanning 15 domains.

\subsection{Data Augmentation and Training Mixture}

To increase coverage beyond observed datasets, we augment using complementary synthesis strategies.
We apply TSMixup \citep{ansari2024chronos}, 
and KernelSynth \citep{ansari2024chronos}, a Gaussian-process generator 
to produce controlled temporal patterns. 
We additionally construct multivariate systems with explicit endogenous-exogenous relationships by imposing linear, nonlinear, seasonal, shock-based, lagged, cointegration, and Granger-style dependencies among sampled univariate series. 
Finally, we generate specialized synthetic episodes for source separation and classification, allowing the model to observe controlled task mappings that are difficult to obtain at scale from labeled real data. 
Full augmentation procedures and parameter settings are provided in Appendix~\ref{sec:augmentation_construction}.

Across all sources, we pool approximately 72.5M uni- and multi-variate time series. 
During training, series are sampled from this pool using a fixed mixture: 10\% KernelSynth-generated, 50\% multivariate with covariates, and 40\% univariate series. 
These series are used to construct instruction-conditioned in-context episodes for forecasting, imputation/reconstruction, anomaly detection, classification, and source de-mixing (see App.~\ref{sec:task_classes}). 
The \textit{prompt construction}, \textit{task instantiations}, \textit{inference protocols}, \textit{curriculum learning} (App.~\ref{sec:curriculum}) steps, and \textit{hyperparameters} are described in Appendix~\ref{app:training_details}.

\subsection{Episodic Amortized Meta-Learning}
\label{sec:training_objective}

As illustrated in Figure~\ref{fig:iclTasksandAnalogy}, training is episodic: each sample is an instruction-conditioned prompt $\mathcal{P}$ containing support examples and a query. 
The support examples define the task through input-output demonstrations, and the model learns
\(
f_\theta(\mathcal{P}) \approx Y_q^{\mathrm{fut}},
\)
where $Y_q^{\mathrm{fut}}$ is the withheld query output. 
Different task families correspond to different interpretations of this output, e.g., a forecast, reconstruction, anomaly mask, class-code sequence, or separated latent component. 
This induces \emph{amortized} meta-learning: task adaptation is learned during training and executed at inference through contextual demonstrations, without parameter updates.

Training minimizes the expected query loss over prompts
by following a specific training curriculum (Appendix~\ref{sec:curriculum}).
The prompt distribution varies target dimensionality, covariate availability, example-query structural alignment, and the task semantics encoded by example futures. 
To encourage genuine ICL rather than query-only shortcuts, the training curriculum includes support-dependent forecasting transforms, episode-local label remapping for classification, and cross-task ambiguity episodes in which the same query window can require different outputs depending on the support demonstrations.

For probabilistic prediction, we use the pinball loss with $|Q|$ quantile levels. 
For a query $q$ with $d_x^{(q)}$ target components and horizon $H$, the loss is
\[
\mathcal{L}_{\text{QR}} =
\frac{1}{|Q| H}
\sum_{p \in Q} \sum_{t=1}^{H}
\max\big(p(y_t - \hat{y}_t^{(p)}), (p - 1)(y_t - \hat{y}_t^{(p)})\big),
\]
where $y \in \{x_{q}^{\text{fut},(j)}\}_{j=1}^{d_x^{(q)}}$ is the target and
$\hat{y}^{(p)} \in \{\hat{x}_{q}^{\text{fut},(j)}\}_{j=1}^{d_x^{(q)}}$ is the prediction at quantile level $p$.

\section{Experiments}
\label{sec:experiments}

We evaluate iAmTime on large-scale forecasting benchmarks and complementary non-forecasting tasks, assessing: 
(i) zero-shot performance across domains, frequencies, and horizons; 
(ii) benefits from variate- and covariate-informed inference; 
(iii) the effect of in-context demonstrations; and 
(iv) task adaptation using the same architecture.

\paragraph{Benchmarks.}
We use two comprehensive time-series foundation-model benchmarks. 
fev-bench \citep{shchur2025fev} contains 100 forecasting tasks across diverse domains, with and without covariates, while GIFT-Eval \citep{aksu2024gift} contains 24 datasets across short-, medium-, and long-horizon settings and multiple frequencies, yielding 97 evaluation settings. 
We ensure no overlap between the pretraining corpus and either benchmark.
For multi-task comparisons, following prior work \citep{gao2024units, goswami2024moment}, we use 10 datasets for classification, 4 for imputation, and 5 for anomaly detection. 
We additionally report classification results on 120 UCR datasets.

\paragraph{Baselines.}
We compare against recent time-series foundation models including TiRex-1.0/2.0 \citep{auer2025tirex, podest2026tirex}, Toto-2.0 \citep{khwaja2026toto}, Chronos-2 \citep{ansari2025chronos}, TimesFM-2.5 \citep{das2024context}, Moirai-2.0 \citep{woo2024unified}, and TabPFN-3 \citep{grinsztajn2026tabpfn}, together with AutoARIMA, AutoETS, and their ensemble.
For classification, we include ROCKET \citep{dempster2019rocket}, MiniROCKET \citep{dempster2021minirocket}, ICL-classifiers \citep{feofanov2026mantisv2, o2026rocketpfn}, and Chronos-2/Bolt with linear-probe heads \citep{alain2016understanding}. 
For other tasks, we compare against task-specific baselines and strong Transformer models.

\paragraph{Metrics.}
We follow each benchmark's official metrics. 
Forecasting is evaluated using mean absolute scaled error (MASE) for point accuracy and continuous ranked probability score (CRPS), approximated by mean weighted quantile loss (WQL), for probabilistic performance. 
Scores are normalized by the seasonal naive baseline and aggregated across tasks. 
Following \citet{shchur2025fev}, we report average win rate ($W$), the fraction of pairwise comparisons won, and skill score ($S$), the average percentage improvement over seasonal naive baseline. 
For classification we report accuracy and F1; for anomaly detection, precision and F1; and for imputation, MSE and MAE.

\subsection{Zero-Shot Forecasting and Gains from ICL}
\label{sec:results}

\begin{figure*}[t]
\centering

\begin{subfigure}[t]{.48\textwidth}
  \centering
  \resizebox{.99\linewidth}{!}{%
    \input{images/gift_eval_overall.pgf}
  }
  \caption{Results on the GIFT-Eval benchmark.}
  \label{fig:overall_gift_eval}
\end{subfigure}
\hfill
\begin{subfigure}[t]{.48\textwidth}
  \centering
  \resizebox{.99\linewidth}{!}{%
    \input{images/fev_bench_overall.pgf}
  }
  \caption{Results on the fev-bench benchmark.}
  \label{fig:overall_fev_bench}
\end{subfigure}

\vspace{0.6em}

\begin{subfigure}[t]{.32\textwidth}
  \centering
  \resizebox{.99\linewidth}{!}{%
    \input{images/imputation_overall.pgf}
  }
  \caption{Imputation results.}
  \label{fig:overall_imputation}
\end{subfigure}
\begin{subfigure}[t]{.33\textwidth}
  \centering
  \resizebox{.99\linewidth}{!}{%
    \input{images/classification_overall.pgf}
  }
  \caption{Classification results.}
  \label{fig:overall_classification}
\end{subfigure}
\begin{subfigure}[t]{.33\textwidth}
  \centering
  \resizebox{.99\linewidth}{!}{%
    \input{images/anomaly_detection_overall.pgf}
  }
  \caption{Anomaly detection results.}
  \label{fig:overall_anomaly_detection}
\end{subfigure}

\caption{
  Aggregate performance across forecasting, imputation, classification,
  and anomaly detection benchmarks. Results are evaluated using
  five seeds, with standard deviations reported in the plots.
  For non-forecasting tasks, each model (except iAmTime) is separately trained on each dataset.
}
\label{fig:overall_performance}
\end{figure*}

\begin{table}[t]
\caption{
The average win rate and skill score with respect to WQL
metric, on the fev-bench dataset. 
Higher values are better for both.
Baseline results and the imputation
strategy for handling data leakage in certain tasks are both taken 
from \citealt{shchur2025fev}. 
Bold and underline indicate the best and second-best performance, respectively.
}
\label{tab:win_skill_fev}
\centering
\setlength{\tabcolsep}{2pt}
\resizebox{\columnwidth}{!}{
    \begin{tabular}{lrrrrrrrrrrrrrrrrr}
    \toprule
    Model & iAmTime & Toto-2 & Toto-2 & Chronos & Toto-2 & TiRex & TimesFM  & TabPFN & CITRAS & TabPFN & Toto & Moirai & Toto-2 & Chronos & Stat. & Auto & Seasonal \\
     & w ICL & 2.5B & 1B & 2 & 313m & 2 & 2.5 & TS-3 & FM & TS & 1.0 & 2.0 & 4m & Bolt & Ens. & ARIMA & Naive \\
    \midrule
    Avg. Win Rate (\%) & \textbf{75.9} & \underline{74.6} & 74.4 & 73.8 & 71.6 & 68.0 & 64.2 & 59.2 & 51.4 & 50.6 & 50.2 & 47.2 & 44.6 & 40.7 & 25.0 & 21.1 & 6.7 \\
    Skill Score (\%) & \textbf{51.6} & 49.2 & 49.2 & \underline{51.5} & 49.0 & 49.5 & 50.8 & 48.8 & 46.0 & 46.6 & 45.3 & 44.8 & 44.7 & 43.2 & 22.4 & 24.5 & 0.0 \\
    Median runtime (s) & 1.8 & 23.8 & 11.5 & 2.7 & 4.5 & 1.4 & 16.8 & 1795.8 & 3.8 & 725.5 & 91.9 & 2.5 & \textbf{0.9} & \underline{1.0} & 725.8 & 198.7 & 2.3 \\
    Data Leakage & 0.0 & 0.0 & 0.0 & 0.0 & 0.0 & 0.0 & 10.0 & 0.0 & 0.0 & 0.0 & 8.0 & 28.0 & 0.0 & 0.0 & 0.0 & 0.0 & 0.0 \\
    \bottomrule
    \end{tabular}
}
\end{table}

\begin{table}[t]
\caption{
   Average Skill Score (\%) w.r.t. SQL of iAmTime with and without demonstrations, on various subsets of fev-bench.
    The \textbf{best} and \underline{second-best} scores are highlighted.
 }
\label{tab:icl_covariates_effect_fev}
\centering
\setlength{\tabcolsep}{3pt}
\resizebox{\columnwidth}{!}{
   \begin{tabular}{llcccccccccccc}
      \toprule
      & & iAmTime & iAmTime & Chronos & TiRex & TimesFM & Toto & TabPFN & CITRAS & Moirai & Chronos & Stat. & Seasonal \\
      Subset & Metric & w/ ICL & w/o ICL & 2 & 2 & 2.5 & 1.0 & TS-3 & FM & 2.0 & Bolt Base & Ensemble & Naive \\

      \midrule
      \multirow[t]{2}{*}{Univariate} & Avg. Win Rate (\%) & \textbf{81.50} & \underline{75.90} & 74.40 & 67.60 & 60.20 & 39.20 & 54.00 & 39.80 & 39.20 & 34.90 & 27.60 & 5.70 \\
      & Skill Score (\%) & \textbf{37.80} & \underline{37.10} & 37.00 & 35.40 & 34.90 & 31.50 & 32.70 & 31.30 & 30.00 & 30.10 & 16.60 & 0.00 \\
      \midrule

      \multirow[t]{2}{*}{Multivariate} & Avg. Win Rate (\%) & 
      \textbf{78.30} & 74.50 & 72.70 & 71.00 & 60.80 & \underline{75.50} & 35.70 & 46.20 & 40.20 & 31.80 & 11.70 & 1.60 \\
      & Skill Score (\%) & \textbf{58.20} & \textbf{58.20} & \underline{57.90} & 56.70 & 56.80 & 57.40 & 53.60 & 54.20 & 54.70 & 52.10 & 24.90 & 0.00 \\
      \midrule

      \multirow[t]{2}{*}{Covariate} & Avg. Win Rate (\%) & 
      \underline{73.30} & \textbf{73.50} & 72.30 & 70.80 & 63.00 & 42.00 & 52.60 & 50.60 & 41.80 & 36.40 & 21.30 & 3.60 \\
      & Skill Score (\%) & \textbf{48.40} & \textbf{48.40} & 47.00 & 44.80 & \underline{47.80} & 35.90 & 43.30 & 39.00 & 37.10 & 35.90 & 21.00 & 0.00 \\

      \bottomrule
   \end{tabular}
}
\end{table}

Across both benchmarks, iAmTime achieves competitive or superior performance against strong foundation-model baselines, with the largest gains on heterogeneous tasks, covariate-rich settings, and longer horizons, suggesting instruction-conditioned demonstrations and amortized meta-learning provide benefits beyond scaling historical context alone.

On GIFT-Eval, iAmTime achieves the lowest aggregate CRPS and MASE (Figure~\ref{fig:overall_gift_eval}), outperforming all competing foundation models, while task-specific deep models and classical baselines lag behind. 
Figures \ref{fig:gift_long_agg},\ref{fig:gift_mid_agg},\ref{fig:gift_short_agg} further show iAmTime's strong performance across short-, medium-, and long-horizon settings. 
On fev-bench, iAmTime achieves the strongest overall performance (Figure~\ref{fig:overall_fev_bench}), with the lowest aggregate CRPS and second-lowest MASE across 100 tasks. It also obtains the highest average win rate and skill score (Table~\ref{tab:win_skill_fev}), indicating broad rather than highly localized gains.

To isolate the effect of demonstrations, we follow \citet{ansari2025chronos} and partition fev-bench into 32 univariate tasks without covariates, 26 multivariate tasks without covariates, and 42 covariate-informed tasks with past-only or future-known covariates. iAmTime uses four examples per task. Table~\ref{tab:icl_covariates_effect_fev} shows that demonstrations improve univariate performance and provide smaller gains on multivariate tasks, indicating effective use of support examples through the hierarchical ICL mechanism. They do not further improve covariate-informed tasks, suggesting that explicit covariates already provide sufficient conditioning and additional examples may add less relevant context.

We don't compare against ensembles or agentic methods.
Appendix~\ref{sec:extend_evaluations_appendix} reports additional GIFT-Eval results across horizons, frequencies, and covariates. Appendix~\ref{sec:eval_inference_protocol} describes the inference protocol.
Figure~\ref{fig:gift_eval_prediction_visualizations} in Appendix~\ref{app:icl_adaptation_viz} shows qualitative examples of iAmTime's zero-shot forecasting.

\subsection{Classification Tasks}
\label{sec:classification_tasks}

\begin{figure}[t]
\centering
\begin{subfigure}{.33\textwidth}
  \centering
  \resizebox{0.99\linewidth}{!}{\input{images/classification_univariate.pgf}}
  \caption{
    Univariate subset.
  }
  \label{fig:overall_univariate_eval}
\end{subfigure}
\hfill
\begin{subfigure}{.33\textwidth}
  \centering
  \resizebox{0.99\linewidth}{!}{\input{images/classification_multivariate.pgf}}
  \caption{
    Multivariate subset.
  }
  \label{fig:overall_multivariate_eval}
\end{subfigure}
\hfill
\begin{subfigure}{.32\textwidth}
  \centering
  \resizebox{0.99\linewidth}{!}{\input{images/classification_icl_task_adaptation.pgf}}
  \caption{
    ICL task adaptation subset.
  }
  \label{fig:overall_icl_task_adaptation_eval}
\end{subfigure}
\caption{
    Aggregate performance on 120 classification tasks: 
    Uni and Multivariate subsets using Linear Probe (LP), 
    and the ICL task-adaptation subset
    from 120 UCR datasets
    (higher is better).
}
\label{fig:overall_classification_ucr_eval}
\end{figure}

We evaluate classification in two settings: \emph{in-context classification}, where the model predicts a query class from a few labeled demonstrations without parameter updates or task-specific head; and frozen-embedding linear probe, which measures representation quality independently of the ICL mechanism. Figure~\ref{fig:overall_classification} shows that iAmTime (w/ ICL) learns meaningful classification behavior from demonstrations, outperforming many baselines, and iAmTime+LP learns competitive representations.

\paragraph{ICL classification via demonstrations.}
Each support example contains a time series and a class label encoded as a constant output sequence; the query contains only the input series, requiring the model to infer the episode-local label mapping from demonstrations using the same architecture and decoding interface as forecasting. Figure~\ref{fig:overall_icl_task_adaptation_eval} shows strongest performance in low-cardinality settings, and degrades gracefully with class-cardinality as more episode-local class codes must be inferred from limited support. 
Overall, iAmTime outperforms almost all baselines (Fig.~\ref{fig:overall_classification}) demonstrating pure demonstration-conditioned classification during inference without a task-specific head.

\paragraph{Embedding-based linear probes.}
We also extract frozen iAmTime embeddings and train linear probes on 120 UCR classification splits. We compare against Chronos-family forecasting encoders under the same protocol and dedicated supervised classifiers such as ROCKET and MiniROCKET. As shown in Fig.~\ref{fig:overall_classification_ucr_eval}, iAmTime learns competitive representations among forecasting foundation models and improves over Chronos-family encoders on several subsets, particularly where native multivariate structure is informative. Dedicated TSC methods remain strong specialized baselines, so this experiment measures representation quality rather than replacing task-specific classifiers.

Detailed classification protocols, datasets, and per-dataset results are provided in Appendix~\ref{app:classification_details}.

\subsection{Imputation Tasks}
\label{sec:imputation_tasks}

We evaluate block imputation by masking varying percentages of contiguous patches of length 8 and reconstructing the missing values. As shown in Fig.~\ref{fig:overall_imputation}, iAmTime with four in-context examples achieves the best aggregate MSE and MAE, outperforming all models including multi-task models such as MOMENT and UniTS. The substantial gap between iAmTime with demonstrations and the same checkpoint evaluated with zero examples isolates the benefit of demonstration-conditioned inference: demonstrations change the task from forecasting (default) to imputation, thereby achieving reconstruction without parameter updates. These results show that the learned ICL mechanism extends beyond forecasting to structured missing-value recovery within the same model and interface.

\subsection{Anomaly Detection Tasks}
\label{sec:anomaly_detection_tasks}

We evaluate two anomaly-detection interfaces using the same iAmTime checkpoint. In \emph{direct-mask} prediction (iAmTime-DM), demonstrations specify a binary anomaly-mask output and the model predicts the query mask directly; in \emph{reconstruction} mode (iAmTime-RC), demonstrations instead specify reconstruction, and anomalies are identified from reconstruction error as in conventional approaches. Figure~\ref{fig:overall_anomaly_detection} shows that direct-mask prediction already outperforms all baselines, including multi-task models MOMENT and UniTS, while reconstruction achieves the strongest overall performance. Importantly, the two settings require markedly different output semantics, yet iAmTime switches between them solely through the demonstrations supplied at inference time. 
Moreover, iAmTime (w/o) ICL fails to perform reliable anomaly detection.
This provides direct evidence that the model adapts the task and output format through context rather than relying on a fixed task-specific head.

\subsection{Task Adaptation: Demonstrations as Task Specifications}
\label{sec:demonstration_interventions}

\begin{table}[t]
\centering
\caption{
Inference-time interventions on demonstrations using iAmTime.
Reordering intact demonstrations has negligible effect, while corrupting or removing them degrades task adaptation.
}
\label{tab:demonstration_interventions}
\centering
\setlength{\tabcolsep}{2pt}
\resizebox{\columnwidth}{!}{

\begin{tabular}{lccccc}
\toprule
Changes to Demonstrations
& \begin{tabular}[c]{@{}c@{}}Forecasting\\(fev-univ \\Skill Score \% $\uparrow$)\end{tabular}
& \begin{tabular}[c]{@{}c@{}}Forecasting\\(GIFT Univ.\\CRPS $\downarrow$)\end{tabular}
& \begin{tabular}[c]{@{}c@{}}Classification\\(UCR ICL\\Acc. $\uparrow$)\end{tabular}
& \begin{tabular}[c]{@{}c@{}}Imputation\\Block-8\\(MSE/MAE $\downarrow$)\end{tabular}
& \begin{tabular}[c]{@{}c@{}}Anomaly\\Detection\\(F1 $\uparrow$)\end{tabular}
\\
\midrule

No Change (Correct examples)
& 37.80
& 0.354
& 0.813
& 0.0727 / 0.1265
& 93.57
\\

Intact Input-Output pairs shuffled
& 37.81
& 0.354
& 0.812
& 0.0725 / 0.1270
& 93.54
\\

Outputs only shuffled across inputs
& 30.92
& 0.491
& 0.110
& 0.2148 / 0.2296
& 71.34
\\

Example outputs masked
& 36.77
& 0.347
& 0.211
& 0.1695 / 0.1948
& 78.46
\\

Unrelated-task examples
& 20.36
& 0.912
& 0.184
& 0.2386 / 0.2471
& 67.82
\\

No examples
& 37.10
& 0.351
& 0.206
& 0.1730 / 0.1967
& 78.09
\\

\bottomrule
\end{tabular}
}
\end{table}

The architecture and instruction-conditioned meta-training already yield competitive forecasts, while examples provide modest, context-dependent gains. However, their main role is \emph{task specification and adaptation}. Table~\ref{tab:demonstration_interventions} shows that reordering intact pairs has negligible effect, whereas breaking the input-output mapping causes broad degradation; shuffling outputs reduces classification accuracy from $0.813$ to $0.110$ and strongly harms imputation, anomaly detection, and forecasting.
Providing unrelated-task demonstrations (e.g., classification examples for a forecasting query) degrades performance the most, showing that they are crucial and essential for task specification.
Appendix~\ref{app:icl_adaptation_viz} visualizes the effect of demonstrations on task adaptation, showing that the model can learn to perform different tasks on the same query depending on the examples provided.
Appendix~\ref{app:mechanistic_analysis} further examines its internal mechanisms, attention representation, task-state restorations, and decoder routing.


\subsection{Ablation Study on Architecture and Multi-Task Adaptation}

\begin{table}[t]
\centering
\caption{
Fine-grained ablations of training, structural, task-adaptation, output-specialization, task-family, and curriculum components of iAmTime.
}
\label{tab:fine_grained_ablation}
\scriptsize
\setlength{\tabcolsep}{2pt}
\renewcommand{\arraystretch}{1.12}

\resizebox{\columnwidth}{!}{
\begin{tabular}{lllccccc}
\toprule
Ablation Type
& Variant
& Change from full iAmTime
& \makecell[c]{Forecasting\\fev Skill\\Score (\%) $\uparrow$}
& \makecell[c]{Forecasting\\GIFT\\CRPS $\downarrow$}
& \makecell[c]{Classification\\UCR ICL\\Acc./F1 $\uparrow$}
& \makecell[c]{Imputation\\Block-8\\MSE/MAE $\downarrow$}
& \makecell[c]{Anomaly\\Detection\\F1 $\uparrow$}
\\
\midrule

Full system
& iAmTime
& Full System
& 46.8
& 0.465
& 0.813 / 0.775
& 0.0727 / 0.1265
& 93.57
\\

\midrule

\multirow[c]{3}{*}{Training}
& -NoMeta
& Forecasting tasks and forecasting examples only
& 46.4
& 0.468
& 0.224 / 0.176
& 0.1740 / 0.1968
& 74.30
\\

& -NoExmp
& No training demonstrations; auxiliary meta-tasks also removed
& 46.4
& 0.471
& 0.238 / 0.184
& 0.1935 / 0.2257
& 72.10
\\

& -Meta-NoExmp
& Perform multi-task training; remove examples and Token Stream
& 47.5
& 0.458
& 0.240 / 0.191
& 0.0814 / 0.1305
& 92.20
\\

\midrule

\multirow[c]{2}{*}{Structural}
& -NoToks-WExmp
& Remove Token Stream but retain examples and multi-task training
& 38.6
& 0.569
& 0.502 / 0.521
& 0.1048 / 0.1542
& 81.55
\\

& -NoToks-NoExmp
& Remove Token Stream, examples, and auxiliary meta-tasks
& 46.2
& 0.489
& 0.207 / 0.158
& 0.1430 / 0.1767
& 76.80
\\

Variate Repr.
& -NoPEF
& Remove Per-example fusion attention (Patch Stream)
& 37.8
& 0.583
& 0.515 / 0.531
& 0.0981 / 0.1498
& 80.31
\\

\midrule

\multirow[c]{3}{*}{Task Adapt.}
& -NoCEA
& Remove token-level Cross-Example Attention (Token Stream)
& 44.8
& 0.494
& 0.263 / 0.198
& 0.1404 / 0.1607
& 78.72
\\

& -NoTWrite
& Remove Token Write/FiLM
& 43.9
& 0.516
& 0.246 / 0.194
& 0.1709 / 0.1954
& 79.05
\\

& -SharedToks
& Tie role-token embeddings while preserving tokens and masks
& 44.9
& 0.496
& 0.689 / 0.645
& 0.0962 / 0.1479
& 89.53
\\

\midrule

O/p Specz.
& -UniMoE
& Replace adaptive routing with uniform expert weights
& 45.8
& 0.479
& 0.776 / 0.734
& 0.0798 / 0.1339
& 91.84
\\

\midrule

\multirow[c]{4}{*}{Task-family}
& -NoCls
& Remove only classification episodes
& 46.3
& 0.468
& 0.274 / 0.218
& 0.0734 / 0.1272
& 93.21
\\

& -NoImp
& Remove only imputation episodes
& 46.2
& 0.469
& 0.808 / 0.770
& 0.1421 / 0.1784
& 93.06
\\

& -NoAD
& Remove only anomaly-detection episodes
& 46.3
& 0.468
& 0.810 / 0.772
& 0.0736 / 0.1271
& 80.64
\\

& -NoSynthCls
& Remove only synthetic classification episodes
& 46.9
& 0.460
& 0.611 / 0.568
& 0.0800 / 0.1259
& 85.41
\\

\midrule

Curriculum
& -NoCurriculum
& Perform multi-task training without curriculum
& 30.2
& 0.701
& 0.524 / 0.581
& 0.1878 / 0.2413
& 66.26
\\

\bottomrule
\end{tabular}
}
\end{table}

Beyond the inference-time no-example ablation in Table~\ref{tab:icl_covariates_effect_fev}, Table~\ref{tab:fine_grained_ablation} isolates training, structural, task-adaptation, task-family, and curriculum components. Appendix~\ref{sec:ablation_appendix} additionally reports Tables~\ref{tab:ablation_fev} and~\ref{tab:ablation_gift} for forecasting against baseline models, and Table~\ref{tab:classification_ablation_ucr} for UCR in-context classification.

The ablations reveal a coherent division of labor. Curriculum learning and Per-Example Fusion (PEF) are the dominant forecasting mechanisms, while Cross-Example Attention (CEA) and Token Write/FiLM incur only modest forecasting costs but are critical for task adaptation: removing either sharply reduces classification accuracy and approximately doubles imputation error. Distinct token roles benefit all tasks, while adaptive MoE routing provides a smaller but consistent gain. Task-family removals are selective: NoCls, NoImp, and NoAD primarily damage classification, imputation, and anomaly detection, respectively, showing that these are not simply weaker checkpoints. Notably, Meta-NoExmp slightly improves forecasting while largely collapsing ICL classification, indicating a small tradeoff between peak forecasting and broader task adaptation. Overall, the results show that PEF and curriculum primarily drive forecasting quality, whereas CEA, Token Write, semantic tokens, demonstrations, and multi-task instruction-conditioned training enable reliable inference-time task specification and adaptation.



\section{Conclusion}

We introduced iAmTime, an instruction-conditioned time-series foundation model that adapts through in-context demonstrations rather than task-specific fine-tuning. By combining semantic tokenization, hierarchical multi-scope attention, and task-conditioned decoding, the model infers task mappings from examples and applies them to queries within the forward pass. Our results show that this enables strong zero-shot forecasting and  non-forecasting task adaptation, suggesting a path toward reusable time-series foundation models whose behavior can be specified through contextual examples.

\subsection{Limitations and Future Work}
\label{sec:limitations}

iAmTime remains sensitive to data quality, distribution shift, and the relevance of in-context examples. Its outputs should be used as decision support, especially in high-stakes settings. Future work should improve example selection, expand meta-training task distributions, study calibration under shift, and extend instruction-conditioned adaptation to richer domains and multimodal time-series settings.

\newpage
\bibliographystyle{abbrvnat}
\bibliography{main}


\clearpage
\appendix

\section{ICL Prompts and Task Adaptation}
\label{app:icl_adaptation_viz}

\setlength{\remainingpageheight}{%
    \dimexpr\pagegoal-\pagetotal-\baselineskip\relax
}
\setlength{\promptimageheight}{\remainingpageheight}
\noindent
\begin{minipage}[t][\remainingpageheight][s]{\linewidth}
    \centering
    \vspace{0pt}

    \includegraphics[
        width=\linewidth,
        height=\promptimageheight,
        keepaspectratio
    ]{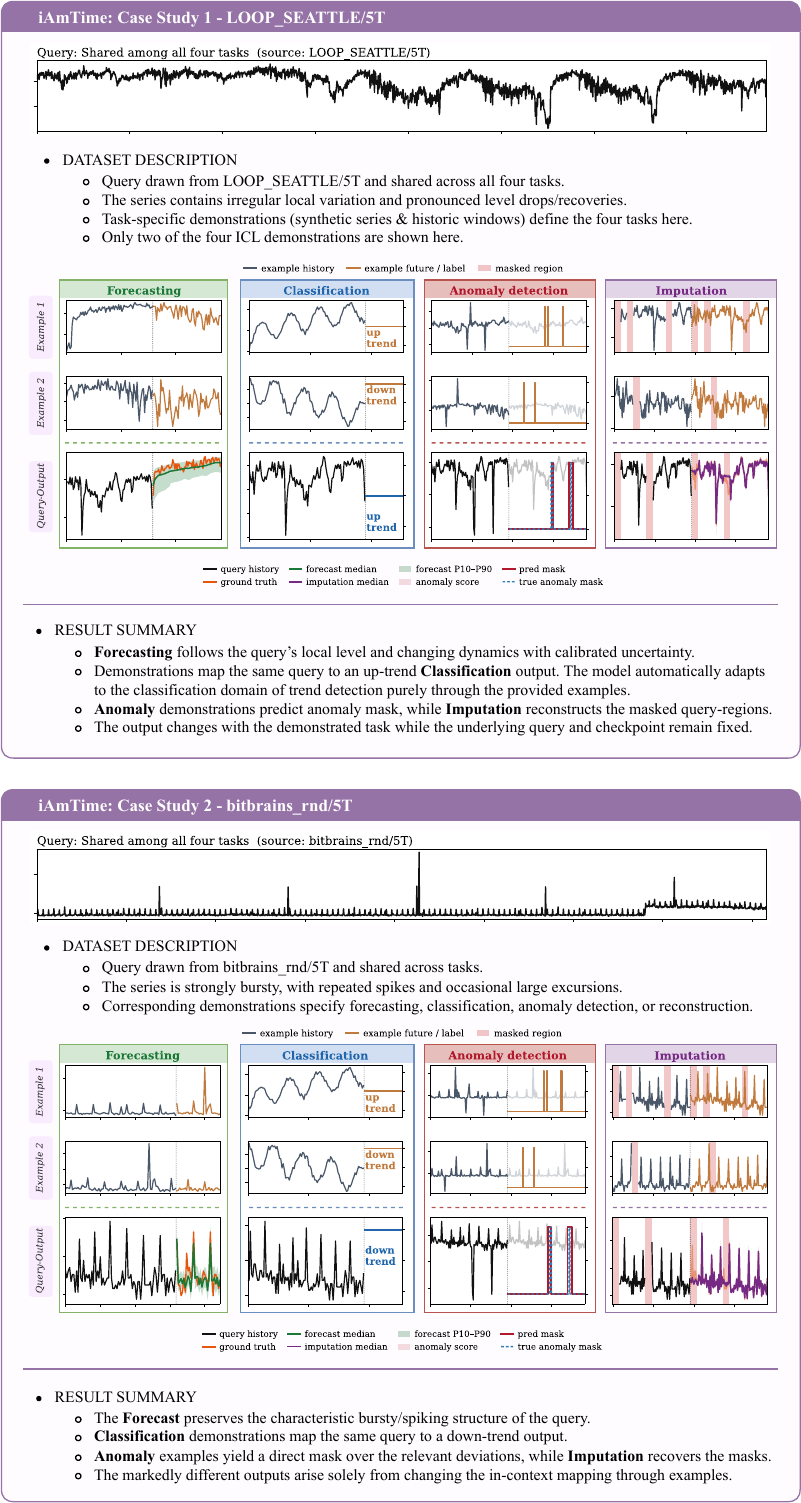}
\end{minipage}

\clearpage

\noindent
\begin{minipage}[t][0.99\textheight][s]{\linewidth}
    \vspace{0pt}
    \centering

    \includegraphics[
        width=\linewidth,
        height=0.47\textheight,
        keepaspectratio
    ]{images/ICL\_prompt\_3.pdf}

    \vspace{10pt}

    \includegraphics[
        width=\linewidth,
        height=0.088\textheight,
        keepaspectratio
    ]{images/forecast\_visual\_1.pdf}

    \vfill

    \includegraphics[
        width=\linewidth,
        height=0.088\textheight,
        keepaspectratio
    ]{images/forecast\_visual\_2.pdf}

    \vfill

    \includegraphics[
        width=\linewidth,
        height=0.088\textheight,
        keepaspectratio
    ]{images/forecast\_visual\_3.pdf}

    \vfill

    \includegraphics[
        width=\linewidth,
        height=0.088\textheight,
        keepaspectratio
    ]{images/forecast\_visual\_4.pdf}

    \vfill

    \includegraphics[
        width=\linewidth,
        height=0.088\textheight,
        keepaspectratio
    ]{images/forecast\_visual\_5.pdf}

    \captionof{figure}{
        Visualization of \textsc{iAmTime}'s predictions on some of the
        GIFT-Eval tasks. Comparing 4-example ICL and no ICL.
    }
    \label{fig:gift_eval_prediction_visualizations}
\end{minipage}

\clearpage

\section{Methodology and Architecture Details}
\label{app:method_details}

\begin{figure}[htbp]
    \centering
    \includegraphics[
        width=0.99\linewidth,
        keepaspectratio
    ]{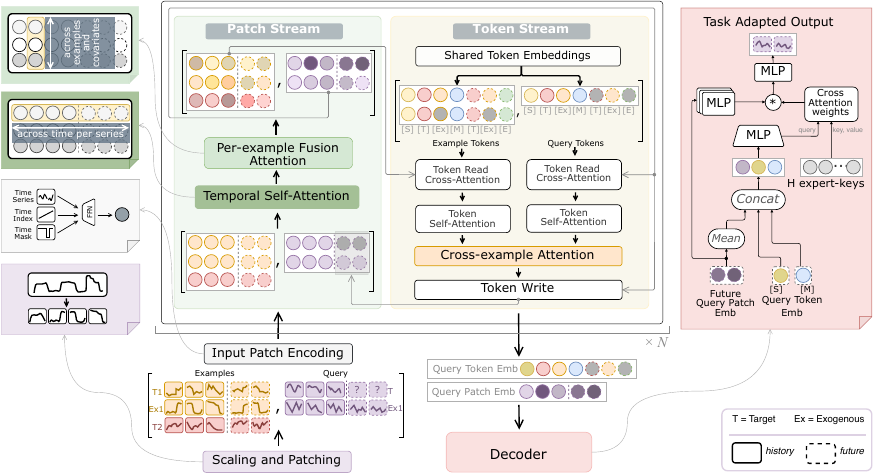}
    \caption{
        \textsc{iAmTime} architecture (center).
        The components on the \textbf{left} represent the patch-stream computations.
        The \textbf{right} side shows task adaptation through the decoder's mixture-of-experts.
    }
    \label{fig:iamtime_arch}
\end{figure}

\subsection{ICL Input Construction and Token Semantics}
\label{app:input_and_token_details}

This section provides additional details on the construction of the in-context
learning (ICL) input used by iAmTime. The main paper introduces the
high-level prompt structure; here we describe the underlying
components, masks, covariates, examples, queries, and semantic tokens in
detail.

\paragraph{Time-series components and masks.}
The most granular object in our formulation is a one-dimensional time-series
component
\(
x^{(j)} = (x^{(j)}_1,\ldots,x^{(j)}_T) \in \mathbb{R}^{T},
\)
where $j$ indexes a component. Each component is associated with a binary
observation mask
\(
m^{(j)} = (m^{(j)}_1,\ldots,m^{(j)}_T), 
\qquad 
m^{(j)}_t \in \{0,1\},
\)
where $m^{(j)}_t=1$ indicates that the value at timestep $t$ is observed, and
$m^{(j)}_t=0$ indicates that it is missing or intentionally withheld. Missing
values are replaced with zeros after mask construction, so that the model can
distinguish true zeros from unobserved entries through the mask channel.

\paragraph{Targets and covariates.}
A multivariate time-series instance is represented as a collection of target
components and covariates:
\[
X = \{x^{(1)},\ldots,x^{(d_x)}\}, 
\qquad
Z = \{z^{(1)},\ldots,z^{(d_z)}\},
\]
where $d_x$ denotes the number of target components and $d_z$ denotes the number
of covariate components. The target set $X$ may be univariate ($d_x=1$) or
multivariate ($d_x>1$), and each target instance may be accompanied by any
number of corresponding covariates. 
Covariates may be 
past-only covariates available over the historical window, i.e., $d_z^{\text{hist}} \geq 1$ and $d_z^{\text{fut}} = 0$;
or known covariates available over history and forecast horizon (e.g., calendar, price plan), i.e., $d_z^{\text{hist}}, d_z^{\text{fut}} \geq 1$.
The formulation also supports the special
case $d_z=0$, where no covariates are provided.
This flexibility enables the model to handle diverse 
real-world forecasting scenarios.

Categorical covariates are converted into scalar-valued covariate series by
ordinal encoding. Specifically, each categorical value is mapped to a real-valued
scalar and then repeated across time to form a constant one-dimensional
covariate series. This allows categorical, static, and dynamic covariates to be
represented using the same component-level interface.

\paragraph{Examples and query.}
An ICL episode consists of a set of example demonstrations followed by a query.
Each example contains both an input side and a demonstrated output side:
\[
\mathcal{E}_i =
\left(
X_i^{\mathrm{hist}}, Z_i^{\mathrm{hist}},
X_i^{\mathrm{fut}}, Z_i^{\mathrm{fut}}
\right),
\]
where $X_i^{\mathrm{hist}}$ and $Z_i^{\mathrm{hist}}$ denote historical target
and covariate components, while $X_i^{\mathrm{fut}}$ and
$Z_i^{\mathrm{fut}}$ denote future target values and future-known covariates.
Thus, each example demonstrates a mapping from historical inputs and available
future covariates to an output target sequence.

The query has the same input structure but omits the future target values:
\(
\mathcal{Q} =
\left(
X_q^{\mathrm{hist}},
Z_q^{\mathrm{hist}},
Z_q^{\mathrm{fut}}
\right).
\)
The model must predict the withheld output
\(
Y_q^{\mathrm{fut}} = X_q^{\mathrm{fut}}.
\)

For non-forecasting tasks, $Y_q^{\mathrm{fut}}$ may instead represent a
reconstructed sequence, a denoised signal, a class label encoded as a constant
series, or another task-specific output defined by the example demonstrations.

\paragraph{Semantic role tokens.}
To make the structure of each example and query explicit, we introduce learned
semantic tokens that identify the role of each region in the prompt. The token
set is
\[
\{
[\texttt{START}], [\texttt{TARGET}], [\texttt{EXOG}], 
[\texttt{MID}], [\texttt{FUTURE\_EXOG}], [\texttt{END}]
\}.
\]
In the implementation, these high-level segment tokens are used by the
token-read mechanism to summarize specific semantic regions. The semantic roles
are summarized in Table~\ref{tab:token_semantics}.

\begin{table}[t]
\caption{Semantic roles of structural tokens. Each token is associated with a specific subset of series and temporal regions, and encodes a distinct summary used for structured reasoning and task inference.
The query has no \texttt{TARGET\_FUT} and \texttt{END} tokens.
}
\label{tab:token_semantics}
\centering
\begin{tabular}{c|c|l}
\toprule
\textbf{Token} & \textbf{Series Index} & \textbf{Semantic Meaning (Token-Read)} \\
\midrule
\texttt{[START]} & - & global example summary \\
\texttt{[TARGET] (HIST)} & $X_i^{\text{hist}}$ or $X_q^{\text{hist}}$  & summary of past target behavior \\
\texttt{[EXOG]$_k$} & $Z_i^{\text{hist},(k)}$ or $Z_q^{\text{hist},(k)}$  & summary of past exogenous driver $k$ \\
\texttt{[MID]} & -  & summary of input conditions \\
\texttt{[TARGET] (FUT)} & $X_i^{\text{fut}}$  & summary of demonstrated output \\
\texttt{[FUTURE\_EXOG]$_k$} & $Z_i^{\text{fut},(k)}$ or $Z_q^{\text{fut},(k)}$  & summary of known future driver $k$ \\
\texttt{[END]} & -  & full example summary \\
\bottomrule
\end{tabular}
\end{table}

\paragraph{Example serialization.}

Each example is serialized as a structured prompt:
\[
\begin{aligned}
\mathsf{Ser}(\mathcal{E}_i) = 
&[\texttt{START}]\; \oplus \\
&[\texttt{TARGET}]\; \oplus_{j=1}^{d_x^{(i)}}
\left(
X_{i}^{\mathrm{hist},(j)}
\right) \oplus
[\texttt{EXOG}]\; 
\oplus_{k=1}^{d_z^{(i)}}
\left(
Z_i^{\text{hist},(k)}
\right) \oplus \\
&[\texttt{MID}]\; \oplus \\
&[\texttt{TARGET}]\; \oplus_{j=1}^{d_x^{(i)}}
\left(
X_{i}^{\mathrm{fut},(j)}
\right) \oplus
[\texttt{FUTURE\_EXOG}]\; 
\oplus_{k=1}^{d_z^{(i)}}
\left(
Z_i^{\text{fut},(k)}
\right) \oplus \\
&[\texttt{END}],
\end{aligned}
\]
where $\oplus$ denotes concatenation. The order of target and covariate
components is kept consistent within an episode. If a component is absent, its
corresponding token slot is marked invalid by the token-validity mask.

\paragraph{Query serialization.}
The query is serialized analogously, except that future target values and the
terminal \texttt{END} token are omitted:
\[
\begin{aligned}
\mathsf{Ser}(\mathcal{Q}) =
&[\texttt{START}]\; \oplus \\
&[\texttt{TARGET}]\; \oplus_{j=1}^{d_x^{(q)}}
\left(
X_{q}^{\mathrm{hist},(j)}
\right) \oplus
[\texttt{EXOG}]\; 
\oplus_{k=1}^{d_z^{(q)}}
\left(
Z_q^{\text{hist},(k)}
\right) \oplus \\
&[\texttt{MID}]\; \oplus \\
&[\texttt{FUTURE\_EXOG}]\; 
\oplus_{k=1}^{d_z^{(q)}}
\left(
Z_q^{\text{fut},(k)}
\right).
\end{aligned}
\]
Because the query future target is withheld, token slots corresponding to
\texttt{TARGET\_FUT} and \texttt{END} are invalid for the query block.

\paragraph{Full ICL prompt.}
The complete input prompt is obtained by concatenating $N$ examples followed by
the query:
\(
\mathcal{P}
=
\mathsf{Ser}(\mathcal{E}_1)
\oplus \cdots \oplus
\mathsf{Ser}(\mathcal{E}_N)
\oplus
\mathsf{Ser}(\mathcal{Q}).
\)
This construction makes the task explicit through demonstrations rather than a
task identifier. The example futures define the mapping to be performed, and the
query provides the input on which that mapping must be applied.

\paragraph{Role of explicit tokens.}
The explicit role and boundary tokens provide discrete anchors for the
hierarchical encoder. They allow the model to construct per-example
representations, align historical inputs with demonstrated outputs inside each
example, identify the query boundary, and condition decoding on mappings
retrieved from the example set. They also enable the token-read and token-write
mechanisms to enforce structured information flow through token validity masks
and token-to-region allow masks, described next.

\subsection{Patch-Based Time-Series Encoding}
\label{sec:patch_encoder}

Each time-series component (target or covariate) is processed independently to construct a structured patch representation. 
To each series $x^{(j)} \in \mathbb{R}^{T}$, we first apply \textit{standardization} followed by a $\sinh^{-1}$ transformation to stabilize scale and heavy-tailed distributions, as in \citet{ansari2025chronos}:
$
\tilde{x}^{(j)} = \sinh^{-1} \left( \frac{x^{(j)} - \mu^{(j)}}{\sigma^{(j)} + \epsilon} \right),
$
where, $\mu^{(j)}$ and $\sigma^{(j)}$ are computed using only the historical segment of the series \citep{kim2021reversible, burbidge1988alternative, uniejewski2018efficient}.

Next, we augment the representation with explicit structural information as in \citet{ansari2025chronos} by defining the relative time index:
$
r = \left[-\frac{T}{C}, \dots, 0, \dots, \frac{H-1}{C}\right],
$
where, $C$ denotes the maximum context length. 
Then, we partition the signal $\tilde{x}^{(j)}$, time index $j$, and mask $m^{(j)}$ into non-overlapping patches of length $p$:
\[
\tilde{x}^{(j)} \rightarrow \{\tilde{x}^{(j)}_{(1)}, \dots, \tilde{x}^{(j)}_{(P)}\}, \quad
r \rightarrow \{r_{(1)}, \dots, r_{(P)}\}, \quad
m^{(j)} \rightarrow \{m^{(j)}_{(1)}, \dots, m^{(j)}_{(P)}\},
\]
where, $P = \lceil T / p \rceil$. 
When the sequence length is not divisible by $p$, zero-padding is applied appropriately to the context (left) or future (right) segments.
Then for each patch index $k$, we concatenate and map into the model embedding space using a residual projection network:
\[
h^{(j)}_{k} = f_\phi(\left[\tilde{x}^{(j)}_{(k)}, \; r_{(k)}, \; m^{(j)}_{(k)}\right]), \quad f_\phi: \mathbb{R}^{3p} \rightarrow \mathbb{R}^{D},
\]
where, $D$ is the hidden dimension and $\phi$ denotes learnable parameters.
Stacking across patches yields the final representation:
$
H^{(j)} = \{h^{(j)}_{1}, \dots, h^{(j)}_{P}\} \in \mathbb{R}^{P \times D}.
$

History and future segments are encoded independently for each component and 
subsequently concatenated along the patch dimension, ensuring that temporal boundaries are preserved prior to higher-level attention operations.

\subsection{Hierarchical Multi-Scope Transformer Encoder Details}
\label{app:encoder_details}

This section provides additional implementation details for the hierarchical encoder described in Section~\ref{sec:methodology}. 
Each ICL prompt contains $N$ example blocks and one query block. 
For a block $b \in \{1,\ldots,N,q\}$, let
\(
H_b \in \mathbb{R}^{S_b \times P_b \times D}
\)
denote the patch representation, where $S_b$ is the number of target and covariate components, 
$P_b$ is the number of patches, and $D$ is the hidden dimension. 
Each block also contains structural token representations
\(
T_b \in \mathbb{R}^{R \times D},
\)
where $R$ is the number of token slots. 
Some slots may be invalid depending on the block structure; for example, query blocks do not contain future target values, so \texttt{TARGET} and \texttt{END} are masked out.

To avoid unrestricted global attention over all patches from examples/query,
and to preserve the information paths required for in-context learning,
each encoder layer applies six structured operations over two main components
\emph{Patch Stream} and \emph{Token Stream}.

\subsubsection{Patch Stream}

\paragraph{Temporal self-attention.}
Temporal self-attention is applied independently to each time-series component along the patch dimension:
\[
H_{b,s,:,:}
\leftarrow
\mathrm{Attn}_{\mathrm{time}}
\left(
H_{b,s,:,:},
H_{b,s,:,:},
H_{b,s,:,:}
\right).
\]
We use T5-style self-attention with rotary positional embeddings (RoPE) \citep{su2024roformer}, which have been widely used in modern transformer architectures \citep{ansari2025chronos,touvron2023llama}. 
Since this operation is applied independently per component, it captures temporal dependencies while remaining agnostic to whether the component is a target or covariate.

\paragraph{Per-example fusion attention.}
After temporal attention, we apply self-attention across the series dimension within each example or query block. 
For each patch index $k$,
\[
H_{b,:,k,:}
\leftarrow
\mathrm{Attn}_{\mathrm{series}}
\left(
H_{b,:,k,:},
H_{b,:,k,:},
H_{b,:,k,:}
\right).
\]
This operation models target-covariate interactions inside each block while preserving example 
(or query) boundaries before cross-example interaction is introduced.

\subsubsection{Token Stream}
\label{sec:token_stream}

\paragraph{Token Read (cross-attention).}
Token read converts patch representations into semantic summaries. 
For each block, tokens act as queries and flattened patch representations act as keys and values:
\[
T_b
\leftarrow
\mathrm{Attn}_{\mathrm{read}}
\left(
Q=T_b,\;
K=\mathrm{flat}(H_b),\;
V=\mathrm{flat}(H_b)
\right);
\quad
\mathrm{flat}(H_b)\in\mathbb{R}^{(S_bP_b)\times D}
\]

\begin{figure*}[t]
    \centering
    \includegraphics[
        width=0.99\linewidth,
        keepaspectratio
    ]{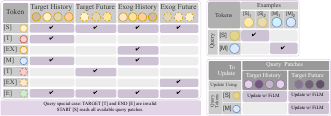}
    \caption{
        Token-patch interaction structure used in the encoder's token stream. \textbf{Left:} Token read cross-attention region matrix. 
        Each semantic token attends only to its designated input region, enforced via allow masks. 
        For query blocks, \texttt{TARGET (FUT)} and \texttt{END} are invalid, and \texttt{START} reads all available regions. 
        \textbf{Right (top):} Cross-example attention. The query retrieves information from the examples using corresponding \texttt{START} and \texttt{MID} tokens.
        \textbf{Right (bottom):} Token write region matrix. The query \texttt{START} token applies global FiLM conditioning to all query patches, whereas the query \texttt{MID} token applies additional FiLM conditioning only to future patches.
    }
    \label{fig:token_region_matrix}
\end{figure*}

Unlike unrestricted cross-attention, token read is constrained by an allow mask
\(
A_b \in \{0,1\}^{R \times (S_bP_b)}.
\)
It determines which patch regions each token may attend to. 
For example, \texttt{TARGET (HIST)}  attends only to target-history patches, \texttt{EXOG$_k$} attends only to covariate-history patches, \texttt{TARGET (FUT)} attends only to future targets, and \texttt{FUTURE\_EXOG$_k$} attends only to future-known covariates. 
The \texttt{MID} token attends to historical regions, while \texttt{START} and \texttt{END} attend globally where valid. 
For query blocks, \texttt{TARGET (FUT)} and \texttt{END} are invalid. 
The full token-region structure is shown in Figure~\ref{fig:token_region_matrix}.

This restricted attention prevents information leakage across semantic regions, while producing token summaries of meaningful time-series segments.

\paragraph{Token self-attention.}
After token read, tokens interact within each block:
\(
T_b
\leftarrow
\mathrm{Attn}_{\mathrm{tok}}(T_b,T_b,T_b).
\)
This allows information from target history, covariate history, future-known variables, and demonstrated outputs to be integrated across token slots. 

\paragraph{Cross-example attention.}
It is the main operation through which the query retrieves demonstrated mappings from the example set. 
Let $T_{\mathrm{ex}} \in \mathbb{R}^{N \times R \times D}$ denote the $N$ example-token 
representations and $T_q \in \mathbb{R}^{R \times D}$ denote the query-token representations. 
We restrict this to the \texttt{START}, \texttt{MID} tokens:
\[
U_q =
[
T_q^{\texttt{START}},
T_q^{\texttt{MID}}
],
\qquad
U_{\mathrm{ex}} =
[
T_{\mathrm{ex}}^{\texttt{START}},
T_{\mathrm{ex}}^{\texttt{MID}}
].
\]
The query summary tokens updates to
\(
U_q
\leftarrow
\mathrm{Attn}_{\mathrm{ex}}
\left(
Q=U_q,\;
K=U_{\mathrm{ex}},\;
V=U_{\mathrm{ex}}
\right),
\)
and written back into the corresponding query token slots. 
By attending to these tokens across examples, the query forms a latent task representation from demonstrated input-output mappings during inference.

\paragraph{Token write via FiLM conditioning.}
Finally, the updated query tokens are injected back into the query patch representation using feature-wise linear modulation (FiLM) \citep{perez2018film}. 

First the query's updated \texttt{START} and \texttt{MID} tokens generate 
modulation parameters:
\[
(\gamma_{\texttt{START}},\beta_{\texttt{START}})
=
W_{\texttt{START}}T_q^{\texttt{START}},
\qquad
(\gamma_{\texttt{MID}},\beta_{\texttt{MID}})
=
W_{\texttt{MID}}T_q^{\texttt{MID}}.
\]

Then, the modulation is applied in two stages, corresponding to distinct conditioning roles:
\begin{itemize}
    \item \emph{Global conditioning:} For the query patch tensor
    \(
    H_q \in \mathbb{R}^{S_q \times P_q \times D},
    \)
    the \texttt{START} token performs global conditioning over all query patches:
    \(
    H_q
    \leftarrow
    H_q \odot (1+\gamma_{\texttt{START}})
    +
    \beta_{\texttt{START}}.
    \)
    \item \emph{Future conditioning:} The \texttt{MID} token performs future conditioning only over the forecast horizon. 
    Let $\mathcal{I}_{\mathrm{fut}}$ denote the future patch indices:
    \(
    H_q^{(\mathcal{I}_{\mathrm{fut}})}
    \leftarrow
    H_q^{(\mathcal{I}_{\mathrm{fut}})}
    \odot
    (1+\gamma_{\texttt{MID}})
    +
    \beta_{\texttt{MID}}.
    \)
\end{itemize}

\texttt{START} encodes global task-level information, while \texttt{MID} localizes conditioning to future patches.

\paragraph{Summary.}
Each encoder layer performs structured information routing across four levels: temporal dynamics within each component, target-covariate interactions within each block, semantic summarization through tokens, and task retrieval across examples. 
Stacking $L$ encoder layers refines both the example-derived task representation and the task-conditioned query patches used by the decoder.

\clearpage

\section{Training Details}
\label{app:training_details}

\begin{algorithm}[h]
\caption{Amortized Meta-Learning via Instruction-Conditioned Training}
\label{alg:amortized-meta-learning}
\begin{algorithmic}
\STATE {\bfseries Input:} Task distribution $\mathcal{T}$; example count $N$; model $f_\theta$; serialization function $\mathsf{Ser}(\cdot)$
\STATE {\bfseries Output:} Trained parameters $\theta$

\WHILE{not converged}
    \STATE Sample a meta-training task $\tau \sim \mathcal{T}$ (Section \ref{sec:task_classes}) (Table \ref{tab:implementation-mapping})
    
    \STATE Sample $N$ example demonstrations $\{\mathcal{E}_i\}_{i=1}^N$ from task domain $\tau$
    
    \STATE $\mathcal{E}_i = (X_i^{\text{hist}}, Z_i^{\text{hist}}, X_i^{\text{fut}}, Z_i^{\text{fut}})$
    
    \STATE Sample a query instance
    \STATE $\mathcal{Q} = (X_q^{\text{hist}}, Z_q^{\text{hist}}, Z_q^{\text{fut}})$
    \STATE Withhold query targets $Y_q^{\text{fut}}$
    
    \STATE Construct the in-context prompt:
    \STATE $\mathcal{P} = \mathsf{Ser}(\mathcal{E}_1) \oplus \cdots \oplus \mathsf{Ser}(\mathcal{E}_N) \oplus \mathsf{Ser}(\mathcal{Q})$
    
    \STATE Predict query output $\hat{Y}_q^{\text{fut}} = f_\theta(\mathcal{P})$
    
    \STATE Compute loss $\mathcal{L}(\hat{Y}_q^{\text{fut}}, Y_q^{\text{fut}})$
    
    \STATE Update parameters $\theta \leftarrow \theta - \eta \nabla_\theta \mathcal{L}$
\ENDWHILE

\end{algorithmic}
\end{algorithm}

We train iAmTime using episodic instruction-conditioned in-context
learning. Each training sample is a self-contained episode consisting of $K$
support demonstrations and one query. The support examples define the intended
input-output mapping, and the model must apply this mapping to the query without
parameter updates. Formally, for prompt $\mathcal{P}$, the
model learns
\(
f_\theta(\mathcal{P}) \mapsto \hat{Y}_q^{\mathrm{fut}}.
\)

A key design principle is that the same query input may correspond to different
targets under different support demonstrations. This prevents the model from
solving tasks using only query statistics or global task shortcuts, and instead
encourages genuine in-context adaptation.

\subsection{Heterogeneous Prompt Construction}
\label{sec:heterogenous_structures}

Training prompts are sampled from a distribution over heterogeneous time-series
structures and task families. Each prompt may vary in the number of target
dimensions $d_x$, the number of covariates $d_z$, and the semantic role of each
component, such as [\texttt{TARGET}], [\texttt{EXOG}], or
[\texttt{FUTURE\_EXOG}]. We include prompts with univariate targets
($d_x=1$), multivariate targets ($d_x>1$), past-only covariates,
future-known covariates, and no covariates ($d_z=0$). The support
examples may either match or differ from the query, depending on the task. This
exposes the model to varying degrees of structural alignment between
demonstrations and queries, improving robustness to real-world heterogeneity.

\subsection{Amortized Meta-Learning via Episodic Training}
\label{sec:amortized_meta_training}

The model is trained over a distribution of episodes rather than isolated
supervised examples. Each prompt $\mathcal{P}$ specifies both the data
distribution and the task semantics through support demonstrations. Training
minimizes
\[
\min_{\theta}
\mathbb{E}_{\mathcal{P}\sim p(\mathcal{P})}
\left[
\mathcal{L}\left(f_\theta(\mathcal{P}),Y_q^{\mathrm{fut}}\right)
\right],
\]
where $\mathcal{L}$ is the quantile regression loss defined in
Section~\ref{sec:training_objective}. We refer to this procedure as
\emph{amortized meta-learning}: adaptation is learned across a distribution of
tasks during training and executed at inference time through a single forward
pass conditioned on demonstrations.

Algorithm~\ref{alg:amortized-meta-learning} summarizes the episodic training
procedure. At each step, a task family is sampled, an episode-specific support
set and query are constructed, the prompt is serialized using the ICL input
format, and the model is optimized to predict the withheld query output. Because
task identity is not provided explicitly, the model must infer the relevant
mapping from the support examples.

\subsection{Meta-Training Task Classes}
\label{sec:task_classes}

\begin{table*}[t]
\caption{
Implementation mapping of example and query components for different meta-training
tasks. Each task input is serialized with the same 
example-query format, with task semantics differentiated by the example futures.
(All tasks include covariates 
$Z_i^{\text{hist}}, Z_i^{\text{fut}}, Z_q^{\text{hist}}, Z_q^{\text{fut}}$
unless specified.)
}
\label{tab:implementation-mapping}
\centering
\resizebox{\linewidth}{!}{
    \begin{tabular}{lcccccc}
        \toprule
        \textbf{Task} 
        & \textbf{Example History} 
        & \textbf{Example Future} 
        & \textbf{Query History} 
        & \textbf{Model Output} 
        & \textbf{Meta-Training} 
        & \textbf{Inference } \\
        & & & & & & \textbf{Adaptation} \\
        \midrule
        
        Forecasting
        & $X_{i,t:t+L}^{\text{hist}}$
        & $X_{i,t+L:t+L+H}^{\text{fut}}$
        & $X_q^{\text{hist}}$
        & $\hat{Y}_q^{\text{fut}}$ 
        & \checkmark & \checkmark \\ [8pt]
        
        Imputation / Recon.
        & $\check{X}_i^{\text{hist}}$
        & $X_i^{\text{hist}}$
        & $\check{X}_q^{\text{hist}}$
        & $\hat{X}_q^{\text{hist}}$  
        & \checkmark & \checkmark \\ [1pt]
        & (masked)
        & (clean)
        & (missing)
        & (reconstructed)
        & & \\ [8pt]
        

        Anomaly Detection
        & $X_{i,\text{corr}}^{\text{hist}}$
        & $M_i^{\text{hist}}$
        & $X_{q,\text{corr}}^{\text{hist}}$
        & $\hat{M}_q^{\text{hist}}$
        & \checkmark & \checkmark \\ [8pt]
        
        Classification$^{\dagger}$
        & $X_i^{\text{hist}}$
        & $c_i \cdot \mathbf{1}$
        & $X_q^{\text{hist}}$
        & $\hat{c}_q \cdot \mathbf{1}$  
        & \checkmark & \checkmark \\ [8pt]

        Source De-mixing$^{\bullet\dagger\ddagger}$
        & $\tilde{X}_i^{\text{hist}}=\sum_{m=0}^{M-1}s_i^{(m)}$
        & $s_i^{(c^\star)}$
        & $\tilde{X}_q^{\text{hist}}=\sum_{m=0}^{M-1}s_q^{(m)}$
        & $\hat{s}_q^{(c^\star)}$
        & \checkmark & \xmark \\ [8pt]

        \midrule
        \multicolumn{7}{c}
        {
        $^{\dagger}$ does not have $Z_i^{\text{fut}}, Z_q^{\text{fut}}$;
        $\quad$
        $^{\ddagger}$ does not have $Z_i^{\text{hist}}, Z_q^{\text{hist}}$;
        $\quad$
        $^{\bullet}$ $c^\star$ is the prompt-level target source index 
        (App.~\ref{sec:source_demixing_task}) shared by all examples and query
        }
        \\
        \bottomrule
    \end{tabular}
}
\end{table*}

We meta-train across supervised and self-supervised
instruction-conditioned tasks, including forecasting, imputation/reconstruction,
anomaly detection, property prediction/classification, and source de-mixing.
All tasks are expressed using the same prompt interface and target format: each
support example demonstrates an input-output mapping, and the query asks the
model to apply the same mapping to a held-out input. The task-specific
construction details are described below (and Table~\ref{tab:implementation-mapping}).

\subsubsection{Forecasting (analogous to next-token prediction).}
The forecasting task trains the model to predict future values given historical
context, mirroring next-token prediction in language models.

\emph{Data generation.}
Given a raw time series $X = \{x_t\}_{t=1}^{T}$, we sample a window of length $L$
and define:
\[
X^{\text{hist}} = X_{t:t+L}, \qquad
X^{\text{fut}} = X_{t+L:t+L+H},
\]
where $H$ is the forecast horizon.
If covariates $Z$ are available, we slice them analogously, including future
covariates $Z^{\text{fut}}$ when known.

\emph{ICL instantiation.}
Each example $\mathcal{E}_i$ provides a historical–future pair
$(X_i^{\text{hist}}, Z_i^{\text{hist}}, Z_i^{\text{fut}}) \mapsto X_i^{\text{fut}}$.
The query $\mathcal{Q}$ contains only $(X_q^{\text{hist}}, Z_q^{\text{hist}}, Z_q^{\text{fut}})$,
and the model predicts $Y_q^{\text{fut}} = X_q^{\text{fut}}$.
We use a variable support count $K \in \{0, \ldots, 4\}$ to provides robustness 
to varying shot counts at inference.

\subsubsection{Imputation / reconstruction (analogous to masked span prediction).}
\label{sec:imputation_reconstruction_task}
This task trains the model to reconstruct missing segments of a time series,
analogous to span corruption in encoder–decoder language models.

\emph{Data generation.}
Given a valid window $X$, we sample a binary mask $M \in \{0,1\}^T$ using random
patch masking, and construct a corrupted input:
\[
\tilde{X} = X \odot (1 - M).
\]
The mask $M$ is provided as an additional covariate to indicate missingness.

\emph{ICL instantiation.}
Examples demonstrate reconstruction by mapping corrupted inputs to clean outputs.
For the query, masked values are withheld and the model predicts
$Y_q^{\text{fut}} = X_q^{\text{hist}}$, corresponding to the full
reconstructed series.

\subsubsection{Anomaly detection (analogous to denoising).}
\label{sec:anomaly_detection_task}
This task trains the model to
receive an anomaly-injected series as history and output a binary indicator sequence as the future, where 1 indicates an anomaly and 0 indicates normal behavior.

\emph{Data generation.}
Starting from a clean series $X$, we sample a binary mask $M \in \{0,1\}^T$ and
inject synthetic anomalies such as spikes,
level shifts, or additive noise to obtain ($\epsilon$ is structured noise):
\[
X^{\text{corr}} = X + \epsilon \odot (M).
\]

\emph{ICL instantiation.}
Examples demonstrate the mapping from corrupted series to anomaly masks.
For the query, the anomaly mask is withheld, and the model predicts
it $Y_q^{\text{fut}} = M_q^{\text{hist}}$.
We inject anomalies of the chosen type to all examples and the query.
The episode-level anomaly type selection means the model must read support examples to understand which anomalous patterns are relevant — a level shift might be normal in one episode but anomalous in another.

\subsubsection{Property prediction / Classification (analogous to sequence classification).}
This task trains the model to identify global properties of a time series, such
as seasonality, trend, and regime.

\emph{Data generation.}
Labels are obtained from raw data via:
\begin{itemize}
    \item  KernelSynth generator \citet{ansari2024chronos} that introduce known properties (kernels),
    \item introducing Censor Augmentation and Spike Injection \citet{auer2025tirex} into series,
    \item statistical tests with feature-based heuristics (e.g., seasonality, stability, trend strength) using methods in \citet{yang2024comprehensive},
    \item clustering of time-series into classes using dataset-domain as seeds \citet{saha2022seeded}, and
    \item from UCR classification datasets with known 
    labels \citet{dau2019ucr, bagnall2018uea}.

\end{itemize}

Class labels are encoded as
constant target series:
$
X_i^{\text{fut}} = c_i \cdot \mathbf{1},
$
where $c_i$ is a class-specific scalar.

\emph{ICL instantiation.}
Since the model predicts time-series outputs, class labels are encoded as
constant target series:
$
Y^{\text{fut}} = c \cdot \mathbf{1},
$
where $c$ is a class-specific scalar. At least one example per class is included in the support set, and the query belongs to one of the classes represented in the examples.

Classification uses \emph{episode-local randomized label codes}: since the model applies InstanceNorm internally (subtracting loc, dividing by scale), the raw label codes are pre-compensated so that after normalization the model sees the intended code value:
\[
\tilde{c}_{i} = c_i \cdot \sigma_i + \mu_i
\]
where $(\mu_i, \sigma_i)$ are the InstanceNorm statistics of the respective history. 
Also, every episode draws a fresh permutation of codes-to-labels. This prevents the model from memorizing a fixed global mapping and forces it to read support examples to determine the coding scheme.

\subsubsection{Source de-mixing (analogous to unshuffling / separation).}
\label{sec:source_demixing_task}
This task trains the model to separate additive mixtures into latent source
components, analogous to unshuffling or source separation objectives in NLP and
vision. Unlike covariate-assisted de-mixing, the source to be extracted is not
provided as an exogenous variable; instead, it is specified through the support
examples in the ICL prompt.

\emph{Data generation.}
Each source de-mixing episode contains $K$ support examples and one query.
For a fixed episode, we sample $M$ latent source indices
$
m \in \{0,\dots,M-1\},
$
and assign each source index a concept type:
\[
\tau_m \in 
\{\texttt{sinusoid}, \texttt{trend}, \texttt{piecewise\_trend},
\texttt{spiky}, \texttt{autoregressive}\}.
\]
These concept assignments are held fixed across all support examples and the
query. We then sample a target source index
$
c^\star \sim \mathrm{Unif}\{0,\dots,M-1\},
$
which defines the component that the demonstrations ask the model to extract.

For each support example $i$, we independently generate $M$ source signals
$
s_i^{(0)}, \dots, s_i^{(M-1)} \in \mathbb{R}^{F},
$
where each $s_i^{(m)}$ is sampled according to its assigned concept type
$\tau_m$. The observed mixture is formed additively:
\(
\tilde{X}_i = \sum_{m=0}^{M-1} s_i^{(m)}.
\)
The latent target component is the selected source:
\(
X_i^{\text{fut}} = s_i^{(c^\star)}.
\)
No exogenous covariates are provided:
$
Z_i^{\text{hist}} = \emptyset
$
and
$
Z_i^{\text{fut}} = \emptyset.
$

\emph{ICL instantiation.}
Each support example $\mathcal{E}_i$ demonstrates how to extract the same latent
source type from a mixture:
\(
(X_i^{\text{hist}} = \tilde{X}_i,\;
Z_i^{\text{hist}} = \emptyset)
\;\mapsto\;
X_i^{\text{fut}} = s_i^{(c^\star)}.
\)
The query $\mathcal{Q}$ is generated using the same episode-level concept
assignments and the same target source index $c^\star$, but with independently
sampled source parameters:
\(
\tilde{X}_q = \sum_{m=0}^{M-1} s_q^{(m)}.
\)
The query contains only the mixed signal:
$
X_q^{\text{hist}} = \tilde{X}_q
$
with
$
Z_q^{\text{hist}} = \emptyset
$
and
$
Z_q^{\text{fut}} = \emptyset.
$

The model predicts
$
Y_q^{\text{fut}} = s_q^{(c^\star)},
$
corresponding to the same latent component type demonstrated in the support
examples. Since all examples in an episode share the same concept assignments
and target source index but use independently sampled parameters, the model must
infer from the demonstrations which component type to extract and apply that
extraction rule to the query mixture.

\subsection{Inference and Task Adaptation}
\label{sec:inference_adaptation}

At inference, the parameters are fixed. Given support
demonstrations $\mathcal{S}=\{\mathcal{E}_i\}_{i=1}^{K}$ and a query
$\mathcal{Q}$, the model predicts
\(
\hat{Y}_q^{\mathrm{fut}}
=
f_\theta(\mathcal{P}),
\quad
\mathcal{P}
=
\mathsf{Ser}(\mathcal{E}_1)\oplus\cdots\oplus
\mathsf{Ser}(\mathcal{E}_K)\oplus\mathsf{Ser}(\mathcal{Q}).
\)
Support implicitly specify the task to be performed. For
forecasting, the output is the future continuation of the target series; for
classification, it is a constant sequence encoding the inferred class; for
anomaly detection, it is an anomaly mask; and for imputation or reconstruction,
it is the recovered signal. 

\paragraph{Scope of inference tasks.}
We implement forecasting, classification, imputation, reconstruction, and anomaly detection at inference time.
Other tasks, such as source de-mixing,
primarily serve as self-supervised objectives during training.
While these tasks improve representation quality and in-context learning
capability, they are not directly evaluated at inference time due to the
difficulty of specifying them solely through demonstrations in practical
settings.

\subsection{Curriculum Learning}
\label{sec:curriculum}

The episodic training distribution combines tasks with different
levels of difficulty. Query-only forecasting provides a stable supervised signal,
whereas support-conditioned forecasting, classification, anomaly detection,
imputation, and source separation require the model to infer task from
demonstrations. To stabilize optimization and reduce shortcut learning, we adopt
a progressive curriculum
\citep{elman1993learning,sanger2002neural,bengio2009curriculum,wu2020curricula}.
The model is trained in four phases, each initialized from the best checkpoint of
the previous phase. Table~\ref{tab:curriculum_task_mixture} summarizes the task
mixture across phases.
The mixture weights are determined based on preliminary experiments to balance learning progress and task diversity, and are not heavily tuned. 

\begin{table}[t]
    \caption{Curriculum task mixture. Percentages denote sampling probabilities within each phase and are normalized during training.}
    \centering
    \small
    \setlength{\tabcolsep}{6pt}
    \renewcommand{\arraystretch}{1.12}
    \begin{tabular}{lcccc}
    \toprule
    \textbf{Task family} & \textbf{A} & \textbf{B} & \textbf{C} & \textbf{D} \\
    \midrule
    Query-only forecasting        & 60\% & 30\% & 15\% & 20\% \\
    Support forecasting           & 40\% & 30\% & --   & 15\% \\
    Support + transforms          & --   & 40\% & 20\% & 15\% \\
    Classification                & --   & --   & 15\% & 12\% \\
    Synthetic classification      & --   & --   & 10\% & 8\%  \\
    Anomaly detection             & --   & --   & 10\% & 10\% \\
    Imputation / reconstruction   & --   & --   & 10\% & 10\% \\
    Source separation             & --   & --   & 5\%  & 5\%  \\
    Cross-task ambiguity          & --   & --   & optional & 5\% \\
    \bottomrule
    \end{tabular}
    \label{tab:curriculum_task_mixture}
\end{table}

\paragraph{Phase A: forecasting backbone stabilization.}
The first phase ensures that the episodic architecture preserves strong
forecasting behavior. Training is dominated by query-only forecasting, with a
smaller fraction of support-conditioned forecasting using windows from
the same series or dataset.

\paragraph{Phase B: forecasting ICL specialization.}
The second phase increases support dependence by introducing episode-specific
forecasting transforms. Each episode samples one or more transforms, such as
affine scaling, additive trend, seasonal injection, piecewise scaling, power
distortion, warping:
\[
\tilde{\mathbf{x}}
=
T_1 \circ T_2 \circ \cdots \circ T_k(\mathbf{x}),
\qquad
k \in \{1,2,3\}.
\]
Because the active transformation varies across episodes, support examples
become informative about the mapping that should be applied to the query.

\paragraph{Phase C: multitask episodic training.}
The third phase introduces the full set of instruction-conditioned tasks into
the same training loop, including forecasting, classification, anomaly detection,
imputation/reconstruction, and source separation. All tasks share the same prompt 
and output format.

\paragraph{Phase D: mixed training with anti-shortcut episodes.}
The final phase mixes all task families and adds cross-task ambiguity episodes,
where the same query window can require different outputs depending on the
support demonstrations. This 
encourages demonstration-conditioned task inference.

This curriculum aligns the difficulty of the training distribution with the
model's evolving representational capacity and has been shown to benefit
ICL in Transformers \citep{garg2022can}.

\subsection{Hyperparameters and Training Configuration}
\label{app:hyperparameters}

Tables~\ref{tab:curriculum_hparams} and~\ref{tab:model_hparams} summarize
the training and architectural settings. Pretraining used eight A100
40\,GB GPUs for 500K steps and 90 hours (720 GPU-hours), with batch size
256, BF16, and gradient checkpointing. Phases A-D used
72/144/252/252 GPU-hours, respectively. All ablations and retrainings
required approximately 12,960 GPU-hours; offline data-generation cost is
excluded. Evaluation used one A100 GPU at approximately 400 query series
per second and four demonstration-selection seeds where applicable.

With a patch size $p=16$ and a query horizon $H$, this yields $\lceil H/16\rceil$ future patches. The decoder maps each future-patch embedding to $16\times9$ outputs (16 patches at nine quantiles). The model generates up to 64 patches ($1{,}024$ time-steps) in parallel. 
Longer horizons use blockwise autoregression.

\begin{table}[t]
    \caption{
        Training and model hyperparameters.
        (a) Optimization hyperparameters for curriculum phases.
        (b) Main architectural hyperparameters for \textsc{iAmTime}.
    }
    \centering
    \small
    \setlength{\tabcolsep}{3pt}
    \renewcommand{\arraystretch}{1.12}

    \subcaptionbox{
    \label{tab:curriculum_hparams}
    }[0.60\linewidth]{
        \centering
        \begin{tabular}{lcccc}
        \toprule
        \textbf{Setting} & \textbf{A} & \textbf{B} & \textbf{C} & \textbf{D} \\
        \midrule
        Steps & 50k & 100k & 175k & 175k \\
        Budget & 10\% & 20\% & 35\% & 35\% \\
        LR & $1.0{\times}10^{-4}$ & $8.0{\times}10^{-5}$ & $5.0{\times}10^{-5}$ & $3.0{\times}10^{-5}$ \\
        \midrule
        Batch & \multicolumn{4}{c}{256} \\
        Optimizer & \multicolumn{4}{c}{AdamW} \\
        Weight decay & \multicolumn{4}{c}{0.01} \\
        Schedule & \multicolumn{4}{c}{Cosine, 5\% warm-up} \\
        Precision & \multicolumn{4}{c}{BF16} \\
        Grad. ckpt. & \multicolumn{4}{c}{Enabled} \\
        \bottomrule
        \end{tabular}
    }
    \hfill
    \subcaptionbox{
    \label{tab:model_hparams}
    }[0.36\linewidth]{
        \centering
        \begin{tabular}{lc}
        \toprule
        \textbf{Hyperparameter} & \textbf{Value} \\
        \midrule
        Input/Output patch & 16 \\
        Max output steps & 64 \\
        Hidden dim. $D$ & 768 \\
        Encoder layers & 12 \\
        MoE experts & 4 \\
        Quantile levels & 9 \\
        Max query context & 4096 \\
        Max example context & 1024 \\
        \midrule
        Total model params & 300M \\
        \bottomrule
        \end{tabular}
    }
    \label{tab:training_model_hparams}
\end{table}

\subsection{Anti-Shortcut Mechanisms}
\label{sec:anti_shortcut_training}

A central challenge in multi-task episodic training is preventing the model from
using shortcuts that bypass support examples. We therefore introduce several
anti-shortcut mechanisms.

\paragraph{Episode-local label remapping.}
For classification tasks, class labels are not assigned fixed global scalar
codes. Instead, each episode samples a random permutation of evenly spaced label
codes:
\[
\mathrm{codes} =
\mathrm{Permute}\left(\mathrm{linspace}(1,9,|\mathcal{C}|)\right),
\]
where $\mathcal{C}$ is the set of classes in the episode. This preserves
separation between class codes while forcing the model to infer the class-code
mapping from support examples.

\paragraph{Support-dependent ambiguity.}
For support-conditioned forecasting, we introduce episode-specific transforms
that are applied consistently within an episode but vary across episodes. These
transforms include affine scaling, additive trends, seasonal injection,
piecewise scaling, power distortion, and time warping. For an episode with
$k$ transforms,
\[
\tilde{\mathbf{x}}
=
T_1\circ T_2 \circ \cdots \circ T_k(\mathbf{x}),
\qquad
k\in\{1,2,3\}.
\]
Since the active transform is demonstrated only in the support examples, the
query alone is insufficient to determine the correct output.

\paragraph{Cross-task ambiguity.}
We additionally use a cross-task ambiguity sampler in which the same query
window can appear under multiple task semantics, such as forecasting, anomaly
detection, imputation, or classification. Each task builder constructs its own
support set while keeping the query window fixed. This forces the model to use
the support demonstrations to infer whether the desired output is, for example,
a forecast, an anomaly mask, a reconstruction, or a class label.

\clearpage

\section{Training Data Augmentation and Synthetic Construction}
\label{sec:augmentation_construction}
To increase diversity and induce controllable relationships, we augment the base
corpora using three complementary strategies.

\subsection{Time-series mixup augmentation (TSMixup)}
\label{sec:tsmixup}
First, we apply the TSMixup \citet{ansari2024chronos} procedure introduced in Chronos, which generates
synthetic mixtures of time series to increase diversity and induce compositional structure.
For each augmented sample, we first draw a random integer
$
k \sim \mathcal{U}\{1, K\},
$
and a segment length
$
\ell \sim \mathcal{U}\{\ell_{\min}, \ell_{\max}\}.
$
We then sample $k$ univariate time-series segments.
To ensure compatibility across different magnitudes and scales, each segment is
normalized using z-score scaling.
Mixing weights
$
\lambda
$
are sampled from a symmetric Dirichlet distribution,
$
\lambda \sim \text{Dir}(\alpha),
$
and the augmented series is formed as a convex combination:
\[
x^{\text{TSMixup}}_{1:\ell}
= \sum_{i=1}^{k} \lambda_i \tilde{x}^{(i)}_{1:\ell}.
\]

In our implementation, we set $K=3$, $\ell_{\min}=128$, $\ell_{\max}=2048$, and
$\alpha=1.5$, and generate approximately 30 million augmented univariate time
series.
Unlike prior usage focused primarily on forecasting, these mixed series later
serve as inputs for multiple task classes, including source de-mixing and
anomaly correction.

\subsection{Kernel-based synthetic generation (KernelSynth)}
\label{sec:kernelsynth}
Second, we synthetic time series generated using the
KernelSynth \citet{ansari2024chronos} procedure from Chronos, 
to further enrich the training distribution with controlled temporal structures. It is a Gaussian-process-based synthetic data generator which constructs a 
covariance function by randomly composing kernels from a
bank $\mathcal{K}$, including Linear (trend), RBF (smooth local variation),
Periodic (seasonality), Rational Quadratic (multi-scale variation), White Noise,
and Constant kernels.
For each synthetic instance, we sample a number of basis kernels
and compose them using random binary operations
$
op \in \{+, \times\},
$
where addition corresponds to superposition of independent processes and
multiplication induces interactions (e.g., modulating seasonality by trend).
This yields a composite kernel function $\kappa(t, t')$.
A synthetic time series is then sampled from a
Gaussian process:
\[
x \sim \mathcal{GP}(0, \kappa(t, t'))
\]
Using this procedure, we generate approximately 10 million synthetic univariate
time series with diverse spectral and structural properties.

\subsection{Multivariate construction with covariate relations}
Third, leveraging the augmented univariate corpus, we construct multivariate
time series with endogenous and exogenous covariate relationships using a
custom multivariate generation pipeline.
This yields approximately 50 million multivariate series with varying numbers
of targets and covariates.

To generate multivariate time series with structured endogenous and exogenous
dependencies, we introduce a multivariate construction pipeline that imposes
explicit mathematical relationships between independently sampled univariate
series.
The goal is to synthesize entangled systems in which target components
(\emph{endogenous}) depend on auxiliary drivers (\emph{exogenous}) through a
diverse family of causal, nonlinear, and temporally lagged transformations.

\paragraph{Initialization and normalization.}
We begin by sampling a set of $N$ independent univariate time-series segments
\[
\{x^{(1)}, \dots, x^{(N)}\}, \qquad x^{(i)} \in \mathbb{R}^{L},
\]
from a source pool (e.g., real data, TSMixup, or KernelSynth).
The segment length $L$ is sampled randomly, and each series is sliced using a
cyclic iterator to ensure uniform coverage.

To enable stable mathematical composition across heterogeneous magnitudes, each
series is normalized by its mean absolute value:
\[
\tilde{x}^{(i)}_t = \frac{x^{(i)}_t}{\frac{1}{L}\sum_{s=1}^{L} |x^{(i)}_s|}.
\]

\paragraph{Role assignment.}
The normalized series are partitioned into:
\begin{itemize}
    \item endogenous (target) subsets. 
    We enforce that at least $60\%$ of the series are assigned as endogenous (internal factors).
    \item exogenous (covariate) subsets. The remaining series are assigned as exogenous (external factors).
\end{itemize}

Exogenous series influence endogenous series but are never influenced by them. Endogenous series can influence each other. This mirrors real-world covariate relationships where external factors (e.g., weather, holidays) affect the target but not vice versa.
The 5 time-dependent transformations described below interact with the endogenous/exogenous structure in two distinct ways based on their nature:

\begin{itemize}
    \item  "Mixing" transformations (Linear Combination, Nonlinear Modulation):
    These replace or blend a target series using the base series' values, creating direct mathematical dependencies between series.
    The base series is always excluded from its own target list (to avoid blending a series with itself).
    When exogenous is the base, targets are endogenous only, and 
    when endogenous is the base, targets are other endogenous (peer influence).

    \item "Modifying" transformations (Trend Modification, Seasonality Injection, Shock Injection):
    These additively perturb existing series rather than replacing them — they shift trends, inject seasonal patterns, or add shocks on top of current values. Here,
    the base series is included in the target list.
    When exogenous is the base, targets are all endogenous and itself.
    When endogenous is the base, targets are all endogenous including itself.
\end{itemize}

\paragraph{Time-dependent transformations.}
We introduce time-dependent dependencies by modifying target series over
aligned time intervals $[t_{\text{start}}, t_{\text{end}}]$ using a selected
base series $x_{\text{base}}$, which may be endogenous or exogenous.
For a target series $y$, we apply one or more of the following transformations:

\begin{itemize}

    \item \emph{Linear combination.}
    \[
    y_t \leftarrow \alpha x_{\text{base},t} + \beta y_t + \epsilon_t,
    \qquad
    \epsilon_t \sim \mathcal{N}(0, \sigma^2),
    \]
    where we use
    $\alpha \sim \mathcal{U}(0.3,0.6)$, $\beta \sim \mathcal{U}(0.3,0.7)$, and
    $\sigma$ is set to $10\%$ of the standard deviation of the base segment.

    \item \emph{Nonlinear modulation.}
    \[
    y_t \leftarrow y_t + \gamma f(x_{\text{base},t}),
    \]
    where $\gamma$ is a scaling coefficient and $f(\cdot)$ is sampled from a library
    of nonlinear functions, including logarithmic, exponential, power-law, and
    hyperbolic tangent transformations.

    \item \emph{Trend modification.}
    We estimate the linear trend of $y$ as $m_{\text{old}} t + c$ and replace it with
    a modified trend:
    \[
    y_t^{\text{new}} = (y_t - m_{\text{old}} t - c) + m_{\text{new}} t,
    \]
    where $m_{\text{new}}$ is obtained by increasing, decreasing, or reversing the
    original slope.

    \item \emph{Seasonality injection.}
    A sinusoidal component is injected with amplitude modulated by the base series:
    \[
    S_t = A \sin\!\left(\frac{2\pi t}{P}\right)
    \left(1 + \alpha\cdot\text{norm}(x_{\text{base},t})\right),
    \qquad
    y_t \leftarrow y_t + S_t.
    \]
    We use $\alpha = 0.3.$

    \item \emph{Shock injection.}
    Discrete shocks are added synchronously across target series:
    \[
    y_t \leftarrow y_t + s \cdot M \cdot d(t),
    \]
    where $s \in \{\pm1\}$, $M$ is the shock magnitude, and $d(t)$ is a decay function
    (constant, linear, or exponential).

\end{itemize}

\paragraph{Time-lagged transformations.}
To induce temporal dependencies, we introduce lagged relationships between a
leader series $x_{\text{leader}}$ and a follower series $y_{\text{follower}}$.

\begin{itemize}
    \item \emph{Lagged influence.}
    \(
    y_{\text{follower},t} \leftarrow
    y_{\text{follower},t} + \alpha x_{\text{leader},t-\ell}.
    \)

    \item \emph{Cointegration (error correction).}
    \(
    \epsilon_t = y_{\text{follower},t} - x_{\text{leader},t}; \;\;
    y_{\text{follower},t} \leftarrow
    y_{\text{follower},t} - \lambda \epsilon_{t-\ell}.
    \)

    \item \emph{Granger-style influence.}
    \(
    y_{\text{follower},t} \leftarrow
    y_{\text{follower},t}
    + \sum_{k=1}^{K} \alpha e^{-\beta k} x_{\text{leader},t-k}.
    \)
\end{itemize}
\paragraph{Stabilization and output.}
To prevent numerical instability, all series are clipped to lie within $\pm 5$
standard deviations of their empirical mean.
The final output consists of a multivariate target set $X$, a covariate set $Z$,
and metadata describing the induced dependencies.
These constructed systems serve as inputs for downstream instruction-conditioned
meta-training tasks.

\subsection{Source separation episode synthesis}
Fourth, we construct synthetic source-separation episodes to directly train the
model on example-conditioned component extraction. This task is designed as a
strong test of instruction-conditioned adaptation: the model observes additive
mixtures as inputs and must infer, from support demonstrations alone, which
latent component should be extracted. This resembles classical source separation
settings \citep{hyvarinen2000independent,comon1994independent}, but is
formulated as an in-context learning problem rather than as a fixed
decomposition objective.

Each source-separation episode contains $K$ support examples and one query.
For a fixed episode, we sample $M$ source indices
$
m \in \{0,\dots,M-1\}
$
and assign each source a concept type from a finite concept bank:
\[
\tau_m \in 
\{\texttt{sinusoid}, \texttt{trend}, \texttt{piecewise\_trend},
\texttt{spiky}, \texttt{autoregressive}\}.
\]
The assignments $\{\tau_m\}_{m=0}^{M-1}$ are fixed within the episode, so that
source $m$ always corresponds to the same conceptual component. We then sample a
target source index
$
c^\star \sim \mathrm{Unif}\{0,\dots,M-1\},
$
which defines the component demonstrated by the support examples and extracted
from the query.

For each support example $i$, we sample parameters from episode-level ranges and
generate $M$ source signals
$
s_i^{(0)},\dots,s_i^{(M-1)}.
$
The observed mixture is formed additively:
\[
\tilde{x}_i(t) = \sum_{m=0}^{M-1} s_i^{(m)}(t),
\]
and the target component is
$
s_i^{(c^\star)}.
$
The component generators are:
\[
\begin{aligned}
\text{Sinusoid:}\quad
&s(t) = a\sin(2\pi f t+\varphi)+\epsilon_t,\\
\text{Trend:}\quad
&s(t) = bt+c+\epsilon_t,\\
\text{Piecewise trend:}\quad
&s(t) =
b_1t\mathbf{1}[t\leq \tau]
+
\big(b_1\tau+b_2(t-\tau)\big)\mathbf{1}[t>\tau]
+\epsilon_t,\\
\text{Spiky:}\quad
&s(t)=\sum_{j=1}^{J}h_j
\exp\!\left(-\frac{(t-\tau_j)^2}{2w_j^2}\right)+\epsilon_t,\\
\text{Autoregressive:}\quad
&s(t)=\alpha s(t-1)+\epsilon_t.
\end{aligned}
\]
Parameters such as amplitude, frequency, phase, trend slope, spike location, and
autoregressive coefficient are sampled independently per source within the
episode-level parameter ranges. This creates support diversity while preserving
the same latent extraction rule across the episode.

We instantiate two modes. In \emph{future-component} mode, the mixture history
has length $H$ and the model predicts the selected component over the future
horizon:
\[
X_i^{\mathrm{hist}}=\tilde{x}_i[0:H],
\qquad
X_i^{\mathrm{fut}}=s_i^{(c^\star)}[H:H+F].
\]
In \emph{reconstruction-component} mode, the model extracts the selected
component over the same window:
\[
X_i^{\mathrm{hist}}=\tilde{x}_i[0:F],
\qquad
X_i^{\mathrm{fut}}=s_i^{(c^\star)}[0:F].
\]
The query is generated using fresh source parameters sampled from the same
episode-level ranges, with the same concept assignments and the same target
index $c^\star$. No exogenous channels are provided. Thus, the support examples
act as instructions specifying which latent source concept should be extracted
from the query mixture.

\subsection{Synthetic classification episode synthesis}
\label{sec:synthetic_classification_data}
Finally, we generate synthetic classification episodes to avoid limiting
classification training to the size and label coverage of real-world labeled
datasets. The goal is to expose the model to controlled decision boundaries over
time-series structure while preserving the same instruction-conditioned
example-query format used for forecasting and other tasks. This complements
standard time-series classification benchmarks and models \citep{fawaz2020deep}
by producing arbitrarily many labeled episodes with known generative factors.

Each synthetic classification episode first samples a family group, and the
classes within the episode are defined by discriminative properties of that
family. We use three groups:
\[
\begin{array}{lll}
\text{Waveform} &:
& \{\texttt{low freq},\texttt{mid freq},\texttt{high freq}\},\\
\text{Regime} &:
& \{\texttt{trending},\texttt{mean reverting},\texttt{volatile}\},\\
\text{Motif} &:
& \{\texttt{motif present},\texttt{motif absent}\}.
\end{array}
\]
The waveform family is generated from sinusoidal processes with class-specific
frequency bands. The regime family is generated from stochastic processes with
different global dynamics, such as trend-dominated, mean-reverting, or
high-variance behavior. The motif family is generated by inserting or omitting a
deterministic temporal pattern inside a random-walk background.

For each episode, class labels are assigned randomized episode-local codes, using
the same scalar code representation as real-data classification. Specifically,
if a sample belongs to class $c$, its output target is encoded as a constant
series:
\(
Y^{\mathrm{fut}} = \rho(c)\cdot \mathbf{1},
\)
where $\rho(c)$ is the episode-local scalar code assigned to class $c$. Support
examples demonstrate the mapping from time-series inputs to these codes, and the
query must predict the code corresponding to its class.

This synthetic sampler allows us to generate classification tasks in which the
discriminative feature is known by construction, but the label identity is
episode-dependent. Consequently, the model cannot rely on a fixed global class
index; instead, it must infer the label semantics from the support examples and
apply the induced decision rule to the query.

\clearpage
\section{Extended Forecasting Evaluations}
\label{sec:extend_evaluations_appendix}

This section presents additional experimental results complementing Section \ref{sec:results}. 

\begin{figure}[h]
\centering
\begin{subfigure}{.48\textwidth}
  \centering
  \resizebox{.99\linewidth}{!}{\input{images/gift_eval_overall_rank.pgf}}
  \caption{Overall CRPS and MASE scores on GIFT-Eval.
  Lower values are better. ``Zero-shot Models'' are not trained on this data.
  }
  \label{fig:overall_gift_eval_rank}
\end{subfigure}\hfill
\begin{subfigure}{.48\textwidth}
  \centering
  \resizebox{.99\linewidth}{!}{\input{images/gift_long_agg.pgf}}
  \caption{Aggregated CRPS-Rank and MASE-Rank on long-term GIFT-Eval results.
  Lower values are better. ``Zero-shot Models'' are not trained on this data.}
  \label{fig:gift_long_agg}
\end{subfigure}
\caption{Overall and long-term performance on the GIFT-Eval benchmark.}
\label{fig:gift_eval_1}
\end{figure}

\begin{figure}[h]
\centering
\begin{subfigure}{.48\textwidth}
  \centering
  \resizebox{.85\linewidth}{!}{\input{images/gift_medium_agg.pgf}}
  \caption{Aggregated CRPS and MASE scores on medium-term GIFT-Eval results.
  Lower values are better. ``Zero-shot Models'' are not trained on this data.}
  \label{fig:gift_mid_agg}
\end{subfigure}\hfill
\begin{subfigure}{.48\textwidth}
  \centering
  \resizebox{.85\linewidth}{!}{\input{images/gift_short_agg.pgf}}
  \caption{Aggregated CRPS and MASE scores on short-term GIFT-Eval results.
  Lower values are better. ``Zero-shot Models'' are not trained on this data.}
  \label{fig:gift_short_agg}
\end{subfigure}
\caption{Term length performance on the GIFT-Eval benchmark.}
\label{fig:gift_eval_2}
\end{figure}

\subsection{Inference Protocol for Forecasting Evaluation}
\label{sec:eval_inference_protocol}
For forecasting evaluation, we use the same support-query prompt interface as during training. For each evaluation episode, a test time-series from the benchmark is selected as the \textit{query} input. The \textit{demonstrations} are constructed by sampling historical windows of length $[64 ... 1024] + \texttt{prediction\_length}$ which is split into the example's history and future. There is no overlap or leakage between the query and the demonstrations.
A set of $K=4$ examples are constructed by default, unless stated otherwise in the corresponding experiment. The query is capped at 4096 time-steps.
The per-dataset metrics for baselines are obtained from their corresponding leaderboards and verified by re-evaluation when possible.

\subsection{Zero-shot generalization on GIFT-Eval}

First, we discuss the overall CRPS and MASE rankings in the zero-shot evaluation on GIFT-Eval. 
This rank-based metric is helpful as it averages rank across evaluation settings,
ensuring robustness of the metrics against outlier performance.
Followed by this, we show the short- , medium- , long-term, 
and univariate-multivariate forecasts 
from GIFT-Eval.

iAmTime achieves the strongest overall performance in the zero-shot setting on GIFT-Eval. It also attains competitive aggregated CRPS and MASE ranks across all evaluation slices (Figure \ref{fig:overall_gift_eval_rank}). 
Despite no training on the benchmark data, iAmTime consistently outperforms or matches task-specific and locally trained models, including methods with partial train–evaluation overlap. 
This advantage persists across long-, medium-, and short-term horizons 
(Figures \ref{fig:gift_long_agg},\ref{fig:gift_mid_agg},\ref{fig:gift_short_agg}).
Performance on different frequency inputs as shown in
Figure \ref{fig:gift_eval_3} suggests that iAmTime maintains strong performance across most frequencies and staying in top 4 for the others.
This demonstrates robust zero-shot generalization across diverse forecasting conditions.

Performance is stable across evaluation slices and horizon length.
It is also stable across input structure, and to make this point more explicit, Table \ref{tab:gift_eval_multivariate_wql} isolates the performance on only the multivariate subset.
In majority of the cases iAmTime achieves the top rank.
This indicates robustness to both forecasting horizon and input dimensionality. 
In contrast, classical statistical baselines and fully local models generally underperform, especially on longer horizons, while other foundation models show competitive but less consistent rankings across settings.
Table \ref{tab:gift_eval_univariate_wql} also demonstrates the performance on only the univariate subset, where iAmTime also achieves the top rank in most cases.

\subsection{Win Rate and Confidence Intervals on fev-bench}

\begin{figure}[t]
    \centering
    \includegraphics[width=\linewidth]{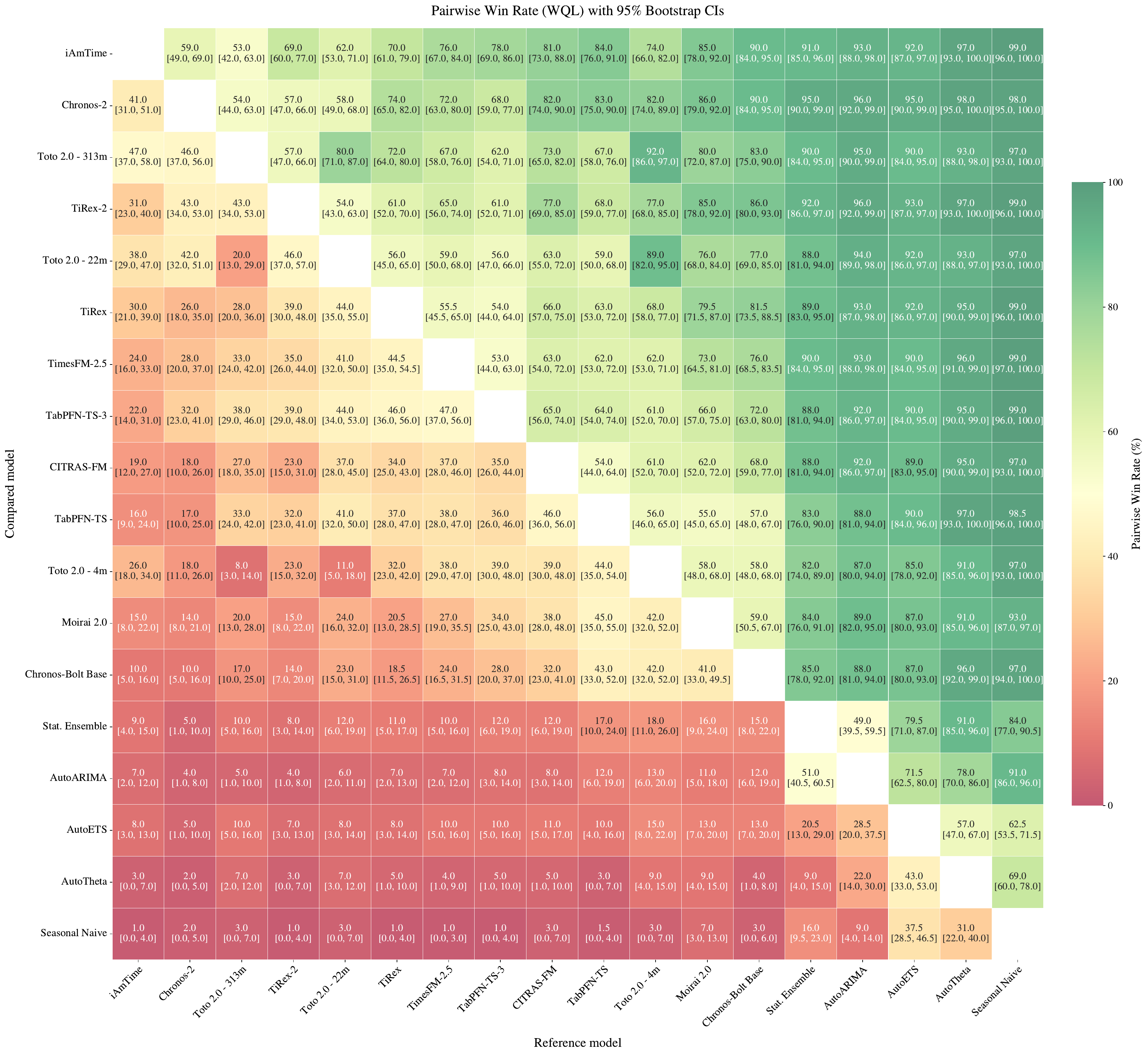}
    \caption{
        The pairwise win rates for all models on fev-bench with 
        95\% 
        confidence intervals (CIs) with respect to WQL metric.
    }
    \label{fig:confidence_intvl_win_rate_fev}
\end{figure}

Figure~\ref{fig:confidence_intvl_win_rate_fev} shows that iAmTime achieves the strongest overall pairwise win-rate profile on fev-bench. It performs comparably to Chronos-2 and Toto 2.0–313m, while its bootstrapped 95\% confidence intervals indicate consistent advantages over the remaining foundation-model and statistical baselines.

\begin{figure}[!t]
\centering

\resizebox{0.96\linewidth}{!}{\input{images/gift_freq_overall_agg.pgf}}

\caption{Result on different frequency inputs on the GIFT-Eval benchmark. 
Here 
SN:Seasonal Naive, 
Chr2:Chronos-2, 
TFM2.5:TimesFM-2.5, 
CBB:Chronos-Bolt Base, 
TiR:TiRex, 
Toto-1:Toto 1.0, 
T2:Toto 2.0, 
Moi2:Moirai 2.0,
TPFN:TabPFN-TS, 
AAR:Auto ARIMA, 
AETS:Auto ETS, 
AT:Auto Theta,
DLin:DLinear, 
}
\label{fig:gift_eval_3}
\end{figure}

\begin{sidewaystable}[p]
\centering

\caption{
CRPS scores of iAmTime compared with various baseline models on the
multivariate data subset of the GIFT-Eval benchmark. Models achieving
the best and second-best scores are highlighted.
}
\label{tab:gift_eval_multivariate_wql}

\begin{adjustbox}{
    max width=0.90\textheight,
    max totalheight=0.60\textwidth,
    center
}
\begin{tabular}{lllllllllllllll}
\toprule

Dataset
& \makecell[c]{iAmTime}
& \makecell[c]{TiRex-2}
& \makecell[c]{Toto 2.0\\2.5B}
& \makecell[c]{Chronos-2}
& \makecell[c]{Toto 2.0\\1B}
& \makecell[c]{TiRex}
& \makecell[c]{TimesFM-2.5}
& \makecell[c]{Toto 1.0}
& \makecell[c]{Toto 2.0\\22M}
& \makecell[c]{Toto 2.0\\313M}
& \makecell[c]{Moirai 2.0}
& \makecell[c]{Chronos\\Bolt Base}
& \makecell[c]{TabPFN-TS}
& \makecell[c]{Seasonal\\Naive} \\

\midrule

bitbrains\_fast\_storage/5T/long & 0.654 & 0.688 & 0.591 & 0.703 & \textbf{0.556} & 0.672 & 0.831 & 0.669 & \underline{0.578} & 0.590 & 0.807 & 0.748 & 0.885 & 1.177 \\
bitbrains\_fast\_storage/5T/medium & 0.590 & 0.596 & 0.582 & 0.623 & \textbf{0.537} & 0.638 & 0.766 & 0.629 & \underline{0.567} & 0.576 & 0.694 & 0.755 & 0.949 & 1.198 \\
bitbrains\_fast\_storage/5T/short & 0.402 & 0.384 & 0.340 & 0.391 & \textbf{0.338} & 0.380 & 0.397 & 0.371 & 0.351 & \underline{0.339} & 0.427 & 0.454 & 0.662 & 1.210 \\
bitbrains\_fast\_storage/H/short & 0.785 & 0.668 & \textbf{0.556} & 0.664 & \underline{0.557} & 0.700 & 0.782 & 0.623 & 0.574 & 0.588 & 0.615 & 0.774 & 0.670 & 1.022 \\
bitbrains\_rnd/5T/long & \underline{0.552} & 0.595 & 0.587 & 0.873 & 0.558 & 0.632 & 0.743 & 0.589 & 0.593 & \textbf{0.536} & 0.570 & 0.756 & 0.819 & 1.175 \\
bitbrains\_rnd/5T/medium & \textbf{0.546} & 0.625 & 0.619 & 1.005 & 0.614 & 0.604 & 0.846 & 0.628 & 0.562 & \underline{0.555} & 0.596 & 0.605 & 0.819 & 1.169 \\
bitbrains\_rnd/5T/short & 0.384 & 0.398 & 0.381 & 0.416 & \underline{0.370} & 0.404 & 0.408 & 0.399 & 0.389 & \textbf{0.369} & 0.404 & 0.438 & 0.608 & 1.102 \\
bitbrains\_rnd/H/short & 0.656 & 0.598 & 0.623 & 0.804 & 0.608 & 0.611 & 0.610 & 0.593 & \textbf{0.575} & \underline{0.590} & 0.670 & 0.624 & 0.742 & 1.243 \\
bizitobs\_application/10S/long & 0.050 & 0.052 & 0.056 & \textbf{0.045} & 0.056 & 0.052 & 0.053 & 0.053 & 0.054 & 0.055 & 0.056 & 0.109 & 0.049 & \underline{0.046} \\
bizitobs\_application/10S/medium & 0.033 & 0.036 & 0.040 & \textbf{0.026} & 0.040 & 0.038 & \underline{0.033} & 0.034 & 0.040 & 0.041 & 0.037 & 0.104 & 0.041 & 0.043 \\
bizitobs\_application/10S/short & 0.011 & 0.011 & 0.010 & \underline{0.010} & 0.011 & 0.011 & \textbf{0.009} & 0.012 & 0.013 & 0.011 & 0.013 & 0.054 & 0.015 & 0.035 \\
bizitobs\_l2c/5T/long & \textbf{0.221} & 0.520 & 0.352 & 0.298 & 0.354 & \underline{0.269} & 0.279 & 0.533 & 0.457 & 0.371 & 0.300 & 0.738 & 0.306 & 0.648 \\
bizitobs\_l2c/5T/medium & \textbf{0.203} & 0.325 & 0.243 & 0.247 & 0.253 & 0.251 & \underline{0.241} & 0.316 & 0.282 & 0.272 & 0.261 & 0.445 & 0.261 & 0.520 \\
bizitobs\_l2c/5T/short & 0.071 & 0.072 & \textbf{0.051} & 0.069 & \underline{0.055} & 0.076 & 0.072 & 0.069 & 0.068 & 0.059 & 0.084 & 0.074 & 0.084 & 0.262 \\
bizitobs\_l2c/H/long & \textbf{0.245} & 0.267 & 0.255 & 0.267 & \underline{0.254} & 0.268 & 0.273 & 0.369 & 0.266 & 0.262 & 0.321 & 0.278 & 0.292 & 0.941 \\
bizitobs\_l2c/H/medium & \textbf{0.226} & 0.257 & 0.235 & 0.236 & \underline{0.233} & 0.252 & 0.237 & 0.356 & 0.238 & 0.233 & 0.274 & 0.254 & 0.237 & 0.904 \\
bizitobs\_l2c/H/short & 0.189 & 0.204 & \textbf{0.174} & \underline{0.176} & 0.187 & 0.212 & 0.179 & 0.199 & 0.183 & 0.185 & 0.235 & 0.189 & 0.210 & 0.521 \\
bizitobs\_service/10S/long & \underline{0.050} & 0.051 & 0.053 & 0.051 & 0.053 & 0.053 & \textbf{0.050} & 0.051 & 0.052 & 0.053 & 0.054 & 0.113 & 0.052 & 0.053 \\
bizitobs\_service/10S/medium & \underline{0.018} & 0.023 & 0.024 & 0.022 & 0.025 & 0.023 & \textbf{0.018} & 0.027 & 0.031 & 0.025 & 0.034 & 0.096 & 0.041 & 0.048 \\
bizitobs\_service/10S/short & 0.011 & 0.011 & \textbf{0.010} & 0.010 & \underline{0.010} & 0.012 & 0.010 & 0.011 & 0.012 & 0.010 & 0.014 & 0.051 & 0.019 & 0.040 \\
ett1/15T/long & 0.237 & \textbf{0.223} & \underline{0.227} & 0.241 & 0.231 & 0.234 & 0.255 & 0.251 & 0.234 & 0.236 & 0.268 & 0.298 & 0.259 & 0.340 \\
ett1/15T/medium & 0.232 & \textbf{0.224} & \underline{0.230} & 0.233 & 0.231 & 0.237 & 0.251 & 0.260 & 0.237 & 0.233 & 0.260 & 0.281 & 0.253 & 0.322 \\
ett1/15T/short & \underline{0.154} & 0.163 & 0.157 & 0.165 & 0.157 & 0.161 & 0.157 & 0.162 & \textbf{0.153} & 0.159 & 0.160 & 0.158 & 0.167 & 0.241 \\
ett1/D/short & \textbf{0.262} & 0.285 & 0.275 & \underline{0.274} & 0.280 & 0.277 & 0.300 & 0.284 & 0.276 & 0.275 & 0.287 & 0.287 & 0.298 & 0.408 \\
ett1/H/long & 0.274 & \underline{0.259} & 0.266 & 0.275 & 0.266 & \textbf{0.258} & 0.288 & 0.267 & 0.269 & 0.267 & 0.323 & 0.311 & 0.295 & 0.471 \\
ett1/H/medium & 0.262 & \textbf{0.245} & 0.258 & 0.261 & 0.258 & \underline{0.252} & 0.287 & 0.254 & 0.258 & 0.256 & 0.287 & 0.303 & 0.283 & 0.435 \\
ett1/H/short & 0.178 & 0.176 & 0.177 & \textbf{0.171} & \underline{0.175} & 0.176 & 0.188 & 0.194 & 0.184 & 0.177 & 0.185 & 0.181 & 0.194 & 0.240 \\
ett1/W/short & 0.287 & 0.286 & 0.256 & 0.271 & 0.262 & 0.278 & \textbf{0.241} & 0.263 & 0.249 & 0.256 & \underline{0.249} & 0.296 & 0.284 & 0.312 \\
ett2/15T/long & 0.102 & 0.091 & 0.090 & 0.093 & 0.090 & 0.092 & 0.097 & \textbf{0.088} & 0.091 & \underline{0.089} & 0.102 & 0.111 & 0.101 & 0.133 \\
ett2/15T/medium & 0.095 & \underline{0.088} & 0.090 & \textbf{0.087} & 0.090 & 0.089 & 0.094 & 0.093 & 0.090 & 0.089 & 0.098 & 0.110 & 0.100 & 0.124 \\
ett2/15T/short & 0.065 & 0.065 & \textbf{0.060} & 0.062 & 0.060 & 0.066 & 0.063 & 0.068 & 0.061 & \underline{0.060} & 0.066 & 0.067 & 0.073 & 0.096 \\
ett2/D/short & 0.091 & 0.092 & 0.090 & 0.094 & \textbf{0.088} & 0.094 & 0.092 & 0.111 & \underline{0.089} & 0.090 & 0.093 & 0.094 & 0.126 & 0.153 \\
ett2/H/long & 0.117 & \underline{0.104} & 0.107 & 0.105 & 0.105 & 0.114 & \textbf{0.101} & 0.108 & 0.107 & 0.107 & 0.109 & 0.117 & 0.139 & 0.208 \\
ett2/H/medium & 0.111 & 0.103 & 0.110 & 0.109 & 0.107 & 0.107 & \textbf{0.101} & \underline{0.102} & 0.106 & 0.108 & 0.111 & 0.115 & 0.121 & 0.186 \\
ett2/H/short & \underline{0.063} & 0.064 & 0.065 & 0.064 & 0.065 & 0.064 & 0.064 & 0.065 & 0.066 & 0.065 & 0.064 & \textbf{0.063} & 0.073 & 0.089 \\
ett2/W/short & 0.087 & 0.085 & 0.088 & 0.090 & \underline{0.084} & 0.087 & 0.087 & 0.106 & 0.085 & \textbf{0.081} & 0.085 & 0.088 & 0.099 & 0.134 \\
jena\_weather/10T/long & \textbf{0.047} & 0.049 & \underline{0.047} & 0.051 & 0.047 & 0.049 & 0.050 & 0.050 & 0.048 & 0.047 & 0.060 & 0.064 & 0.053 & 0.237 \\
jena\_weather/10T/medium & \underline{0.046} & 0.049 & \textbf{0.046} & 0.050 & 0.046 & 0.048 & 0.049 & 0.049 & 0.047 & 0.046 & 0.059 & 0.057 & 0.054 & 0.212 \\
jena\_weather/10T/short & 0.027 & 0.028 & 0.025 & 0.030 & \underline{0.025} & 0.027 & 0.028 & 0.027 & 0.026 & \textbf{0.025} & 0.036 & 0.033 & 0.034 & 0.155 \\
jena\_weather/D/short & \underline{0.044} & 0.045 & 0.051 & 0.047 & 0.047 & 0.044 & 0.045 & 0.051 & 0.055 & 0.053 & \textbf{0.043} & 0.045 & 0.047 & 0.211 \\
jena\_weather/H/long & 0.059 & \textbf{0.052} & 0.061 & 0.059 & 0.060 & 0.059 & \underline{0.055} & 0.057 & 0.057 & 0.057 & 0.058 & 0.062 & 0.103 & 0.419 \\
jena\_weather/H/medium & 0.050 & \textbf{0.049} & 0.051 & \underline{0.050} & 0.051 & 0.053 & 0.051 & 0.053 & 0.053 & 0.051 & 0.055 & 0.054 & 0.058 & 0.343 \\
jena\_weather/H/short & 0.044 & \textbf{0.040} & 0.041 & 0.042 & 0.041 & \underline{0.041} & 0.043 & 0.042 & 0.041 & 0.041 & 0.042 & 0.042 & 0.042 & 0.154 \\

\bottomrule
\end{tabular}
\end{adjustbox}

\end{sidewaystable}

\begin{sidewaystable}[p]
\centering

\caption{
CRPS scores on the
univariate data subset of the GIFT-Eval benchmark. The \textbf{best} and \underline{second-best} scores are highlighted.
}
\label{tab:gift_eval_univariate_wql}

\begin{adjustbox}{
    max width=0.90\textheight,
    max totalheight=0.58\textwidth,
    center
}
\begin{tabular}{lllllllllllllll}
\toprule

Dataset
& \makecell[c]{iAmTime}
& \makecell[c]{TiRex-2}
& \makecell[c]{Toto 2.0\\2.5B}
& \makecell[c]{Chronos-2}
& \makecell[c]{Toto 2.0\\1B}
& \makecell[c]{TiRex}
& \makecell[c]{TimesFM-2.5}
& \makecell[c]{Toto 1.0}
& \makecell[c]{Toto 2.0\\22M}
& \makecell[c]{Toto 2.0\\313M}
& \makecell[c]{Moirai 2.0}
& \makecell[c]{Chronos\\Bolt Base}
& \makecell[c]{TabPFN-TS}
& \makecell[c]{Seasonal\\Naive} \\

\midrule

car\_parts/M/short & 0.980 & 0.953 & 1.180 & 0.966 & 1.262 & 0.995 & 0.942 & \textbf{0.899} & 0.995 & 1.091 & \underline{0.936} & 0.995 & 0.970 & 1.722 \\
covid\_deaths/D/short & 0.032 & 0.032 & \underline{0.025} & 0.035 & \textbf{0.024} & 0.032 & 0.035 & 0.027 & 0.029 & 0.032 & 0.028 & 0.047 & 0.041 & 0.127 \\
electricity/15T/long & \textbf{0.069} & 0.074 & 0.072 & \underline{0.072} & 0.073 & 0.078 & 0.077 & 0.086 & 0.078 & 0.073 & 0.083 & 0.084 & 0.081 & 0.113 \\
electricity/15T/medium & \textbf{0.069} & 0.075 & 0.072 & \underline{0.071} & 0.073 & 0.079 & 0.077 & 0.086 & 0.079 & 0.073 & 0.080 & 0.083 & 0.083 & 0.113 \\
electricity/15T/short & \textbf{0.076} & 0.084 & 0.083 & 0.078 & 0.085 & 0.091 & 0.095 & 0.099 & 0.089 & 0.084 & \underline{0.077} & 0.082 & 0.097 & 0.165 \\
electricity/D/short & \textbf{0.049} & 0.055 & 0.057 & 0.058 & 0.057 & 0.054 & 0.054 & 0.059 & 0.058 & 0.057 & \underline{0.054} & 0.055 & 0.063 & 0.104 \\
electricity/H/long & \textbf{0.075} & 0.084 & 0.091 & 0.088 & 0.090 & 0.094 & \underline{0.082} & 0.083 & 0.095 & 0.091 & 0.097 & 0.098 & 0.108 & 0.153 \\
electricity/H/medium & \textbf{0.071} & 0.075 & 0.076 & 0.076 & 0.077 & 0.079 & \underline{0.073} & 0.075 & 0.082 & 0.077 & 0.080 & 0.081 & 0.088 & 0.127 \\
electricity/H/short & 0.064 & 0.065 & 0.072 & 0.068 & 0.073 & \underline{0.064} & 0.065 & 0.069 & 0.076 & 0.073 & 0.065 & \textbf{0.064} & 0.072 & 0.106 \\
electricity/W/short & \textbf{0.043} & 0.050 & 0.054 & 0.056 & 0.056 & 0.048 & \underline{0.046} & 0.064 & 0.057 & 0.056 & 0.066 & 0.047 & 0.055 & 0.099 \\
hierarchical\_sales/D/short & 0.571 & 0.576 & 0.578 & 0.579 & 0.580 & \textbf{0.570} & 0.574 & \underline{0.570} & 0.579 & 0.579 & 0.577 & 0.576 & 0.592 & 1.736 \\
hierarchical\_sales/W/short & \underline{0.341} & 0.349 & 0.346 & \textbf{0.341} & 0.346 & 0.348 & 0.348 & 0.356 & 0.347 & 0.349 & 0.352 & 0.353 & 0.345 & 0.832 \\
hospital/M/short & 0.053 & 0.052 & \underline{0.050} & 0.051 & 0.050 & 0.051 & 0.051 & 0.052 & 0.051 & \textbf{0.049} & 0.052 & 0.057 & 0.054 & 0.062 \\
kdd\_cup\_2018/D/short & \textbf{0.357} & 0.372 & 0.378 & 0.367 & 0.378 & 0.376 & 0.375 & 0.387 & 0.381 & 0.378 & 0.389 & 0.372 & \underline{0.362} & 0.675 \\
kdd\_cup\_2018/H/long & \underline{0.403} & 0.436 & 0.451 & 0.445 & 0.451 & 0.441 & 0.440 & 0.457 & 0.450 & 0.451 & 0.518 & \textbf{0.300} & 0.478 & 0.936 \\
kdd\_cup\_2018/H/medium & \underline{0.397} & 0.425 & 0.430 & 0.417 & 0.432 & 0.421 & 0.419 & 0.441 & 0.426 & 0.429 & 0.495 & \textbf{0.301} & 0.450 & 0.759 \\
kdd\_cup\_2018/H/short & \underline{0.319} & 0.370 & 0.381 & 0.374 & 0.379 & 0.378 & 0.374 & 0.403 & 0.379 & 0.379 & 0.426 & \textbf{0.246} & 0.418 & 0.548 \\
loop\_seattle/5T/long & \textbf{0.066} & \underline{0.072} & 0.074 & 0.080 & 0.074 & 0.084 & 0.076 & 0.077 & 0.079 & 0.073 & 0.087 & 0.129 & 0.090 & 0.127 \\
loop\_seattle/5T/medium & \textbf{0.063} & \underline{0.067} & 0.067 & 0.074 & 0.067 & 0.078 & 0.071 & 0.072 & 0.074 & 0.067 & 0.080 & 0.116 & 0.087 & 0.117 \\
loop\_seattle/5T/short & 0.047 & 0.048 & 0.046 & \underline{0.046} & 0.046 & 0.048 & 0.049 & 0.048 & 0.048 & \textbf{0.046} & 0.046 & 0.055 & 0.053 & 0.081 \\
loop\_seattle/D/short & \textbf{0.039} & 0.042 & 0.043 & 0.042 & 0.043 & 0.042 & \underline{0.041} & 0.044 & 0.043 & 0.043 & 0.043 & 0.044 & 0.043 & 0.103 \\
loop\_seattle/H/long & \underline{0.056} & \textbf{0.052} & 0.057 & 0.059 & 0.058 & 0.060 & 0.057 & 0.065 & 0.061 & 0.060 & 0.066 & 0.076 & 0.063 & 0.187 \\
loop\_seattle/H/medium & 0.056 & \textbf{0.050} & 0.058 & 0.063 & 0.060 & 0.064 & \underline{0.055} & 0.064 & 0.065 & 0.063 & 0.069 & 0.076 & 0.067 & 0.162 \\
loop\_seattle/H/short & \underline{0.049} & \textbf{0.047} & 0.056 & 0.058 & 0.057 & 0.058 & 0.051 & 0.063 & 0.061 & 0.058 & 0.062 & 0.065 & 0.063 & 0.104 \\
m4\_daily/D/short & 0.021 & 0.021 & 0.021 & 0.023 & 0.021 & 0.021 & 0.022 & 0.022 & 0.021 & \underline{0.021} & \textbf{0.020} & 0.021 & 0.023 & 0.024 \\
m4\_hourly/H/short & 0.028 & 0.030 & \underline{0.021} & 0.023 & 0.023 & \textbf{0.020} & 0.024 & 0.035 & 0.032 & 0.023 & 0.024 & 0.025 & 0.030 & 0.038 \\
m4\_monthly/M/short & \textbf{0.086} & 0.094 & 0.092 & \underline{0.090} & 0.092 & 0.091 & 0.094 & 0.097 & 0.091 & 0.092 & 0.095 & 0.094 & 0.094 & 0.122 \\
m4\_quarterly/Q/short & 0.073 & 0.075 & 0.073 & 0.073 & \underline{0.073} & \textbf{0.073} & 0.076 & 0.078 & 0.074 & 0.073 & 0.075 & 0.077 & 0.078 & 0.098 \\
m4\_weekly/W/short & 0.037 & 0.038 & \textbf{0.036} & 0.037 & 0.037 & 0.037 & 0.037 & 0.049 & 0.038 & \underline{0.037} & 0.040 & 0.038 & 0.037 & 0.061 \\
m4\_yearly/A/short & \textbf{0.109} & 0.124 & 0.120 & \underline{0.112} & 0.119 & 0.116 & 0.127 & 0.122 & 0.116 & 0.119 & 0.116 & 0.121 & 0.118 & 0.137 \\
m\_dense/D/short & 0.061 & 0.066 & 0.063 & \textbf{0.058} & 0.064 & 0.068 & 0.063 & 0.075 & 0.069 & 0.063 & 0.070 & 0.069 & \underline{0.061} & 0.227 \\
m\_dense/H/long & 0.131 & 0.118 & 0.124 & \underline{0.113} & 0.122 & 0.115 & \textbf{0.112} & 0.128 & 0.127 & 0.129 & 0.120 & 0.170 & 0.165 & 0.419 \\
m\_dense/H/medium & 0.122 & \underline{0.116} & 0.119 & 0.117 & 0.117 & 0.116 & \textbf{0.113} & 0.121 & 0.121 & 0.123 & 0.120 & 0.157 & 0.160 & 0.377 \\
m\_dense/H/short & 0.127 & 0.131 & 0.126 & 0.128 & 0.128 & 0.128 & 0.128 & 0.148 & 0.134 & \underline{0.126} & 0.132 & \textbf{0.125} & 0.155 & 0.275 \\
restaurant/D/short & \textbf{0.251} & \underline{0.254} & 0.260 & 0.254 & 0.260 & 0.255 & 0.257 & 0.297 & 0.265 & 0.260 & 0.260 & 0.264 & 0.263 & 0.677 \\
saugeen/D/short & 0.347 & 0.357 & 0.359 & 0.349 & 0.353 & 0.358 & \underline{0.333} & 0.353 & 0.352 & 0.356 & \textbf{0.328} & 0.338 & 0.373 & 0.585 \\
saugeen/M/short & 0.291 & 0.306 & 0.286 & 0.287 & 0.282 & 0.297 & 0.298 & 0.299 & 0.284 & \underline{0.280} & 0.291 & 0.296 & \textbf{0.276} & 0.445 \\
saugeen/W/short & \textbf{0.338} & 0.367 & 0.351 & \underline{0.347} & 0.359 & 0.350 & 0.378 & 0.390 & 0.370 & 0.353 & 0.419 & 0.363 & 0.395 & 0.734 \\
solar/10T/long & \textbf{0.264} & 0.307 & \underline{0.288} & 0.290 & 0.289 & 0.321 & 0.331 & 0.352 & 0.327 & 0.300 & 0.408 & 0.443 & 0.331 & 0.674 \\
solar/10T/medium & \textbf{0.270} & 0.324 & 0.316 & \underline{0.290} & 0.314 & 0.334 & 0.340 & 0.353 & 0.335 & 0.334 & 0.412 & 0.436 & 0.326 & 0.655 \\
solar/10T/short & \underline{0.363} & 0.529 & 0.423 & 0.386 & 0.431 & 0.541 & 0.562 & 0.541 & 0.459 & 0.425 & \textbf{0.350} & 0.511 & 0.458 & 0.860 \\
solar/D/short & \textbf{0.266} & 0.276 & 0.279 & \underline{0.267} & 0.275 & 0.273 & 0.282 & 0.290 & 0.277 & 0.278 & 0.303 & 0.287 & 0.269 & 0.559 \\
solar/H/long & 0.320 & \textbf{0.169} & 0.296 & 0.340 & 0.298 & \underline{0.287} & 0.324 & 0.331 & 0.325 & 0.302 & 0.332 & 0.405 & 0.351 & 1.078 \\
solar/H/medium & 0.290 & \textbf{0.164} & 0.299 & 0.363 & 0.303 & \underline{0.285} & 0.314 & 0.331 & 0.318 & 0.311 & 0.333 & 0.368 & 0.313 & 0.946 \\
solar/H/short & 0.276 & \textbf{0.154} & 0.311 & 0.346 & 0.331 & \underline{0.273} & 0.337 & 0.328 & 0.320 & 0.317 & 0.342 & 0.298 & 0.336 & 0.592 \\
solar/W/short & 0.140 & 0.149 & 0.161 & 0.139 & 0.154 & 0.144 & 0.170 & 0.186 & 0.164 & 0.150 & 0.163 & \underline{0.133} & \textbf{0.120} & 0.210 \\
sz\_taxi/15T/long & \textbf{0.196} & \underline{0.197} & 0.201 & 0.201 & 0.200 & 0.198 & 0.200 & 0.202 & 0.201 & 0.199 & 0.217 & 0.248 & 0.242 & 0.428 \\
sz\_taxi/15T/medium & \textbf{0.200} & \underline{0.201} & 0.203 & 0.202 & 0.203 & 0.202 & 0.203 & 0.205 & 0.203 & 0.201 & 0.214 & 0.244 & 0.230 & 0.379 \\
sz\_taxi/15T/short & \underline{0.199} & 0.201 & 0.199 & 0.200 & \textbf{0.199} & 0.200 & 0.201 & 0.203 & 0.201 & 0.200 & 0.201 & 0.202 & 0.209 & 0.309 \\
sz\_taxi/H/short & \textbf{0.134} & 0.136 & \underline{0.134} & 0.135 & 0.135 & 0.135 & 0.136 & 0.137 & 0.136 & 0.135 & 0.136 & 0.136 & 0.140 & 0.214 \\
temperature\_rain/D/short & \textbf{0.375} & 0.545 & 0.560 & 0.539 & 0.560 & 0.547 & 0.553 & 0.560 & 0.564 & 0.560 & 0.561 & \underline{0.538} & 0.569 & 1.268 \\
us\_births/D/short & 0.018 & 0.020 & 0.017 & \underline{0.017} & 0.018 & 0.023 & 0.018 & 0.026 & 0.021 & 0.018 & 0.020 & 0.026 & \textbf{0.016} & 0.120 \\
us\_births/M/short & \textbf{0.012} & 0.016 & 0.013 & 0.013 & 0.015 & \underline{0.012} & 0.013 & 0.013 & 0.014 & 0.014 & 0.017 & 0.019 & 0.015 & 0.017 \\
us\_births/W/short & 0.013 & 0.013 & \textbf{0.010} & 0.011 & 0.011 & 0.012 & 0.013 & 0.014 & 0.013 & 0.011 & 0.011 & 0.013 & \underline{0.011} & 0.019 \\

\bottomrule
\end{tabular}
\end{adjustbox}

\end{sidewaystable}

\clearpage

\clearpage
\section{Extended Classification Evaluation \& Details}
\label{app:classification_details}

We evaluate on multiple univariate and multivariate classification datasets from the UCR-UEA collections \citep{dau2019ucr, bagnall2018uea}.
The univariate UCR suite spans sensors, device traces, motion, physiology, and similar domains. The archive was designed so that evaluation is standardized via fixed train/test splits across these datasets.
The Multivariate UEA suite includes datasets from similar domains but with multiple channels with: synchronized channels, predefined splits, and typically equal-length tensors (channels $\times$ time). Domains include speech, motion, physiology, and related sensing setups.

\subsection{In-Context Classification Protocol}
\label{app:icl_classification_protocol}

We evaluate in-context classification on UCR classification datasets using the same support-query prompt interface used by iAmTime for forecasting. 
We use the subset of data from Table~\ref{tab:ucr_classification_dataset} with the column \texttt{Test Subset} marked as \texttt{3}.
For each evaluation episode, a test time series from a dataset is selected as the query input. 
The support set is constructed from the corresponding training split. 

Let $C$ denote the number of classes in the dataset. 
We sample the number of support examples $K$ uniformly subject to
\(
K \in [C, 8],
\)
and ensure that the support set contains at least one example from each class. 
This guarantees that the class semantics are identifiable from the prompt while keeping the number of demonstrations small.

Each support example consists of an input time series and a class label represented as a constant output sequence. 
Specifically, for a support instance with class $y_i \in \mathcal{C}$, we encode the output as
\[
Y_i^{\mathrm{fut}} = c(y_i)\cdot \mathbf{1},
\]
where $c(y_i)$ is the scalar code assigned to class $y_i$. 
To prevent the model from relying on a global label-code convention, the class-code mapping is sampled independently for each episode. 
For a dataset with $C$ classes, we construct $C$ evenly spaced scalar codes and randomly permute their assignment to classes:
\(
\{c_1,\ldots,c_C\}
=
\mathrm{Permute}
\left(
\mathrm{linspace}(1,9,C)
\right).
\)
Thus, the same semantic class may correspond to different scalar codes across episodes, forcing the model to infer the label mapping from the support examples.

The query contains only the input time series, with the future label sequence withheld. 
The model predicts a sequence $\hat{Y}_q^{\mathrm{fut}}$, and is reduced to a scalar prediction by averaging over the output horizon.

The predicted class is obtained by nearest-code decoding:
\[
\hat{y}_q
=
\arg\min_{y \in \mathcal{C}}
\left|
\hat{c}_q - c(y)
\right|.
\]
This protocol directly measures whether iAmTime can infer an episode-local classification rule from demonstrations and apply it to a held-out query without parameter updates or a task-specific classification head.

\subsection{Embedding-Based Linear Probe Protocol}
\label{app:linear_probe_protocol}

We also evaluate classification using a frozen-encoder linear-probe protocol to assess the representational quality of iAmTime independently of generative in-context decoding. 
We test separately on subsets of the univariate and multivariate datasets
in Table~\ref{tab:ucr_classification_dataset}. 
The column \texttt{Test Subset} is marked as \texttt{1} for univariate datasets and \texttt{2} for multivariate datasets.
For each dataset, the original train/test split is preserved. 
Each model is used as a frozen feature extractor, and a linear classifier is trained on embeddings extracted from the training split and evaluated on the corresponding test split.

For iAmTime and Chronos-2, which natively support multivariate inputs, the full univariate or multivariate time series is passed to the model jointly. 
The encoder output is used as the representation and flattened across the relevant token, patch, and variate dimensions to form a fixed-dimensional feature vector. 
For Chronos-family models that operate primarily in a univariate setting, each variate is embedded independently and then concatenated to obtain a single feature vector. 
We use flatten-concatenation as the pooling strategy.

We train three standard linear classifiers on top of the frozen embeddings: ridge classification, logistic regression, and linear SVM. 
All classifiers use the default hyperparameters provided by \texttt{scikit-learn}. 
For each dataset and encoder, we report the best result among the three probe types:
\[
\mathrm{score}(f)
=
\max
\{
\mathrm{score}_{\mathrm{ridge}},
\mathrm{score}_{\mathrm{logreg}},
\mathrm{score}_{\mathrm{linearsvc}}
\}.
\]
This evaluation is intended to measure whether the frozen representations are linearly separable for downstream time-series classification. 
Because only the probe is trained and the encoder remains fixed, performance reflects the transferability and discriminative structure of the learned time-series embeddings rather than task-specific fine-tuning of iAmTime.

\subsection{Comparison with In-Context and Instruction-Style Classifiers}
\label{app:icl_classification_baselines}

\begin{table}[t]
\centering
\caption{
Comparison with directly relevant in-context, multi-task, and foundation-model classification baselines across 10 multivariate datasets. Lower rank is better.
}
\label{tab:icl_classification_baselines}
\scriptsize
\setlength{\tabcolsep}{3pt}
\renewcommand{\arraystretch}{1.10}

\resizebox{\columnwidth}{!}{
\begin{tabular}{llcccc}
\toprule
Model Paradigm
& Model
& Mean Acc. $\uparrow$
& Median Acc. $\uparrow$
& Mean Rank $\downarrow$
& Median Rank $\downarrow$
\\
\midrule

Multi-Task
& UniTS-PMT
& 69.11
& 74.05
& 6.90
& 8.00
\\

Foundation Model for Classification
& MantisV2
& 66.94
& 67.95
& 6.25
& 5.75
\\

Multi-Task
& MOMENT+LP
& 68.40
& 71.90
& 6.40
& 7.25
\\

ICL based Classification
& TiCT
& 60.66
& 66.85
& 8.90
& 10.00
\\

ICL based Classification
& TimEE
& 70.10
& 70.65
& 6.15
& 7.25
\\

ICL based Classification
& RocketPFN
& \underline{72.86}
& 77.20
& 5.10
& 5.00
\\

Foundation Model for Forecasting
& Chronos-2+LP
& 70.69
& 77.00
& \underline{4.40}
& \underline{4.00}
\\

Multi-Task
& Chronos-Bolt-M+LP
& 67.37
& 70.85
& 6.50
& 6.50
\\

ICL based Multi-Task
& Chronos-Bolt-ICL
& 69.35
& 72.45
& 5.60
& 5.00
\\

ICL based Multi-Task
& iAmTime (w/ ICL)
& 71.03
& \underline{77.50}
& 5.55
& 5.75
\\

Foundation Model for Multi-Task
& iAmTime+LP (w/o ICL)
& \textbf{73.34}
& \textbf{79.65}
& \textbf{4.25}
& \textbf{3.50}
\\

\bottomrule
\end{tabular}
}
\end{table}

We further compare against directly relevant in-context and instruction-style classifiers, together with multi-task and linear-probe references. 
Table~\ref{tab:icl_classification_baselines} shows that iAmTime with four demonstrations substantially exceeds TiCT \citep{yeh2025tict}, improves over TimEE \citep{kuken2026timee} in both mean and median accuracy, and remains competitive with RocketPFN. 
RocketPFN achieves the strongest mean accuracy and rank among the direct ICL classifiers, while iAmTime attains the higher median accuracy. 
For completeness, UniTS-ST reaches $75.04\%$ mean accuracy under single-task training, representing a strong but different supervision setting. 
Finally, iAmTime+LP achieves the best aggregate accuracy and rank statistics overall, confirming strong representation quality without conflating linear probing with direct demonstration-conditioned task adaptation.

\subsection{Additional Results}

\begin{figure}[tp]
    \centering
    \begin{subfigure}[t]{\linewidth}
        \resizebox{\linewidth}{!}{\input{images/classification_ucr_icl_by_num_class.pgf}}
        \caption{Univariate, grouped by number of classes.}
        \label{fig:classification_ucr_icl_by_num_class}
    \end{subfigure}

    \begin{subfigure}[t]{\linewidth}
        \resizebox{\linewidth}{!}{\input{images/classification_ucr_icl_by_type.pgf}}
        \caption{Multivariate, grouped by dataset type.}
        \label{fig:classification_ucr_icl_by_type}
    \end{subfigure}

    \caption{Classification performance of iAmTime using ICL demonstrations on the UCR classification benchmark, grouped by number of classes and dataset type, for univariate and multivariate settings.}
    \label{fig:classification_ucr_icl_by_num_class_type}
\end{figure}

\begin{figure}[tp]
    \centering
    \begin{subfigure}[t]{\linewidth}
        \resizebox{\linewidth}{!}{\input{images/classification_ucr_univ_by_num_class.pgf}}
        \caption{Univariate, grouped by number of classes.}
        \label{fig:classification_ucr_univ_by_num_class}
    \end{subfigure}

    \begin{subfigure}[t]{\linewidth}
        \resizebox{\linewidth}{!}{\input{images/classification_ucr_multi_by_num_class.pgf}}
        \caption{Multivariate, grouped by number of classes.}
        \label{fig:classification_ucr_multi_by_num_class}
    \end{subfigure}

    \begin{subfigure}[t]{\linewidth}
        \resizebox{\linewidth}{!}{\input{images/classification_ucr_univ_by_type.pgf}}
        \caption{Univariate, grouped by dataset type.}
        \label{fig:classification_ucr_univ_by_type}
    \end{subfigure}

    \begin{subfigure}[t]{\linewidth}
        \resizebox{\linewidth}{!}{\input{images/classification_ucr_multi_by_type.pgf}}
        \caption{Multivariate, grouped by dataset type.}
        \label{fig:classification_ucr_multi_by_type}
    \end{subfigure}

    \caption{Classification performance of iAmTime and other models on the UCR classification benchmark, grouped by number of classes (top two) and dataset type (bottom two), for univariate and multivariate settings.}
    \label{fig:classification_ucr_by_num_class_type}
\end{figure}

In Fig. \ref{fig:classification_ucr_icl_by_num_class_type}, we break down the in-context classification performance by the number of classes and the dataset family.
Generally, we observe a decreasing trend in performance as the number of classes increases, which is expected due to the increased difficulty of inferring more complex decision boundaries from limited examples.

In Fig. \ref{fig:classification_ucr_by_num_class_type}, we break down the linear probe classification performance by the number of classes and the dataset family. 
ROCKET/MiniROCKET (blue) generally dominate, with iAmTime+LP competitive on many types — particularly strong on Traffic, Simulated, and Motion datasets.

Table \ref{tab:classification_full_results} provides the full results for both in-context and linear-probe classification across the 10 datasets following prior work \citep{gao2024units, goswami2024moment}.

\begin{table}[t]
\centering
\caption{
Classification accuracy (\%) on ten multivariate time-series datasets.
}
\label{tab:classification_full_results}

\scriptsize
\setlength{\tabcolsep}{2pt}
\renewcommand{\arraystretch}{1.08}

\subcaptionbox{
Results for DTW through DLinear.
\label{tab:classification_full_results_a}
}{%
\resizebox{\columnwidth}{!}{%
\begin{tabular}{lccccccccccccccccc}
\toprule
Dataset
& DTW
& \makecell{XG-\\Boost}
& Rocket
& LSTM
& LSTNet
& LSSL
& TCN
& \makecell{Trans-\\former}
& Reformer
& Informer
& Pyraformer
& Autoformer
& \makecell{Station-\\former}
& FEDformer
& ETSformer
& Flowformer
& DLinear \\
\midrule

EthanolConcentration
& 32.3 & 43.7 & 45.2 & 32.3 & 39.9 & 31.1 & 28.9 & 32.7
& 31.9 & 31.6 & 30.8 & 31.6 & 32.7 & 31.2 & 28.1 & 33.8 & 32.6 \\

FaceDetection
& 52.9 & 63.3 & 64.7 & 57.7 & 65.7 & 66.7 & 52.8 & 67.3
& {68.6} & 67.0 & 65.7 & 68.4 & 68.0 & 66.0 & 66.3
& 67.6 & 68.0 \\

Handwriting
& 28.6 & 15.8 & {58.8} & 15.2 & 25.8 & 24.6
& {53.3} & 32.0 & 27.4 & 32.8 & 29.4 & 36.7 & 31.6
& 28.0 & 32.5 & 33.8 & 27.0 \\

Heartbeat
& 71.7 & 73.2 & 75.6 & 72.2 & 77.1 & 72.7 & 75.6 & 76.1
& 77.1 & 80.5 & 75.6 & 74.6 & 73.7 & 73.7 & 71.2 & 77.6 & 75.1 \\

JapaneseVowels
& 94.9 & 86.5 & 96.2 & 79.7 & 98.1 & 98.4
& {98.9} & 98.7 & 97.8 & {98.9}
& 98.4 & 96.2 & {99.2} & 98.4 & 95.9
& {98.9} & 96.2 \\

PEMS-SF
& 71.1 & {98.3} & 75.1 & 39.9 & 86.7 & 86.1 & 68.8 & 82.1
& 82.7 & 81.5 & 83.2 & 82.7 & 87.3 & 80.9 & 86.0 & 83.8 & 75.1 \\

SelfRegulationSCP1
& 77.7 & 84.6 & 90.8 & 68.9 & 84.0 & 90.8 & 84.6 & 92.2
& 90.4 & 90.1 & 88.1 & 84.0 & 89.4 & 88.7 & 89.6 & 92.5 & 87.3 \\

SelfRegulationSCP2
& 53.9 & 48.9 & 53.3 & 46.6 & 52.8 & 52.2 & 55.6 & 53.9
& 56.7 & 53.3 & 53.3 & 50.6 & {57.2} & 54.4
& 55.0 & 56.1 & 50.5 \\

SpokenArabicDigits
& 96.3 & 69.6 & 71.2 & 31.9
& {100.0} & {100.0}
& 95.6 & 98.4 & 97.0
& {100.0} & {99.6}
& {100.0} & {100.0}
& {100.0} & {100.0}
& 98.8 & 81.4 \\

UWaveGestureLibrary
& 90.3 & 75.9 & {94.4} & 41.2 & 87.8 & 85.9 & 88.4
& 85.6 & 85.6 & 85.6 & 83.4 & 85.9 & 87.5 & 85.3 & 85.0
& 86.6 & 82.1 \\

\midrule
Average Acc.
& 67.0 & 66.0 & 72.5 & 48.6 & 71.8 & 70.9 & 70.3 & 71.9
& 71.5 & 72.1 & 70.8 & 71.1 & 72.7 & 70.7 & 71.0 & 73.0 & 67.5 \\

\bottomrule
\end{tabular}%
}%
}

\vspace{0.6em}

\subcaptionbox{
Results for LightTSformer through iAmTime.
\label{tab:classification_full_results_b}
}{%
\resizebox{\columnwidth}{!}{%
\begin{tabular}{lcccccccccccccccc}
\toprule
Dataset
& \makecell{LightTS-\\former}
& GPT4TS
& TimesNet
& UniTS-ST
& \makecell{UNITS-\\SUP}
& \makecell{UNITS-\\PMT}
& MantisV2
& MOMENT
& TiCT
& TimEE
& RocketPFN
& Chronos-2
& \makecell{Chronos-Bolt\\Base}
& \makecell{Chronos-Bolt\\ICL}
& \makecell{iAmTime\\(w/ ICL)}
& \makecell{iAmTime+LP\\(w/o ICL)} \\
\midrule

EthanolConcentration
& 29.7 & 33.5 & 35.7 & 37.6 & 30.9 & 35.2 & 41.4 & 35.7
& 27.4 & {61.2} & {68.8} & 37.0 & 30.5 & 35.8
& 46.1 & 45.1 \\

FaceDetection
& 67.5 & 66.1 & {68.6} & {70.5}
& 65.4 & 58.0 & 63.3 & 63.3 & 63.3 & 63.3 & 64.2 & 66.0
& 51.3 & 62.1 & 68.5 & 66.4 \\

Handwriting
& 26.1 & 31.7 & 32.1 & 29.7 & 30.4 & 30.6 & 28.1 & 30.8
& 20.8 & 42.2 & 15.8 & 31.4 & 35.1 & 34.9 & 31.2 & 38.8 \\

Heartbeat
& 75.1 & 69.8 & 78.0 & 80.0 & 63.9 & 65.4
& {82.9} & 72.2 & 72.2 & 71.7 & 71.3 & 72.8
& 74.1 & 82.8 & 72.5 & {94.7} \\

JapaneseVowels
& 96.2 & 94.6 & 98.4 & 97.8 & 92.2 & 90.3 & 66.3 & 71.6
& 64.1 & 78.4 & 83.1 & 95.9 & {98.9}
& 90.1 & 98.3 & 97.9 \\

PEMS-SF
& 88.4 & 79.2 & 89.6 & 93.1 & 83.2 & 82.7 & 94.2 & 89.6
& 90.2 & {95.4} & 92.9 & 81.2 & 67.6 & 59.5
& 86.7 & 71.4 \\

SelfRegulationSCP1
& 89.8 & 92.2 & 91.8 & {93.9} & 90.1 & 90.9 & 81.9
& 84.0 & 72.0 & 88.7 & 86.7 & 90.5 & 84.3
& {95.0} & 90.1 & 92.2 \\

SelfRegulationSCP2
& 51.1 & 45.6 & {57.2} & {61.1}
& 48.9 & {57.2} & 51.7 & 47.8 & 51.1 & 43.9
& 51.6 & 54.5 & 53.9 & 54.9 & 42.3 & 46.9 \\

SpokenArabicDigits
& {100.0} & 97.5 & 99.0 & 98.9 & 96.8 & 95.5
& 69.6 & 98.1 & 69.6 & 69.6 & 99.5 & 94.5 & 90.5
& 89.3 & 92.1 & 92.1 \\

UWaveGestureLibrary
& 80.3 & 81.9 & 85.3 & 87.8 & 82.2 & 85.3 & 90.0 & 90.9
& 75.9 & 86.6 & {94.7} & 83.1 & 87.5 & 89.1
& 82.5 & 87.9 \\

\midrule
Average Acc.
& 70.4 & 69.2 & {73.6} & {75.0}
& 68.4 & 69.1 & 66.9 & 68.4 & 60.7 & 70.1 & 72.9
& 70.69 & 67.4 & 69.3 & 71.03 & 73.3 \\

\bottomrule
\end{tabular}%
}%
}

\end{table}

\clearpage
\section{Extended Imputation Evaluation \& Details}
\label{app:imputation_details}

\begin{table}[h]
\centering
\caption{Full results of block-wise imputation with patch size 8 tasks on 4 datasets.}
\label{tab:imputation_block8_full_results}

\scriptsize
\setlength{\tabcolsep}{1.8pt}
\renewcommand{\arraystretch}{1.05}

\resizebox{\textwidth}{!}{%
\begin{tabular}{ll*{2}{cc}|*{11}{cc}}
\toprule
\multirow{2}{*}{Dataset}
& \multirow{2}{*}{\makecell{Mask\\Ratio}}
& \multicolumn{2}{c}{iAmTime}
& \multicolumn{2}{c}{\makecell{iAmTime\\(w/o ICL)}}
& \multicolumn{2}{c}{Chronos-2}
& \multicolumn{2}{c}{\makecell{Chronos-Bolt\\Base}}
& \multicolumn{2}{c}{\makecell{Chronos-Bolt\\ICL}}
& \multicolumn{2}{c}{UniTS}
& \multicolumn{2}{c}{MOMENT}
& \multicolumn{2}{c}{GPT4TS}
& \multicolumn{2}{c}{TimesNet}
& \multicolumn{2}{c}{Naive}
& \multicolumn{2}{c}{Linear}
& \multicolumn{2}{c}{Nearest}
& \multicolumn{2}{c}{Cubic} \\

\cmidrule(lr){3-4}
\cmidrule(lr){5-6}
\cmidrule(lr){7-8}
\cmidrule(lr){9-10}
\cmidrule(lr){11-12}
\cmidrule(lr){13-14}
\cmidrule(lr){15-16}
\cmidrule(lr){17-18}
\cmidrule(lr){19-20}
\cmidrule(lr){21-22}
\cmidrule(lr){23-24}
\cmidrule(lr){25-26}
\cmidrule(lr){27-28}

& & MSE & MAE
& MSE & MAE
& MSE & MAE
& MSE & MAE
& MSE & MAE
& MSE & MAE
& MSE & MAE
& MSE & MAE
& MSE & MAE
& MSE & MAE
& MSE & MAE
& MSE & MAE
& MSE & MAE \\
\midrule

\multirow{3}{*}{ETTm1}
& 25\%
& \textbf{0.069} & \textbf{0.140}
& 0.190 & 0.201
& 0.195 & 0.210
& 0.327 & 0.400
& 0.190 & 0.248
& 0.092 & 0.141
& 0.071 & 0.169
& 0.071 & 0.144
& 0.080 & 0.194
& 0.395 & 0.341
& 0.114 & 0.202
& 0.171 & 0.234
& 0.539 & 0.424 \\

& 50\%
& \textbf{0.081} & \textbf{0.148}
& 0.213 & 0.203
& 0.240 & 0.246
& 0.348 & 0.413
& 0.193 & 0.257
& 0.108 & 0.167
& 0.086 & 0.169
& \textbf{0.081} & 0.149
& 0.088 & 0.197
& 0.679 & 0.448
& 0.265 & 0.291
& 0.346 & 0.316
& 1.715 & 0.680 \\

& Avg
& \textbf{0.075} & \textbf{0.144}
& 0.202 & 0.202
& 0.218 & 0.228
& 0.338 & 0.407
& 0.192 & 0.252
& 0.100 & 0.154
& 0.079 & 0.169
& 0.076 & 0.147
& 0.084 & 0.196
& 0.537 & 0.395
& 0.190 & 0.247
& 0.259 & 0.275
& 1.127 & 0.552 \\
\midrule

\multirow{3}{*}{ETTh1}
& 25\%
& \textbf{0.130} & \textbf{0.170}
& 0.261 & 0.200
& 0.274 & 0.261
& 0.495 & 0.509
& 0.223 & 0.308
& 0.139 & 0.171
& 0.142 & 0.238
& 0.278 & 0.267
& 0.154 & 0.261
& 1.311 & 0.686
& 0.833 & 0.540
& 0.954 & 0.581
& 1.433 & 0.772 \\

& 50\%
& \textbf{0.132} & \textbf{0.181}
& 0.299 & 0.279
& 0.301 & 0.289
& 0.496 & 0.518
& 0.235 & 0.328
& 0.157 & \textbf{0.181}
& \textbf{0.132} & 0.231
& 0.213 & 0.243
& 0.192 & 0.267
& 1.103 & 0.643
& 0.840 & 0.559
& 0.963 & 0.601
& 3.681 & 1.204 \\

& Avg
& \textbf{0.131} & \textbf{0.175}
& 0.280 & 0.240
& 0.287 & 0.275
& 0.496 & 0.513
& 0.229 & 0.318
& 0.148 & 0.176
& 0.137 & 0.235
& 0.246 & 0.255
& 0.173 & 0.264
& 1.207 & 0.665
& 0.837 & 0.550
& 0.959 & 0.591
& 2.557 & 0.988 \\
\midrule

\multirow{3}{*}{Electricity}
& 25\%
& \textbf{0.055} & \textbf{0.112}
& 0.102 & 0.202
& 0.100 & 0.213
& 0.407 & 0.502
& 0.292 & 0.366
& 0.059 & 0.116
& 0.093 & 0.211
& 0.073 & 0.184
& 0.121 & 0.246
& 1.447 & 0.857
& 0.654 & 0.554
& 0.815 & 0.582
& 1.619 & 0.769 \\

& 50\%
& \textbf{0.065} & \textbf{0.120}
& 0.172 & 0.301
& 0.169 & 0.254
& 0.419 & 0.519
& 0.310 & 0.367
& 0.068 & 0.124
& 0.092 & 0.210
& 0.075 & 0.185
& 0.126 & 0.249
& 1.581 & 0.915
& 1.002 & 0.712
& 1.239 & 0.766
& 3.978 & 1.213 \\

& Avg
& \textbf{0.060} & \textbf{0.116}
& 0.137 & 0.251
& 0.135 & 0.233
& 0.413 & 0.511
& 0.301 & 0.367
& 0.063 & 0.120
& 0.093 & 0.211
& 0.074 & 0.185
& 0.124 & 0.248
& 1.514 & 0.886
& 0.828 & 0.633
& 1.027 & 0.674
& 2.799 & 0.991 \\
\midrule

\multirow{3}{*}{Weather}
& 25\%
& \textbf{0.022} & 0.065
& 0.072 & 0.088
& 0.074 & 0.089
& 0.113 & 0.206
& 0.071 & 0.102
& 0.037 & \textbf{0.053}
& 0.036 & 0.078
& 0.030 & 0.071
& 0.037 & 0.100
& 0.127 & 0.104
& 0.075 & 0.065
& 0.094 & 0.076
& 0.297 & 0.125 \\

& 50\%
& 0.028 & 0.075
& 0.075 & 0.099
& 0.076 & 0.101
& 0.130 & 0.222
& 0.089 & 0.161
& 0.048 & \textbf{0.068}
& 0.035 & 0.075
& \textbf{0.026} & 0.069
& 0.035 & 0.098
& 0.124 & 0.127
& 0.066 & 0.078
& 0.086 & 0.090
& 0.831 & 0.197 \\

& Avg
& \textbf{0.025} & 0.070
& 0.073 & 0.094
& 0.075 & 0.095
& 0.122 & 0.214
& 0.080 & 0.132
& 0.043 & \textbf{0.060}
& 0.036 & 0.077
& 0.028 & 0.070
& 0.036 & 0.099
& 0.126 & 0.116
& 0.071 & 0.072
& 0.090 & 0.083
& 0.564 & 0.161 \\

\bottomrule
\end{tabular}%
}
\end{table}

This section provides additional details and results for the imputation experiments extending Section~\ref{sec:imputation_tasks}. 
We evaluate on ETTm1, ETTh1, Electricity, and Weather under both contiguous-block and random-point masking. 
The model is not trained on these evaluation datasets for imputation; imputation behavior is learned only from randomly masked pretraining series as described in Section~\ref{sec:imputation_reconstruction_task}.

\subsection{Imputation Protocol}
\label{app:imputation_protocol}

For every test query, demonstrations are sampled from the corresponding training split. 
The input portion of each demonstration contains a masked time series, while its demonstrated ``future'' contains the original target history with the masked positions restored to their true values. 
For exogenous channels, the future portion repeats their corresponding histories. 
The query follows the same masked-input structure, and the model reconstructs the missing target values using the demonstrated mapping. 
We use four demonstrations and report MSE and MAE over the masked positions.

We additionally evaluate the same checkpoint with zero demonstrations (iAmTime w/o ICL). 
Without examples specifying reconstruction as the desired mapping, the model does not reliably infer the imputation task and instead tends toward its learned forecasting behavior. 
The resulting performance gap therefore directly measures the contribution of inference-time demonstrations to task adaptation.

\subsection{Contiguous Block Imputation}
\label{app:block_imputation}

Following the harder block-masking setting used by prior multi-task models such as UniTS \citep{gao2024units} and MOMENT \citep{goswami2024moment}, we randomly mask contiguous subsequences of length 8 until approximately $25\%$ or $50\%$ of the input is hidden. 
Unlike independent point masking, this removes local temporal neighborhoods and requires reconstruction from longer-range context.

Table~\ref{tab:imputation_block8_full_results} shows that iAmTime with four demonstrations achieves the strongest aggregate MSE and MAE across the four datasets, outperforming UniTS, MOMENT, and Chronos-family baselines. 
The advantage over iAmTime without demonstrations is particularly large, confirming that the model uses support examples to switch from forecasting to reconstruction rather than relying only on its pretrained forecasting behavior. 
Performance remains strong as the masked proportion increases from $25\%$ to $50\%$, indicating robustness to substantial contiguous missing regions.

\subsection{Random-Point Imputation}
\label{app:point_imputation}

\begin{table}[p]
\centering
\caption{
Full results of random-point imputation on ETTm1, ETTh1, Electricity,
and Weather. Lower MSE and MAE are better. Best results are bolded,
and second-best results are underlined.
}
\label{tab:imputation_point_full_results}

\scriptsize
\setlength{\tabcolsep}{1.7pt}
\renewcommand{\arraystretch}{1.04}

\subcaptionbox{
iAmTime and recent time-series models.
\label{tab:imputation_point_full_results_a}
}{%
\resizebox{\columnwidth}{!}{%
\begin{tabular}{ll*{10}{cc}}
\toprule
\multirow{2}{*}{Dataset}
& \multirow{2}{*}{\makecell{Mask\\Ratio}}
& \multicolumn{2}{c}{iAmTime}
& \multicolumn{2}{c}{\makecell{iAmTime\\(w/o ICL)}}
& \multicolumn{2}{c}{Chronos-2}
& \multicolumn{2}{c}{\makecell{Chronos-Bolt\\Base}}
& \multicolumn{2}{c}{\makecell{Chronos-Bolt\\ICL}}
& \multicolumn{2}{c}{UniTS}
& \multicolumn{2}{c}{MOMENT}
& \multicolumn{2}{c}{TimesNet}
& \multicolumn{2}{c}{ETSformer}
& \multicolumn{2}{c}{LightTS} \\

\cmidrule(lr){3-4}
\cmidrule(lr){5-6}
\cmidrule(lr){7-8}
\cmidrule(lr){9-10}
\cmidrule(lr){11-12}
\cmidrule(lr){13-14}
\cmidrule(lr){15-16}
\cmidrule(lr){17-18}
\cmidrule(lr){19-20}
\cmidrule(lr){21-22}

& & MSE & MAE
& MSE & MAE & MSE & MAE & MSE & MAE
& MSE & MAE & MSE & MAE & MSE & MAE
& MSE & MAE & MSE & MAE & MSE & MAE \\
\midrule

\multirow{5}{*}{ETTm1}
& 12.5\%
& \textbf{0.011} & \textbf{0.070}
& 0.049 & 0.110
& 0.052 & 0.122
& 0.082 & 0.125
& 0.032 & \underline{0.075}
& \underline{0.015} & 0.079
& 0.016 & 0.090
& 0.019 & 0.092
& 0.067 & 0.188
& 0.075 & 0.180 \\

& 25\%
& \textbf{0.015} & \textbf{0.079}
& 0.060 & 0.123
& 0.062 & 0.125
& 0.087 & 0.128
& 0.032 & 0.093
& \underline{0.017} & \underline{0.082}
& 0.023 & 0.098
& 0.023 & 0.101
& 0.096 & 0.229
& 0.093 & 0.206 \\

& 37.5\%
& \textbf{0.016} & \textbf{0.085}
& 0.078 & 0.126
& 0.073 & 0.125
& 0.090 & 0.132
& 0.036 & 0.095
& \underline{0.019} & \underline{0.088}
& 0.028 & 0.108
& 0.029 & 0.111
& 0.133 & 0.271
& 0.113 & 0.231 \\

& 50\%
& \textbf{0.021} & \textbf{0.090}
& 0.085 & 0.153
& 0.089 & 0.127
& 0.105 & 0.159
& 0.042 & 0.098
& \underline{0.024} & \underline{0.097}
& 0.031 & 0.115
& 0.036 & 0.124
& 0.186 & 0.323
& 0.134 & 0.255 \\

& Avg
& \textbf{0.016} & \textbf{0.081}
& 0.068 & 0.128
& 0.069 & 0.125
& 0.091 & 0.136
& 0.035 & 0.090
& \underline{0.019} & \underline{0.087}
& 0.024 & 0.103
& 0.027 & 0.107
& 0.120 & 0.253
& 0.104 & 0.218 \\
\midrule

\multirow{5}{*}{ETTh1}
& 12.5\%
& \textbf{0.027} & \textbf{0.094}
& 0.090 & 0.178
& 0.100 & 0.174
& 0.124 & 0.185
& 0.045 & 0.122
& \underline{0.032} & \underline{0.118}
& 0.056 & 0.155
& 0.057 & 0.159
& 0.126 & 0.263
& 0.240 & 0.345 \\

& 25\%
& \textbf{0.029} & \textbf{0.102}
& 0.100 & 0.180
& 0.104 & 0.176
& 0.124 & 0.188
& 0.047 & 0.128
& \underline{0.036} & \underline{0.126}
& 0.060 & 0.176
& 0.069 & 0.178
& 0.169 & 0.304
& 0.265 & 0.364 \\

& 37.5\%
& \textbf{0.038} & \textbf{0.117}
& 0.108 & 0.180
& 0.113 & 0.202
& 0.125 & 0.189
& \underline{0.047} & 0.148
& \underline{0.047} & \underline{0.142}
& 0.077 & 0.192
& 0.084 & 0.196
& 0.220 & 0.347
& 0.296 & 0.382 \\

& 50\%
& \textbf{0.046} & \textbf{0.135}
& 0.112 & 0.195
& 0.125 & 0.205
& 0.134 & 0.215
& \underline{0.052} & \underline{0.156}
& 0.060 & 0.160
& 0.099 & 0.208
& 0.102 & 0.215
& 0.293 & 0.402
& 0.334 & 0.404 \\

& Avg
& \textbf{0.035} & \textbf{0.112}
& 0.103 & 0.183
& 0.111 & 0.189
& 0.127 & 0.194
& 0.048 & 0.138
& \underline{0.043} & \underline{0.136}
& 0.073 & 0.183
& 0.078 & 0.187
& 0.202 & 0.329
& 0.284 & 0.373 \\
\midrule

\multirow{5}{*}{Electricity}
& 12.5\%
& \textbf{0.029} & \textbf{0.104}
& 0.067 & 0.175
& 0.072 & 0.181
& 0.102 & 0.198
& 0.058 & 0.130
& \underline{0.031} & \underline{0.112}
& 0.079 & 0.200
& 0.085 & 0.202
& 0.196 & 0.321
& 0.102 & 0.229 \\

& 25\%
& \textbf{0.033} & \textbf{0.110}
& 0.076 & 0.223
& 0.082 & 0.202
& 0.105 & 0.230
& 0.062 & 0.133
& \underline{0.035} & \underline{0.119}
& 0.083 & 0.199
& 0.089 & 0.206
& 0.207 & 0.332
& 0.121 & 0.252 \\

& 37.5\%
& \textbf{0.038} & \textbf{0.123}
& 0.080 & 0.230
& 0.095 & 0.211
& 0.105 & 0.230
& 0.063 & 0.138
& \underline{0.040} & \underline{0.128}
& 0.094 & 0.212
& 0.094 & 0.213
& 0.219 & 0.344
& 0.141 & 0.273 \\

& 50\%
& \textbf{0.043} & \textbf{0.130}
& 0.106 & 0.241
& 0.112 & 0.245
& 0.107 & 0.256
& 0.064 & 0.141
& \underline{0.046} & \underline{0.138}
& 0.099 & 0.212
& 0.100 & 0.221
& 0.235 & 0.357
& 0.160 & 0.293 \\

& Avg
& \textbf{0.036} & \textbf{0.117}
& 0.082 & 0.217
& 0.090 & 0.210
& 0.105 & 0.229
& 0.062 & 0.136
& \underline{0.038} & \underline{0.124}
& 0.089 & 0.206
& 0.092 & 0.210
& 0.214 & 0.339
& 0.131 & 0.262 \\
\midrule

\multirow{5}{*}{Weather}
& 12.5\%
& \textbf{0.020} & \underline{0.040}
& 0.048 & 0.086
& 0.049 & 0.085
& 0.075 & 0.137
& 0.035 & 0.051
& 0.025 & 0.041
& \underline{0.021} & \textbf{0.037}
& 0.025 & 0.045
& 0.057 & 0.141
& 0.047 & 0.101 \\

& 25\%
& \textbf{0.022} & \textbf{0.043}
& 0.050 & 0.086
& 0.051 & 0.085
& 0.087 & 0.148
& 0.044 & 0.081
& \underline{0.026} & \underline{0.044}
& \textbf{0.022} & \underline{0.044}
& 0.029 & 0.052
& 0.065 & 0.155
& 0.052 & 0.111 \\

& 37.5\%
& \underline{0.026} & \textbf{0.043}
& 0.053 & 0.095
& 0.055 & 0.098
& 0.101 & 0.171
& 0.045 & 0.097
& 0.027 & \underline{0.045}
& \textbf{0.025} & 0.051
& 0.031 & 0.057
& 0.081 & 0.180
& 0.058 & 0.121 \\

& 50\%
& \textbf{0.028} & \textbf{0.044}
& 0.073 & 0.101
& 0.061 & 0.103
& 0.118 & 0.177
& 0.052 & 0.099
& \underline{0.029} & \underline{0.049}
& \underline{0.029} & 0.061
& 0.034 & 0.062
& 0.102 & 0.207
& 0.065 & 0.133 \\

& Avg
& \textbf{0.024} & \textbf{0.042}
& 0.056 & 0.092
& 0.054 & 0.093
& 0.095 & 0.158
& 0.044 & 0.082
& \underline{0.026} & \underline{0.045}
& \textbf{0.024} & 0.048
& 0.030 & 0.054
& 0.076 & 0.171
& 0.055 & 0.117 \\

\bottomrule
\end{tabular}%
}%
}

\vspace{0.5em}

\subcaptionbox{
Additional forecasting and reconstruction baselines.
\label{tab:imputation_point_full_results_b}
}{%
\resizebox{\columnwidth}{!}{%
\begin{tabular}{ll*{11}{cc}}
\toprule
\multirow{2}{*}{Dataset}
& \multirow{2}{*}{\makecell{Mask\\Ratio}}
& \multicolumn{2}{c}{DLinear}
& \multicolumn{2}{c}{FEDformer}
& \multicolumn{2}{c}{Stationformer}
& \multicolumn{2}{c}{Autoformer}
& \multicolumn{2}{c}{Pyraformer}
& \multicolumn{2}{c}{Informer}
& \multicolumn{2}{c}{LogTransformer}
& \multicolumn{2}{c}{Reformer}
& \multicolumn{2}{c}{LSTM}
& \multicolumn{2}{c}{TCN}
& \multicolumn{2}{c}{LSSL} \\

\cmidrule(lr){3-4}
\cmidrule(lr){5-6}
\cmidrule(lr){7-8}
\cmidrule(lr){9-10}
\cmidrule(lr){11-12}
\cmidrule(lr){13-14}
\cmidrule(lr){15-16}
\cmidrule(lr){17-18}
\cmidrule(lr){19-20}
\cmidrule(lr){21-22}
\cmidrule(lr){23-24}

& & MSE & MAE
& MSE & MAE & MSE & MAE & MSE & MAE
& MSE & MAE & MSE & MAE & MSE & MAE
& MSE & MAE & MSE & MAE & MSE & MAE
& MSE & MAE \\
\midrule

\multirow{5}{*}{ETTm1}
& 12.5\%
& 0.058 & 0.162
& 0.035 & 0.135
& 0.026 & 0.107
& 0.034 & 0.124
& 0.670 & 0.541
& 0.047 & 0.155
& 0.041 & 0.141
& 0.032 & 0.126
& 0.974 & 0.780
& 0.510 & 0.493
& 0.101 & 0.231 \\

& 25\%
& 0.080 & 0.193
& 0.052 & 0.166
& 0.032 & 0.119
& 0.046 & 0.144
& 0.689 & 0.553
& 0.063 & 0.180
& 0.044 & 0.144
& 0.042 & 0.146
& 1.032 & 0.807
& 0.518 & 0.500
& 0.106 & 0.235 \\

& 37.5\%
& 0.103 & 0.219
& 0.069 & 0.191
& 0.039 & 0.131
& 0.057 & 0.161
& 0.737 & 0.581
& 0.079 & 0.200
& 0.052 & 0.158
& 0.063 & 0.182
& 0.999 & 0.792
& 0.516 & 0.499
& 0.116 & 0.246 \\

& 50\%
& 0.132 & 0.248
& 0.089 & 0.218
& 0.047 & 0.145
& 0.067 & 0.174
& 0.770 & 0.605
& 0.093 & 0.218
& 0.063 & 0.173
& 0.082 & 0.208
& 0.952 & 0.763
& 0.519 & 0.496
& 0.129 & 0.260 \\

& Avg
& 0.093 & 0.206
& 0.062 & 0.177
& 0.036 & 0.126
& 0.051 & 0.150
& 0.717 & 0.570
& 0.071 & 0.188
& 0.050 & 0.154
& 0.055 & 0.166
& 0.989 & 0.786
& 0.516 & 0.497
& 0.113 & 0.243 \\
\midrule

\multirow{5}{*}{ETTh1}
& 12.5\%
& 0.151 & 0.267
& 0.070 & 0.190
& 0.060 & 0.165
& 0.074 & 0.182
& 0.857 & 0.609
& 0.114 & 0.234
& 0.229 & 0.330
& 0.074 & 0.194
& 1.265 & 0.896
& 0.599 & 0.554
& 0.422 & 0.461 \\

& 25\%
& 0.180 & 0.292
& 0.106 & 0.236
& 0.080 & 0.189
& 0.090 & 0.203
& 0.829 & 0.672
& 0.140 & 0.262
& 0.207 & 0.323
& 0.102 & 0.227
& 1.262 & 0.883
& 0.610 & 0.567
& 0.412 & 0.456 \\

& 37.5\%
& 0.215 & 0.318
& 0.124 & 0.258
& 0.102 & 0.212
& 0.109 & 0.222
& 0.830 & 0.675
& 0.174 & 0.293
& 0.210 & 0.328
& 0.135 & 0.261
& 1.200 & 0.867
& 0.628 & 0.577
& 0.421 & 0.461 \\

& 50\%
& 0.257 & 0.347
& 0.165 & 0.299
& 0.133 & 0.240
& 0.137 & 0.248
& 0.854 & 0.691
& 0.215 & 0.325
& 0.230 & 0.348
& 0.179 & 0.298
& 1.174 & 0.849
& 0.648 & 0.587
& 0.443 & 0.473 \\

& Avg
& 0.201 & 0.306
& 0.117 & 0.246
& 0.094 & 0.201
& 0.103 & 0.214
& 0.842 & 0.662
& 0.161 & 0.279
& 0.219 & 0.332
& 0.122 & 0.245
& 1.225 & 0.873
& 0.621 & 0.571
& 0.424 & 0.463 \\
\midrule

\multirow{5}{*}{Electricity}
& 12.5\%
& 0.092 & 0.214
& 0.107 & 0.237
& 0.093 & 0.210
& 0.089 & 0.210
& 0.297 & 0.383
& 0.218 & 0.326
& 0.164 & 0.296
& 0.190 & 0.308
& 0.277 & 0.366
& 0.621 & 0.620
& 0.217 & 0.341 \\

& 25\%
& 0.118 & 0.247
& 0.120 & 0.251
& 0.097 & 0.214
& 0.096 & 0.220
& 0.294 & 0.380
& 0.219 & 0.326
& 0.169 & 0.299
& 0.197 & 0.312
& 0.281 & 0.369
& 0.559 & 0.585
& 0.219 & 0.341 \\

& 37.5\%
& 0.144 & 0.276
& 0.136 & 0.266
& 0.102 & 0.220
& 0.104 & 0.229
& 0.296 & 0.381
& 0.222 & 0.328
& 0.178 & 0.305
& 0.203 & 0.315
& 0.275 & 0.364
& 0.567 & 0.588
& 0.223 & 0.343 \\

& 50\%
& 0.175 & 0.305
& 0.158 & 0.284
& 0.108 & 0.228
& 0.113 & 0.239
& 0.299 & 0.383
& 0.228 & 0.331
& 0.187 & 0.312
& 0.210 & 0.319
& 0.273 & 0.361
& 0.581 & 0.597
& 0.229 & 0.347 \\

& Avg
& 0.132 & 0.260
& 0.130 & 0.259
& 0.100 & 0.218
& 0.101 & 0.225
& 0.297 & 0.382
& 0.222 & 0.328
& 0.175 & 0.303
& 0.200 & 0.313
& 0.277 & 0.365
& 0.582 & 0.597
& 0.222 & 0.343 \\
\midrule

\multirow{5}{*}{Weather}
& 12.5\%
& 0.039 & 0.084
& 0.041 & 0.107
& 0.027 & 0.051
& 0.026 & 0.047
& 0.140 & 0.220
& 0.037 & 0.093
& 0.037 & 0.072
& 0.031 & 0.076
& 0.296 & 0.379
& 0.176 & 0.287
& 0.036 & 0.095 \\

& 25\%
& 0.048 & 0.103
& 0.064 & 0.163
& 0.029 & 0.056
& 0.030 & 0.054
& 0.147 & 0.229
& 0.042 & 0.100
& 0.038 & 0.074
& 0.035 & 0.082
& 0.327 & 0.409
& 0.187 & 0.293
& 0.042 & 0.104 \\

& 37.5\%
& 0.057 & 0.117
& 0.107 & 0.229
& 0.033 & 0.062
& 0.032 & 0.060
& 0.156 & 0.240
& 0.049 & 0.111
& 0.039 & 0.078
& 0.040 & 0.091
& 0.406 & 0.463
& 0.172 & 0.281
& 0.047 & 0.112 \\

& 50\%
& 0.066 & 0.134
& 0.183 & 0.312
& 0.037 & 0.068
& 0.037 & 0.067
& 0.164 & 0.249
& 0.053 & 0.114
& 0.042 & 0.082
& 0.046 & 0.099
& 0.431 & 0.483
& 0.195 & 0.303
& 0.054 & 0.123 \\

& Avg
& 0.052 & 0.110
& 0.099 & 0.203
& 0.032 & 0.059
& 0.031 & 0.057
& 0.152 & 0.235
& 0.045 & 0.104
& 0.039 & 0.076
& 0.038 & 0.087
& 0.365 & 0.434
& 0.183 & 0.291
& 0.045 & 0.108 \\

\bottomrule
\end{tabular}%
}%
}

\end{table}

We also evaluate the conventional random-point setting by independently masking $12.5\%$, $25\%$, $37.5\%$, or $50\%$ of the observations and predicting only the missing values.
Table~\ref{tab:imputation_point_full_results} shows that iAmTime remains highly competitive across all masking ratios and achieves the strongest overall performance across the four datasets. 

Overall, the two settings show that imputation is not implemented through a dedicated reconstruction head. 
The same iAmTime checkpoint performs forecasting or imputation according to the input-output relation supplied by the demonstrations, while removing those demonstrations substantially weakens reconstruction performance.

\clearpage
\section{Extended Anomaly Detection Evaluation \& Details}
\label{app:anomaly_detection_details}

\begin{table*}[h]
\centering
\caption{
Anomaly-detection results on SMD, MSL, SMAP, SWaT, and PSM.
iAmTime-DM predicts anomaly masks directly from demonstrations, while
iAmTime-RC performs reconstruction and scores anomalies using reconstruction
error. We report precision, recall, and F1 for each dataset and average F1
across datasets.
}
\label{tab:anomaly_detection_results}

\scriptsize
\setlength{\tabcolsep}{2.2pt}
\renewcommand{\arraystretch}{1.05}

\resizebox{\textwidth}{!}{%
\begin{tabular}{l*{5}{ccc}|c}
\toprule
\multirow{2}{*}{Model}
& \multicolumn{3}{c}{SMD}
& \multicolumn{3}{c}{MSL}
& \multicolumn{3}{c}{SMAP}
& \multicolumn{3}{c}{SWaT}
& \multicolumn{3}{c}{PSM}
& \multirow{2}{*}{\makecell{Average\\F1}} \\
\cmidrule(lr){2-4}
\cmidrule(lr){5-7}
\cmidrule(lr){8-10}
\cmidrule(lr){11-13}
\cmidrule(lr){14-16}
& P & R & F1
& P & R & F1
& P & R & F1
& P & R & F1
& P & R & F1
& \\
\midrule

LSTM
& 78.52 & 65.47 & 71.41
& 78.04 & 86.22 & 81.93
& 91.06 & 57.49 & 70.48
& 78.06 & 91.72 & 84.34
& 69.24 & 99.53 & 81.67
& 77.97 \\

Transformer
& 83.58 & 76.13 & 79.56
& 71.57 & 87.37 & 78.68
& 89.37 & 57.12 & 69.70
& 68.84 & 96.53 & 80.37
& 62.75 & 96.56 & 76.07
& 76.88 \\

LogTransformer
& 83.46 & 70.13 & 76.21
& 73.05 & 87.37 & 79.57
& 89.15 & 57.59 & 69.97
& 68.67 & 97.32 & 80.52
& 63.06 & 98.00 & 76.74
& 76.60 \\

TCN
& 84.06 & 79.07 & 81.49
& 75.11 & 82.44 & 78.60
& 86.90 & 59.23 & 70.45
& 76.59 & 95.71 & 85.09
& 54.59 & 99.77 & 70.57
& 77.24 \\

Reformer
& 82.58 & 69.24 & 75.32
& 85.51 & 83.31 & 84.40
& 90.91 & 57.44 & 70.40
& 72.50 & 96.53 & 82.80
& 59.93 & 95.38 & 73.61
& 77.31 \\

Informer
& 86.60 & 77.23 & 81.65
& 81.77 & 86.48 & 84.06
& 90.11 & 57.13 & 69.92
& 70.29 & 96.75 & 81.43
& 64.27 & 96.33 & 77.10
& 78.83 \\

Anomaly Transformer
& 88.91 & 82.23 & 85.49
& 79.61 & 87.37 & 83.31
& 91.85 & 58.11 & 71.18
& 72.51 & 97.32 & 83.10
& 68.35 & 94.72 & 79.40
& 80.50 \\

Pyraformer
& 85.61 & 80.61 & 83.04
& 83.81 & 85.93 & 84.86
& 92.54 & 57.71 & 71.09
& 87.92 & 96.00 & 91.78
& 71.67 & 96.02 & 82.08
& 82.57 \\

Autoformer
& 88.06 & 82.35 & 85.11
& 77.27 & 80.92 & 79.05
& 90.40 & 58.62 & 71.12
& 89.85 & 95.81 & 92.74
& 99.08 & 88.15 & 93.29
& 84.26 \\

LSSL
& 78.51 & 65.32 & 71.31
& 77.55 & 88.18 & 82.53
& 89.43 & 53.43 & 66.90
& 79.05 & 93.72 & 85.76
& 66.02 & 92.93 & 77.20
& 76.74 \\

Stationformer
& 88.33 & 81.21 & 84.62
& 68.55 & 89.14 & 77.50
& 89.37 & 59.02 & 71.09
& 68.03 & 96.75 & 79.88
& 97.82 & 96.76 & 97.29
& 82.08 \\

DLinear
& 83.62 & 71.52 & 77.10
& 84.34 & 85.42 & 84.88
& 92.32 & 55.41 & 69.26
& 80.91 & 95.30 & 87.52
& 98.28 & 89.26 & 93.55
& 82.46 \\

ETSformer
& 87.44 & 79.23 & 83.13
& 85.13 & 84.93 & 85.03
& 92.25 & 55.75 & 69.50
& 90.02 & 80.36 & 84.91
& 99.31 & 85.28 & 91.76
& 82.87 \\

LightTS
& 87.10 & 78.42 & 82.53
& 82.40 & 75.78 & 78.95
& 92.58 & 55.27 & 69.21
& 91.98 & 94.72 & 93.33
& 98.37 & 95.97 & 97.15
& 84.23 \\

FEDformer
& 87.95 & 82.39 & 85.08
& 77.14 & 80.07 & 78.57
& 90.47 & 58.10 & 70.76
& 90.17 & 96.42 & 93.19
& 97.31 & 97.16 & 97.23
& 84.97 \\

TimesNet
& 87.95 & 81.54 & 84.62
& 89.55 & 75.29 & 81.80
& 90.14 & 56.56 & 69.50
& 90.76 & 95.35 & 93.00
& 98.50 & 96.29 & 97.38
& 85.26 \\

MOMENT
& 88.64 & 84.22 & 86.37
& 89.73 & 76.49 & 82.58
& 91.76 & 66.29 & 76.97
& 91.57 & 94.76 & 93.14
& 98.56 & 96.29 & 97.41
& 87.29 \\

UniTS
& 89.32 & 86.90 & 88.09
& 89.91 & 77.68 & 83.46
& 93.37 & 76.02 & 83.80
& 92.37 & 94.17 & 93.26
& 98.62 & 96.28 & 97.43
& 89.21 \\

Chronos-2
& 80.99 & 67.80 & 73.81
& 75.55 & 86.80 & 80.78
& 90.11 & 57.54 & 70.23
& 73.37 & 94.52 & 82.61
& 66.15 & 98.77 & 79.23
& 77.33 \\

Chronos-Bolt Base
& 83.52 & 73.13 & 77.98
& 72.31 & 87.37 & 79.13
& 89.26 & 57.36 & 69.84
& 68.76 & 96.93 & 80.45
& 62.91 & 97.28 & 76.40
& 76.76 \\

Chronos-Bolt ICL
& 87.95 & 81.97 & 84.85
& 83.35 & 77.68 & 80.41
& 90.31 & 57.33 & 70.13
& 90.47 & 95.89 & 93.10
& 97.91 & 96.73 & 97.31
& 85.16 \\

\midrule

\textbf{iAmTime-DM}
& 80.80 & 94.77 & 87.23
& 91.37 & 89.36 & 90.35
& 93.61 & 99.34 & 96.39
& 93.31 & 95.00 & 96.54
& 97.77 & 96.90 & 97.33
& \textbf{93.57} \\

\textbf{iAmTime-RC}
& 81.44 & 97.99 & 88.95
& 91.65 & 91.84 & 91.75
& 93.76 & 99.80 & 96.69
& 93.03 & 95.03 & 96.39
& 98.12 & 98.56 & 98.34
& \textbf{94.42} \\

iAmTime (w/o ICL)
& 85.33 & 78.15 & 81.58
& 78.44 & 84.46 & 81.34
& 88.51 & 58.18 & 70.21
& 73.44 & 96.23 & 83.30
& 59.43 & 98.05 & 74.00
& 78.09 \\

\bottomrule
\end{tabular}%
}
\end{table*}

This section provides the evaluation protocol, baseline implementation details, and additional results for the anomaly-detection experiments complementing Section~\ref{sec:anomaly_detection_tasks}. 
Importantly, the anomaly-detection capability is not trained using any of the evaluation datasets: anomaly episodes during meta-training are generated using synthetic anomaly injection as described in Section~\ref{sec:anomaly_detection_task}.

\subsection{Datasets and Evaluation Protocol}
\label{app:anomaly_detection_protocol}

We follow the multi-task anomaly-detection evaluation used by UniTS \citep{gao2024units} on five widely used multivariate datasets: SMD, MSL, SMAP, SWaT, and PSM. 
Their numbers of variables are 38, 55, 25, 51, and 25, respectively. 
For each query window, the support pool is created using synthetic data, while evaluation is performed on the test windows. 
We report precision, recall, and F1 for each dataset and the mean F1 across datasets, under the point-adjustment (PA) protocol \citep{gao2024units} used by all the reported baselines.

iAmTime supports two anomaly-detection formulations using the same trained model and support-query interface. Both set the threshold from a per-dataset anomaly ratio.

\paragraph{Direct-mask prediction (iAmTime-DM).}
Each demonstration contains an observed time-series window as input and its per-timestep binary anomaly mask as the demonstrated output, where $1$ denotes an anomaly and $0$ a normal timestep. 
The query contains only the observed window, and iAmTime directly predicts its anomaly-mask scores from the demonstrated mapping. 
Future values of the original exogenous channels are masked with NaNs. 
This formulation requires the model to infer the anomaly-detection task and its output semantics directly from demonstrations without a task-specific detection head.

\paragraph{Reconstruction-based detection (iAmTime-RC).}
The same checkpoint can instead be prompted for reconstruction. 
For each demonstration, the future target is the raw target history itself, repeated without denoising or removal of anomalies; future exogenous channels similarly repeat their histories. 
For the query, iAmTime produces a reconstruction that empirically follows the underlying series while smoothing anomalous deviations. 
We therefore use the squared reconstruction error,
$\left(x_t-\hat{x}_t\right)^2$, as the anomaly score. 
Thus, iAmTime-DM and iAmTime-RC use the same model but specify substantially different output mappings solely through the demonstrations.

\subsection{Additional Anomaly Detection Results}
\label{app:anomaly_detection_results}

Across five datasets, as shown in Table~\ref{tab:anomaly_detection_results}, iAmTime-DM achieves an average F1 of $93.57$, outperforming the strongest multi-task baselines UniTS ($89.21$) and MOMENT ($87.29$), as well as specialized anomaly-detection models. 
The reconstruction formulation performs even better, with iAmTime-RC reaching an average F1 of $94.42$. 
The gains are consistent across datasets, with particularly strong performance on MSL, SMAP, SWaT, and PSM.

The two variants also demonstrate task adaptation beyond simply improving anomaly-detection accuracy. 
iAmTime-DM directly predicts a binary anomaly representation from demonstrations, whereas iAmTime-RC produces a reconstructed time series from which anomalies are derived through reconstruction error. 
Despite these different output semantics, no task-specific head or parameter update is introduced. 
In contrast, evaluating the same checkpoint without demonstrations reduces average F1 to $78.09$, further showing that the demonstrated mapping is central to the anomaly-detection behavior.

\section{Implementation Details of Baselines}
\label{app:anomaly_baseline_details}

We follow the baseline implementation and multi-task training protocol of UniTS \citep{gao2024units}. 
Because most baselines cannot directly handle heterogeneous variable counts and task outputs with a single interface, each uses data-specific input modules to project different variable dimensions into a shared representation and task-specific output modules for the required prediction format. 
The shared backbone is then trained jointly under the same fully supervised multi-task setting used by UniTS.

For patch-based approaches, patch size and stride are fixed to 16. 
Baseline backbones use three basic modeling blocks following the UniTS protocol, except GPT4TS \citep{zhou2023one}, which uses its prescribed six GPT blocks. 
We compare against UniTS \citep{gao2024units}, MOMENT \citep{goswami2024moment}, GPT4TS \citep{zhou2023one}, and a broad set of Transformer and sequence-model baselines, including the specialized Anomaly Transformer. 
Unlike iAmTime, these approaches rely on dataset- or task-specific input/output modules or adaptation rather than specifying the output behavior through demonstrations at inference time.

\clearpage
\section{Ablations}
\label{sec:ablation_appendix}

\begin{table}[H]
\centering
\caption{
Fine-grained ablations of training, structural, task-adaptation, output-specialization, task-family, and curriculum components of iAmTime.
}
\label{tab:fine_grained_ablation_appendix}
\scriptsize
\setlength{\tabcolsep}{4pt}
\renewcommand{\arraystretch}{1.12}

\subcaptionbox{
Ablation definitions.
\label{tab:fine_grained_ablation_defs}
}[0.99\columnwidth]{
{
\begin{tabular}{lll}
\toprule
Ablation Type & Variant & Change from full iAmTime \\
\midrule

Full system
& iAmTime
& Full System
\\

\midrule

Training
& -NoMeta
& Forecasting tasks and forecasting examples only
\\

Training
& -NoExmp
& No training demonstrations; auxiliary meta-tasks also removed
\\

Training
& -Meta-NoExmp
& Perform multi-task training; remove examples and Token Stream
\\

\midrule

Structural
& -NoToks-WExmp
& Remove Token Stream but retain examples and multi-task training
\\

Structural
& -NoToks-NoExmp
& Remove Token Stream, examples, and auxiliary meta-tasks
\\

Variate Representation
& -NoPEF
& Remove Per-example fusion attention (Patch Stream)
\\

\midrule

Task Adaptation
& -NoCEA
& Remove token-level Cross-Example Attention (Token Stream)
\\

Task Adaptation
& -NoTWrite
& Remove Token Write/FiLM
\\

Task Adaptation
& -SharedToks
& Tie role-token embeddings while preserving tokens and masks
\\

\midrule

Output Specialization
& -UniMoE
& Replace adaptive routing with uniform expert weights
\\

\midrule

Task-family
& -NoCls
& Remove only classification episodes
\\

Task-family
& -NoImp
& Remove only imputation episodes
\\

Task-family
& -NoAD
& Remove only anomaly-detection episodes
\\

Task-family
& -NoSynthCls
& Remove only synthetic classification episodes
\\

\midrule

Curriculum
& -NoCurriculum
& Perform multi-task training without curriculum
\\

\bottomrule
\end{tabular}
}
}

\hfill

\subcaptionbox{
Performance of each ablation across tasks.
\label{tab:fine_grained_ablation_results}
}[0.99\columnwidth]{
{
\begin{tabular}{lccccc}
\toprule
Variant
& \makecell[c]{Forecasting\\fev Skill\\Score (\%) $\uparrow$}
& \makecell[c]{Forecasting\\GIFT\\CRPS $\downarrow$}
& \makecell[c]{Classification\\UCR ICL\\Acc./F1 $\uparrow$}
& \makecell[c]{Imputation\\Block-8\\MSE/MAE $\downarrow$}
& \makecell[c]{Anomaly\\Detection\\F1 $\uparrow$}
\\
\midrule

iAmTime
& 46.8
& 0.465
& 0.813 / 0.775
& 0.0727 / 0.1265
& 93.57
\\

\midrule

-NoMeta
& 46.4
& 0.468
& 0.224 / 0.176
& 0.1740 / 0.1968
& 74.30
\\

-NoExmp
& 46.4
& 0.471
& 0.238 / 0.184
& 0.1935 / 0.2257
& 72.10
\\

-Meta-NoExmp
& 47.5
& 0.458
& 0.240 / 0.191
& 0.0814 / 0.1305
& 92.20
\\

\midrule

-NoToks-WExmp
& 38.6
& 0.569
& 0.502 / 0.521
& 0.1048 / 0.1542
& 81.55
\\

-NoToks-NoExmp
& 46.2
& 0.489
& 0.207 / 0.158
& 0.1430 / 0.1767
& 76.80
\\

-NoPEF
& 37.8
& 0.583
& 0.515 / 0.531
& 0.0981 / 0.1498
& 80.31
\\

\midrule

-NoCEA
& 44.8
& 0.494
& 0.263 / 0.198
& 0.1404 / 0.1607
& 78.72
\\

-NoTWrite
& 43.9
& 0.516
& 0.246 / 0.194
& 0.1709 / 0.1954
& 79.05
\\

-SharedToks
& 44.9
& 0.496
& 0.689 / 0.645
& 0.0962 / 0.1479
& 89.53
\\

\midrule

-UniMoE
& 45.8
& 0.479
& 0.776 / 0.734
& 0.0798 / 0.1339
& 91.84
\\

\midrule

-NoCls
& 46.3
& 0.468
& 0.274 / 0.218
& 0.0734 / 0.1272
& 93.21
\\

-NoImp
& 46.2
& 0.469
& 0.808 / 0.770
& 0.1421 / 0.1784
& 93.06
\\

-NoAD
& 46.3
& 0.468
& 0.810 / 0.772
& 0.0736 / 0.1271
& 80.64
\\

-NoSynthCls
& 46.9
& 0.460
& 0.611 / 0.568
& 0.0800 / 0.1259
& 85.41
\\

\midrule

-NoCurriculum
& 30.2
& 0.701
& 0.524 / 0.581
& 0.1878 / 0.2413
& 66.26
\\

\bottomrule
\end{tabular}
}
}

\end{table}

\subsection{Training Method Ablations}

We first isolate how the training formulation affects forecasting and broader task adaptation. 
All variants use the same evaluation protocols on fev-bench, GIFT-Eval, and the non-forecasting tasks. 
The results in Table~\ref{tab:fine_grained_ablation_appendix}, together with
Tables~\ref{tab:ablation_fev} and~\ref{tab:ablation_gift}, reveal that the components required for peak forecasting are not identical to those required for inference-time task adaptation.

\paragraph{iAmTime.}
The full model is trained on heterogeneous instruction-conditioned tasks using explicit example-query demonstrations and structured semantic tokens processed by the \emph{Token Stream} (Section~\ref{sec:token_stream}). 
It provides the best overall balance across forecasting, classification, imputation, and anomaly detection.

\paragraph{iAmTime-NoMeta and iAmTime-NoExmp.}
iAmTime-NoMeta restricts training to forecasting tasks and forecasting demonstrations, while iAmTime-NoExmp removes demonstrations during training together with the auxiliary meta-task families. 
Both retain competitive forecasting because the hierarchical encoder can still learn strong temporal and covariate representations, but classification and other task-adaptation abilities degrade sharply. 
This separates ordinary forecasting ability from the instruction-conditioned behavior learned through the broader episodic training distribution.

\paragraph{iAmTime-Meta-NoExmp.}
This variant retains multi-task training but removes examples and the Token Stream. 
Interestingly, it slightly improves forecasting over the full model while largely collapsing ICL classification. 
Thus, heterogeneous multi-task training can learn useful general representations even without demonstrations, while explicit example-conditioned task adaptation requires the support-query interface. 
The result also reveals a small trade-off: optimizing for broader task adaptability need not maximize peak forecasting performance.

Together, these training ablations show that demonstrations and multi-task instruction-conditioned training primarily enable adaptation rather than serving as generic forecasting improvements.

\begin{table}[t]
\center
\caption{Win rate and skill score with respect to SQL and sorted by 
CRPS Rank on fev-bench dataset.
\textbf{iAmTime} is the base version trained using Meta-training tasks and examples.
\textbf{iAmTime-NoExmp} ablates the examples (0 examples) while training (and removes the other meta-training task).
\textbf{iAmTime-NoToks-NoExmp} ablates the use of semantic tokens and examples while training.
\textbf{iAmTime-NoToks-WExmp} ablates the use of semantic tokens but keeps the examples while training.
\textbf{iAmTime-NoMeta} uses only forecasting tasks and forecasting examples to train the model.
}
\label{tab:ablation_fev}
\begin{tabular}{lrrrrrr}
    \toprule
 & MASE & MASE  & CRPS & CRPS  & Avg. Win & Skill \\
 &  & Rank &  & Rank & Rate (\%) & Score (\%) \\
 \midrule

\textbf{iAmTime} & 0.646 & 11.170 & 0.484 & 11.260 & 71.6 & 46.8 \\
\textbf{iAmTime-NoMeta} & 0.650 & 13.340 & 0.487 & 12.740 & 66.2 & 46.4 \\
\textbf{iAmTime-NoExmp} & 0.650 & 13.190 & 0.487 & 12.910 & 66.8 & 46.4 \\
Toto-2.0-1B & 0.676 & 14.200 & 0.508 & 13.410 & 64.2 & 44.4 \\
Toto-2.0-2.5B & 0.675 & 14.490 & 0.508 & 13.470 & 65.9 & 44.4 \\
Toto-2.0-313m & 0.679 & 15.130 & 0.510 & 14.080 & 61.5 & 44.2 \\
Chronos-2 & 0.645 & 13.520 & 0.485 & 14.420 & 67.8 & 47.3 \\
\textbf{iAmTime-NoToks-NoExmp} & 0.653 & 15.360 & 0.489 & 15.440 & 60.3 & 46.2 \\
TiRex-2 & 0.663 & 16.080 & 0.505 & 16.610 & 62.0 & 45.5 \\
Toto-2.0-22m & 0.694 & 17.940 & 0.523 & 17.350 & 52.1 & 42.9 \\
TabPFN-TS-3 & 0.694 & 21.250 & 0.512 & 20.010 & 43.5 & 43.1 \\
CITRAS-FM & 0.707 & 22.060 & 0.540 & 22.310 & 40.0 & 41.2 \\
Toto-2.0-4m & 0.720 & 22.210 & 0.553 & 22.870 & 37.6 & 40.7 \\
Moirai-2.0 & 0.720 & 23.940 & 0.552 & 23.520 & 34.4 & 40.2 \\
\textbf{iAmTime-NoToks-WExmp} & 0.721 & 24.140 & 0.586 & 24.260 & 33.0 & 38.6 \\
Chronos-Bolt Base & 0.735 & 25.540 & 0.568 & 25.910 & 29.1 & 38.9 \\
Seasonal Naive & 1.000 & 33.560 & 1.000 & 34.460 & 3.3 & 0.0 \\

\bottomrule
\end{tabular}
\end{table}

\begin{table}[t]
\centering
\caption{
Metrics sorted with respect to CRPS Rank on the GIFT-Eval dataset.
\textbf{iAmTime} is the base version trained using Meta-training tasks and examples.
\textbf{iAmTime-NoExmp} ablates the examples (0 examples) while training (and removes the other meta-training task).
\textbf{iAmTime-NoToks-NoExmp} ablates the use of semantic tokens and examples while training.
\textbf{iAmTime-NoToks-WExmp} ablates the use of semantic tokens but keeps the examples while training.
\textbf{iAmTime-NoMeta} uses only forecasting tasks and forecasting examples to train the model.
}
\label{tab:ablation_gift}
\begin{tabular}{lrrrr}
\toprule
 & MASE & MASE Rank & CRPS & CRPS Rank \\
Model &  &  &  &  \\
\midrule
\textbf{iAmTime} & 0.680 & 4.443 & 0.465 & 4.546 \\
\textbf{iAmTime-NoMeta} & 0.686 & 5.351 & 0.468 & 5.134 \\
\textbf{iAmTime-NoExmp} & 0.691 & 6.299 & 0.471 & 5.876 \\
Toto-2.0-1B & 0.699 & 6.103 & 0.478 & 6.423 \\
Toto-2.0-313m & 0.703 & 6.381 & 0.481 & 6.557 \\
Chronos-2 & 0.698 & 6.464 & 0.485 & 6.928 \\
TiRex-2 & 0.697 & 7.402 & 0.478 & 7.082 \\
\textbf{iAmTime-NoToks-NoExmp} & 0.710 & 7.503 & 0.489 & 7.237 \\
TimesFM-2.5 & 0.705 & 7.680 & 0.490 & 7.732 \\
Moirai-2.0 & 0.728 & 9.814 & 0.516 & 9.948 \\
Toto 1.0 & 0.750 & 10.278 & 0.517 & 9.948 \\
\textbf{iAmTime-NoToks-WExmp} & 0.756 & 10.278 & 0.569 & 10.876 \\
Chronos-Bolt Base & 0.808 & 10.278 & 0.574 & 11.000 \\
TabPFN-TS & 0.771 & 11.784 & 0.544 & 11.289 \\
Seasonal Naive & 1.000 & 14.423 & 1.000 & 14.753 \\

\bottomrule
\end{tabular}
\end{table}

\subsection{Structural Ablations}

\paragraph{Semantic Token Stream.}
We consider two variants that remove semantic role tokens. 
iAmTime-NoToks-NoExmp removes both the Token Stream and demonstrations and is trained only for forecasting; it therefore remains a capable forecaster but has no mechanism for demonstration-conditioned adaptation. 
iAmTime-NoToks-WExmp retains examples and multi-task training but removes the explicit token structure separating targets, covariates, demonstrated outputs, and queries. 
Its substantial degradation shows that providing examples alone is insufficient: without explicit role and boundary information, the model must infer prompt structure, making cross-example information less reliable.

\paragraph{Per-Example Fusion (NoPEF).}
NoPEF removes the Per-Example Fusion attention operating on the Patch Stream, preventing direct integration of target and covariate representations within each example. 
This causes one of the largest forecasting degradations (fev skill $46.8\!\rightarrow\!37.8$ and GIFT CRPS $0.465\!\rightarrow\!0.583$), showing that PEF is a primary forecasting mechanism. 
Its degradation on the non-forecasting tasks further indicates that coherent within-example representations are also important before information can be transferred across demonstrations.

\subsection{Task-Adaptation Mechanism Ablations}

\paragraph{Cross-Example Attention (NoCEA).}
NoCEA removes token-level Cross-Example Attention, which transfers information summarized from support examples into the query representation. 
Forecasting degrades only moderately, whereas classification accuracy falls from $0.813$ to $0.263$, imputation error increases substantially, and anomaly detection also deteriorates. 
This disproportionate effect confirms that Cross-Example Attention is principally responsible for extracting the mapping encoded by demonstrations rather than for basic forecasting capacity.

\paragraph{Token Write (NoTWrite).}
NoTWrite removes the FiLM-based Token Write operation that injects the inferred task representation back into query patches. 
The resulting pattern closely mirrors NoCEA: forecasting remains relatively competitive, but classification, imputation, and anomaly detection degrade sharply. 
Together, NoCEA and NoTWrite identify a clear information pathway: Cross-Example Attention retrieves the demonstrated mapping, while Token Write applies that information to the query representation.

\paragraph{Shared token roles (SharedToks).}
SharedToks ties the role-token embeddings while retaining the tokens and their attention masks. 
Performance decreases across all tasks, but substantially less than when cross-example communication or Token Write is removed. 
Distinct token identities therefore improve separation of targets, covariates, historical inputs, demonstrated outputs, and future regions, while the attention structure itself retains part of the task-adaptation capability.

\subsection{Output Specialization Ablation}

\paragraph{Uniform MoE routing (UniMoE).}
UniMoE replaces learned task-conditioned expert routing with uniform expert weights while preserving the same experts. 
Its relatively small but consistent degradation across forecasting and non-forecasting tasks shows that adaptive routing improves output specialization, but is not the primary source of ICL behavior. 
The main task-adaptation capability instead arises upstream from demonstration-conditioned training and the Token Stream mechanisms.

\subsection{Task-Family and Curriculum Ablations}

\paragraph{Task-family removals.}
We separately remove classification (NoCls), imputation (NoImp), anomaly-detection (NoAD), and synthetic-classification (NoSynthCls) episodes. 
The resulting failures are strongly task-selective: NoCls primarily reduces classification accuracy, NoImp substantially increases imputation error, and NoAD sharply reduces anomaly-detection F1, while forecasting remains nearly unchanged. 
NoSynthCls also substantially reduces classification performance, demonstrating the value of synthetic episodes for learning transferable episode-level mappings. 
These results show that poor non-forecasting performance is not simply due to globally weaker checkpoints; each task family contributes specialized behavior while sharing the same architecture and inference interface.

\paragraph{Curriculum learning (NoCurriculum).}
Removing the curriculum while retaining multi-task training produces the largest overall degradation, reducing fev skill from $46.8$ to $30.2$ and worsening every non-forecasting task. 
The staged curriculum is therefore important for stabilizing optimization across heterogeneous task formats and preventing difficult instruction-conditioned episodes from interfering with the acquisition of fundamental forecasting and representation capabilities.

Overall, the ablations reveal: 
Per-Example Fusion and curriculum learning are the dominant contributors to forecasting quality; Cross-Example Attention and Token Write have smaller forecasting effects but are essential for transmitting demonstrated task mappings; distinct semantic tokens support reliable role separation; and adaptive MoE routing provides consistent specialization benefit. 
The task-family ablations show that individual capabilities are learned selectively rather than arising from an improvement in model quality. 
This behavior is consistent with the broader MetaICL perspective \citep{min2022metaicl}: effective ICL arises from training on structured demonstration-query episodes that align training and inference interfaces, allowing task adaptation to be amortized into a forward pass.

\subsection{Effect of Structural and Training Ablations on Task Adaptation}
\label{sec:classification_ablation}

\begin{table}[t]
\centering
\caption{
Classification ablation results on the overall UCR suite under the in-context classification protocol.
Relative drop is computed with respect to full iAmTime accuracy.
}
\small
\setlength{\tabcolsep}{5pt}
\renewcommand{\arraystretch}{1.12}
\begin{tabular}{lcccc}
\toprule
\textbf{Model variant} & \textbf{Acc.} & \textbf{Acc. Std.} & \textbf{Macro-F1} & \textbf{Rel. drop} \\
\midrule
iAmTime & 0.813 & 0.198 & 0.775 & - \\
iAmTime-NoToks-WExmp & 0.502 & 0.341 & 0.521 & $-38.2\%$ \\
iAmTime-NoExmp & 0.238 & 0.187 & 0.184 & $-70.7\%$ \\
iAmTime-NoMeta & 0.224 & 0.181 & 0.176 & $-72.4\%$ \\
iAmTime-NoToks-NoExmp & 0.207 & 0.169 & 0.158 & $-74.5\%$ \\
\bottomrule
\end{tabular}
\label{tab:classification_ablation_ucr}
\end{table}

We further isolate these effects on UCR classification, where the model must infer an episode-local label mapping from support examples without a task-specific classification head. 
Table~\ref{tab:classification_ablation_ucr} shows that full iAmTime achieves an average accuracy of $0.813$ and Macro-F1 of $0.775$. 
Variants without demonstrations or non-forecasting meta-training approach near-chance performance because they never learn to infer episode-specific label mappings. 
Removing semantic tokens retains partial adaptation when classification demonstrations remain available, whereas removing Cross-Example Attention or Token Write largely eliminates the ability to transfer those mappings to the query.

These results reinforce the distinction between forecasting and task adaptation: strong forecasting can persist without demonstrations, but classification requires the appropriate task-training distribution, demonstrations that define the episode-specific mapping, and architectural mechanisms that transfer this information to the query. 
The full model combines these components to perform classification through the same generative interface used for forecasting.

\subsection{Mechanism Transfer to Chronos-Bolt}
\label{sec:chronos_bolt_transfer}

\begin{table}[t]
\centering
\caption{
Transfer of the proposed ICL architecture and meta-training mechanisms to a Chronos-Bolt Base backbone.
}
\label{tab:chronos_bolt_transfer}
\scriptsize
\setlength{\tabcolsep}{2pt}
\renewcommand{\arraystretch}{1.12}

\resizebox{\columnwidth}{!}{
\begin{tabular}{llccccc}
\toprule
Variant
& Description
& \makecell[c]{Forecasting\\fev Skill\\Score (\%) $\uparrow$}
& \makecell[c]{Forecasting\\GIFT\\CRPS $\downarrow$}
& \makecell[c]{Classification\\UCR ICL\\Acc. (\%) $\uparrow$}
& \makecell[c]{Imputation\\Block-8\\MSE/MAE $\downarrow$}
& \makecell[c]{Anomaly\\Detection\\F1 $\uparrow$}
\\
\midrule

CB
& Original Chronos-Bolt
& 38.9
& 0.574
& n/a
& n/a
& n/a
\\

CB-M
& Matched training without ICL mechanism
& 40.1
& 0.555
& 67.37
& 0.3420 / 0.4111
& 76.76
\\

CB-ICL-0
& ICL checkpoint, zero inference examples
& 41.7
& 0.534
& 21.8
& 0.3261 / 0.2984
& 77.83
\\

CB-ICL-4
& ICL checkpoint, four examples
& 43.2
& 0.512
& 69.35
& 0.2002 / 0.2671
& 85.16
\\

\midrule

$\Delta$
& Improvement: CB-ICL-4 vs.\ CB-M
& +3.1 pp
& 7.7\% lower
& +1.98
& 41.5\% / 35.0\% lower
& +8.4
\\

\bottomrule
\end{tabular}
}
\end{table}

To test the transferability of the proposed mechanisms beyond iAmTime, we augment Chronos-Bolt Base with additional encoder and attention components following the iAmTime design and apply the same meta-training paradigm. Table~\ref{tab:chronos_bolt_transfer} compares the original model (CB), matched training without ICL (CB-M), and the ICL-trained checkpoint with zero or four inference examples (CB-ICL-0/4).

The pattern transfers clearly: CB-ICL-4 improves over CB-M by $+3.1$ fev skill points, $7.7\%$ lower GIFT CRPS, $41.5\%/35.0\%$ lower imputation MSE/MAE, and $+8.4$ anomaly F1. Comparing CB-ICL-0 with CB-ICL-4 isolates the effect of demonstrations, yielding $+47.55$ classification points and $+7.33$ anomaly F1. These results suggest that the architecture and instruction-conditioned meta-training are transferable, while demonstrations provide an additional task-specification benefit.

\subsection{Robustness to Distribution Shift during Inference}

To study this, we train the model by selecting subset of data belonging to 
domain
$C_T$, and perform inference on a non-overlapping subset of domains $C_I$ such that $C_T \cap C_I = \emptyset$.
While doing inference, we provide ICL examples from classes $C_E$ in three ways:
\begin{enumerate}
    \item $C_E \subset C_I$,
    \item $C_E \subset C_T$, and
    \item repeat 1 but perform perturbations on the example time-series.
\end{enumerate}

\begin{table}[t]
\centering
\caption{Results on subset of fev-bench observing the robustness to distribution shift}
\label{tab:icl_distribution_shift_ablation}
\begin{tabular}{r|cc}
\toprule
Inference & Win Rate (\%) & Skill Score (\%) \\
ICL Variant &  &  \\
\midrule
$C_E \subset C_I$ & \textbf{71.3} & 50.1 \\
$C_E \subset C_T$ & \underline{69.0} & 50.8 \\
$C_E \subset C_I$ + random perturbations & 66.2 & 46.7 \\
\bottomrule
\end{tabular}
\end{table}

We do this by training the full iAmTime model, using all meta-training 
task-classes.
We let $C_T = $ all domains other than $\{$"Climate", "Healthcare"$\}$, and
$C_I =\{$"Climate", "Healthcare"$\}$ domains.
The pre-training is done by sampling the pre-training datasets from domains $C_T$
(Tables \ref{tab:datasets_train_chronos}, \ref{tab:datasets_gift_eval_pre}).

The model is evaluated on forecasting-task by creating
queries $\mathcal{Q}$ from $C_I$ in the benchmark datasets
(Tables \ref{tab:dataset_eval_fev}, \ref{tab:dataset_eval_gift_eval}).
For the $C_E \subset C_I$ variant, examples $\{\mathcal{E}_i\}_{i=1}^N$ 
are constructed by randomly selecting
time-series from $C_I$.
For $C_E \subset C_T$, the examples are sampled from $C_T$.
For the third variant, the same $(\mathcal{Q}, \{\mathcal{E}_i\}_{i=1}^N)$
pairs are used from the first variant and perturbations are made on all the
examples $\{\mathcal{E}_i\}_{i=1}^N$, by randomly applying 
the "Time-dependent transformation" functions
from Section \ref{sec:augmentation_construction}.

Table \ref{tab:icl_distribution_shift_ablation} evaluates robustness to distribution shift during inference by varying the source of in-context examples provided to the model. When example demonstrations are drawn from the same unseen target-domain class set as the query ($C_E \subset C_I$), the model achieves the highest win rate, indicating that in-context examples effectively anchor the model to the test-time data distribution. 
When examples are instead drawn from the training-domain classes ($C_E \subset C_T$), performance degrades very slightly but remains strong, demonstrating that the model can still transfer learned forecasting strategies across domains through contextual adaptation. 
Introducing random perturbations to in-domain examples further reduces performance, confirming that the quality and distributional alignment of demonstrations directly influence the effectiveness of in-context task inference. 
These results suggest that the model uses in-context examples primarily as distributional and functional references rather than relying on memorized domain-specific parameters, enabling robust zero-shot generalization under \emph{moderate} distribution shifts.

\subsection{Mechanistic Analysis of Demonstration-Conditioned Adaptation}
\label{app:mechanistic_analysis}


\begin{figure}[ht]
    \captionsetup[figure]{font=small,skip=5pt}
    \captionsetup[subfigure]{labelformat=parens,labelsep=none,font=small,skip=3pt}

    \centering
    \begin{minipage}{0.92\linewidth}
    \centering
    \begin{subfigure}[b]{0.65\linewidth}
        \centering
        \includegraphics[width=\linewidth]{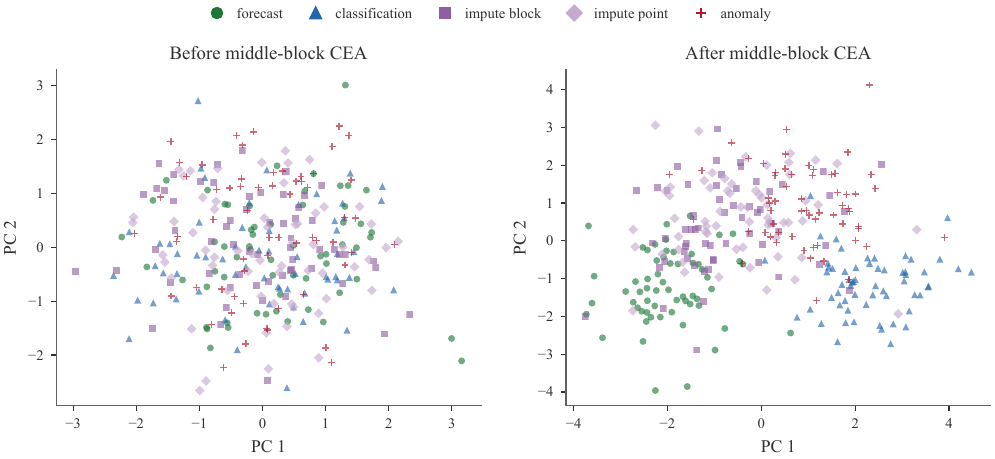}
        \caption{}\label{fig:probe_repr}
    \end{subfigure}\hfill
    \begin{subfigure}[b]{0.33\linewidth}
        \centering
        \includegraphics[width=\linewidth]{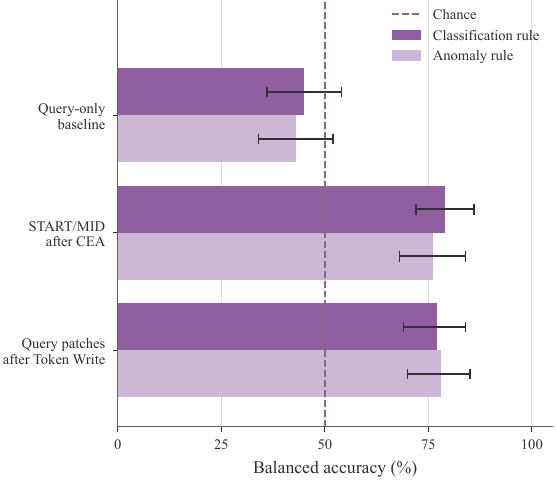}
        \caption{}\label{fig:probe_rules}
    \end{subfigure}

    \medskip
    \begin{subfigure}[b]{0.75\linewidth}
        \centering
        \includegraphics[width=\linewidth]{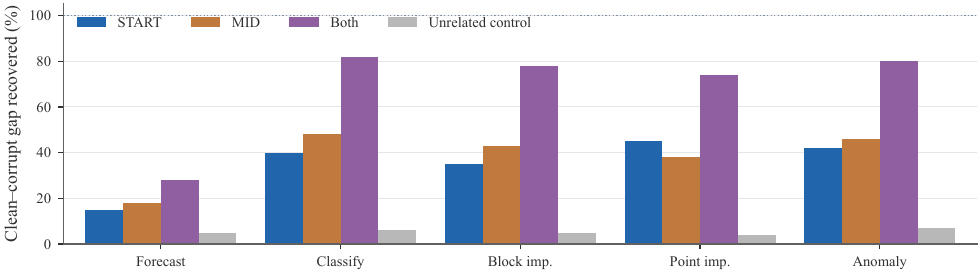}
        \caption{}\label{fig:probe_restoration}
    \end{subfigure}
    \end{minipage}
    \caption{\textbf{Representation organization, rule decodability, and task-state restoration.}
    (a) PCA projections before and after middle-block CEA, colored by task.
    (b) Held-out rule-probe balanced accuracy with confidence intervals;
    the dashed line marks $50\%$ chance accuracy. Probes measure rule
    decodability rather than downstream task performance.
    (c) Recovery after inserting clean query \texttt{[START]} and/or
    \texttt{[MID]} states into corrupted-demonstration runs before Token
    Write. Recovery is $100R\%$ (Eq.~\eqref{eq:task_state_recovery});
    the dashed line marks full recovery.}
    \label{fig:representations_restoration}
\end{figure}

We complement the task-adaptation and output-specialization ablations with
controlled analyses of the frozen iAmTime checkpoint on synthetic episodes
covering forecasting, classification, contiguous-block imputation, random-point
imputation, and anomaly detection. These analyses examine how demonstrated
mappings are represented in the Token Stream, which attention connections
influence predictions, and how task conditioning affects the decoder. Model
parameters remain fixed throughout; only the external representation probes
are fitted and examined.

\paragraph{Representation organization and held-out rule probes.}
Figure~\ref{fig:representations_restoration}(\subref{fig:probe_repr}) compares PCA projections of query representations
immediately before and after the middle-block Cross-Example Attention (CEA)
operation. The post-CEA representations exhibit clearer task-dependent
organization, while contiguous-block and random-point imputation remain
overlapping. To complement this descriptive visualization, we probe the
demonstrated classification and anomaly-detection rules on held-out episodes
(Figure~\ref{fig:representations_restoration}(\subref{fig:probe_rules})).
Classification-rule balanced accuracy increases from $45\%$ for query-only
representations to $79\%$ for the updated \texttt{[START]}/\texttt{[MID]}
states and $77\%$ for query patches after Token Write. The corresponding
anomaly-rule accuracies are $43\%$, $76\%$, and $78\%$. Both query-only
confidence intervals include the $50\%$ chance level, whereas all four
post-conditioning intervals lie above it. These results indicate that rule
information is accessible in both the Token Stream and the conditioned
Patch Stream; they do not require complete separation between task clusters.

\paragraph{Causal restoration of task states.}
We compare correct demonstrations (clean) with altered demonstrated mappings
(corrupt), keeping the query fixed. During a corrupted run, we restore
query \texttt{[START]} state, \texttt{[MID]} state, or both from the clean run
after CEA and before Token Write, and continue forward pass. For query
loss $L$, recovery is the fraction of the clean-corrupt loss gap closed:
\begin{equation}
    R = \frac{L_{\mathrm{corrupt}}-L_{\mathrm{restored}}}
             {L_{\mathrm{corrupt}}-L_{\mathrm{clean}}}.
    \label{eq:task_state_recovery}
\end{equation}
Figure~\ref{fig:representations_restoration}(\subref{fig:probe_restoration}) reports $100R$ as a percentage. Joint
restoration recovers $82\%$, $78\%$, $74\%$, and $80\%$ of the gap for
classification, block imputation, point imputation, and anomaly detection,
respectively, and $28\%$ for forecasting. It exceeds either single-token
intervention in every setting, and unrelated-state controls recover only
$4$-$7\%$. This supports a causal contribution of the
restored query states under the intervention, complementing the
\texttt{NoCEA} and \texttt{NoTWrite} ablations. Because these states also
affect subsequent computations and decoder routing, the intervention does
not isolate FiLM alone.

\begin{figure}[tp]
    \captionsetup[figure]{font=small,skip=5pt}
    \captionsetup[subfigure]{labelformat=parens,labelsep=none,font=small,skip=3pt}

    \centering
    \begin{minipage}{0.99\linewidth}
        \centering
        \begin{minipage}[b]{0.78\linewidth}
            \centering
            \begin{subfigure}[b]{\linewidth}
                \centering
                \includegraphics[width=\linewidth]{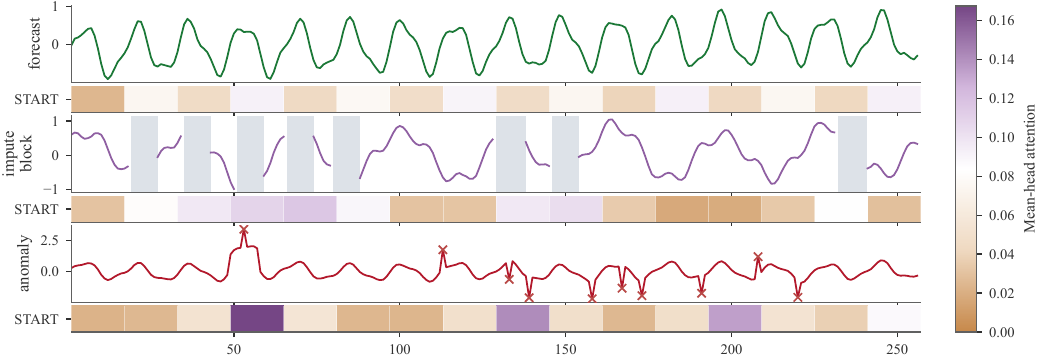}
                \caption{}\label{fig:probe_attention}
            \end{subfigure}

            \medskip
            \begin{subfigure}[b]{\linewidth}
                \centering
                \includegraphics[width=\linewidth]{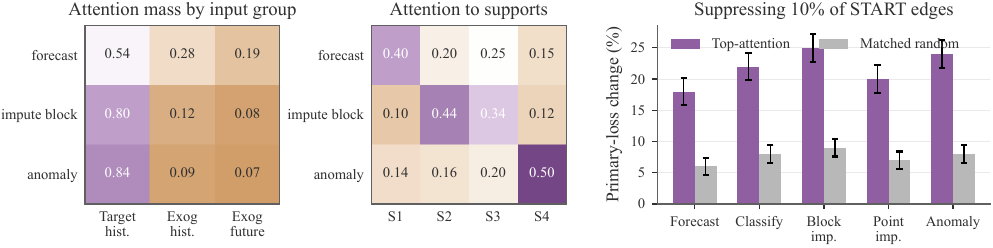}
                \caption{}\label{fig:probe_attention_summary}
            \end{subfigure}
        \end{minipage}\hfill
        \begin{subfigure}[b]{0.21\linewidth}
            \centering
            \includegraphics[width=\linewidth]{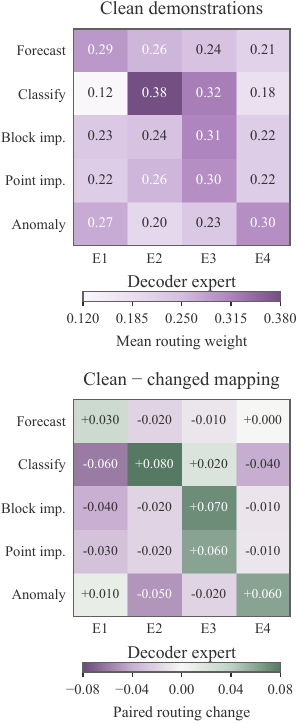}
            \caption{}\label{fig:probe_routing}
        \end{subfigure}
    \end{minipage}
    \caption{\textbf{Attention allocation, functional sensitivity, and decoder routing.}
    (a) Representative series aligned with mean-head query \texttt{[START]}
    Token Read attention over 16-step patches.
    (b) Attention mass by input group, retrieval weights over support examples,
    and primary-loss increases (\%) after suppressing the top-attended $10\%$
    of \texttt{[START]} connections or matched-random connections. Suppression
    is evaluated across all five tasks.
    (c) Mean decoder routing weights under clean demonstrations (top) and
    paired clean-minus-changed-mapping differences (bottom). Expert indices
    refer to the same trained decoder; all tasks share the four experts.}
    \label{fig:attention_routing}
\end{figure}

\paragraph{Attention allocation and functional sensitivity.}
Figure~\ref{fig:attention_routing}(\subref{fig:probe_attention}) aligns query \texttt{[START]} Token Read
attention with the input series at 16-step patch resolution.
Figure~\ref{fig:attention_routing}(\subref{fig:probe_attention_summary})
summarizes input-group attention, cross-example retrieval, and suppression
sensitivity. In the displayed cases, target history
receives approximately $54\%$ of attention mass for forecasting, compared with
$80\%$ for block imputation and $84\%$ for anomaly detection, with the remainder
allocated to covariates. Support attention is also non-uniform. Suppressing
the top-attended $10\%$ of \texttt{[START]} connections increases primary
loss by $18\%$, $22\%$, $25\%$, $20\%$, and $24\%$ for forecasting,
classification, block imputation, point imputation, and anomaly detection,
respectively. A matched random $10\%$ suppression of \texttt{[START]} 
attention yields smaller increases of
$6\%$, $8\%$, $9\%$, $7\%$, and $8\%$ in the same order. This intervention
supports a functional contribution from the selected connections, beyond
the alignment of attention maps.

\paragraph{Task-conditioned expert mixtures.}
Figure~\ref{fig:attention_routing}(\subref{fig:probe_routing}) shows shared, task-dependent decoder mixtures.
Forecasting weights are relatively close to uniform, whereas classification
assigns approximately $70\%$ of its mass to experts 2 and 3. The two
imputation settings have similar mixtures, and each task uses all four
experts. Paired routing changes for classification, imputation, and anomaly
detection further show sensitivity to the demonstrated mapping. These observations are
consistent with adaptive output specialization, complementing the modest
degradation under \texttt{UniMoE} in Appendix~J.4, without implying a
one-to-one correspondence between experts and tasks.


\clearpage
\section{Societal Impact}
\label{sec:societal_impact}

iAmTime may reduce the cost and latency of adapting time-series models across domains such as energy, supply chains, healthcare, transportation, and finance by allowing tasks to be specified through examples rather than retraining. This could make forecasting and related time-series analysis more reusable and accessible in settings with limited labeled data or modeling infrastructure. However, forecasts, anomaly signals, and classification outputs may be unreliable under distribution shift, poor data quality, or misleading in-context demonstrations. In high-stakes settings, these outputs should be treated as decision-support signals rather than automated decisions, with validation on representative data, uncertainty assessment, drift monitoring, and human oversight. Future work should further study robustness, calibration, fairness, and safeguards for example selection.

\section{Code}
\label{sec:code}
The code repository for the model will be made available on GitHub. The model weights, and synthetic data generation scripts will be released alongside the code.

\clearpage

\section{Dataset Details}
\label{sec:datasets}

\subsection{Training Data}

\begin{table}[ht!]
\centering
\caption{
    List of 
    training datasets used from Chronos pre-training corpus, published by
    \cite{ansari2024chronos}
}
\label{tab:datasets_train_chronos}
\begin{tabular}{llrr}
\toprule
\textbf{Name} & \textbf{Domain} & \textbf{\# Series} & \textbf{Avg. Length} \\
\midrule
Mexico City Bikes & Mobility / Transport & 494 & 78{,}313 \\
Brazilian Cities Temperature & Weather / Climate & 12 & 757 \\
Solar (5 Min.) & Energy & 5{,}166 & 105{,}120 \\
Solar (Hourly) & Energy & 5{,}166 & 105{,}120 \\
Spanish Energy and Weather & Energy / Weather & 66 & 35{,}064 \\
Taxi (Hourly) & Mobility / Transport & 2{,}428 & 739 \\
USHCN & Weather / Climate & 6{,}090 & 38{,}653 \\
Weatherbench (Hourly) & Weather / Climate & 225{,}280 & 350{,}639 \\
Weatherbench (Daily) & Weather / Climate & 225{,}280 & 14{,}609 \\
Weatherbench (Weekly) & Weather / Climate & 225{,}280 & 2{,}087 \\
Wiki Daily (100k) & Web / Information & 100{,}000 & 2{,}741 \\
Wind Farms (Hourly) & Energy & 100{,}000 & 8{,}514 \\
Wind Farms (Daily) & Energy & 100{,}000 & 354 \\
Electricity (15 Min.) & Energy & 370 & 113{,}341 \\
Electricity (Hourly) & Energy & 321 & 26{,}304 \\
Electricity (Weekly) & Energy & 321 & 156 \\
KDD Cup 2018 & Energy & 270 & 10{,}897 \\
London Smart Meters & Energy & 5{,}560 & 29{,}951 \\
M4 (Daily) & Business / Economics & 4{,}227 & 2{,}371 \\
M4 (Hourly) & Business / Economics & 414 & 901 \\
M4 (Monthly) & Business / Economics & 48{,}000 & 234 \\
M4 (Weekly) & Business / Economics & 359 & 1{,}035 \\
Pedestrian Counts & Mobility / Urban & 66 & 47{,}459 \\
Rideshare & Mobility / Transport & 2{,}340 & 541 \\
Taxi (30 Min.) & Mobility / Transport & 2{,}428 & 1{,}478 \\
Temperature--Rain & Weather / Climate & 32{,}072 & 725 \\
Uber TLC (Hourly) & Mobility / Transport & 262 & 4{,}344 \\
Uber TLC (Daily) & Mobility / Transport & 262 & 181 \\
\bottomrule
\end{tabular}
\end{table}

\begin{table}[ht!]
\centering
\caption{Subset of pre-training datasets used from GiftEvalPretrain, published by \cite{aksu2024gift}}
\label{tab:datasets_gift_eval_pre}
\begin{tabular}{llrr}
\toprule
\textbf{Name} & \textbf{Domain} & \textbf{\# Series} & \textbf{Avg. Length} \\
\midrule
azure vm traces 2017 & Cloud / Systems & 159{,}472 & 5{,}553 \\
borg cluster data 2011 & Cloud / Systems & 143{,}386 & 3{,}749 \\
bdg-2 panther & Energy & 105 & 8{,}760 \\
bdg-2 fox & Energy & 135 & 17{,}219 \\
bdg-2 rat & Energy & 280 & 16{,}887 \\
bdg-2 bear & Energy & 91 & 16{,}289 \\
lcl & Energy & 713 & 13{,}385 \\
smart & Energy & 5 & 19{,}142 \\
ideal & Energy & 217 & 5{,}785 \\
sceaux & Energy & 1 & 34{,}223 \\
borealis & Energy & 15 & 5{,}551 \\
buildings 900k & Energy & 1{,}795{,}256 & 8{,}761 \\
largest 2017 & Climate & 8{,}196 & 105{,}120 \\
largest 2018 & Climate & 8{,}428 & 105{,}120 \\
largest 2019 & Climate & 8{,}600 & 105{,}120 \\
largest 2020 & Climate & 8{,}561 & 105{,}408 \\
largest 2021 & Climate & 8{,}548 & 105{,}120 \\
PEMS03 & Transport & 358 & 26{,}208 \\
PEMS04 & Transport & 307 & 16{,}992 \\
PEMS07 & Transport & 883 & 28{,}224 \\
PEMS08 & Transport & 170 & 17{,}856 \\
PEMS BAY & Transport & 325 & 52{,}128 \\
LOS LOOP & Transport & 207 & 34{,}272 \\
BEIJING SUBWAY 30MIN & Transport & 276 & 1{,}572 \\
SHMETRO & Transport & 288 & 8{,}809 \\
HZMETRO & Transport & 80 & 2{,}377 \\
Q-TRAFFIC & Transport & 45{,}148 & 5{,}856 \\
subseasonal & Climate & 862 & 16{,}470 \\
subseasonal precip & Climate & 862 & 11{,}323 \\
wind power & Energy & 1 & 7{,}397{,}147 \\
solar power & Energy & 1 & 7{,}397{,}222 \\
kaggle web traffic weekly & Web & 14{,}563 & 114 \\
kdd2022 & Energy & 134 & 35{,}280 \\
godaddy & Web & 3{,}135 & 41 \\
favorita sales & Retail & 111{,}840 & 1{,}244 \\
china air quality & Environment & 437 & 13{,}133 \\
beijing air quality & Environment & 12 & 35{,}064 \\
residential load power & Energy & 271 & 538{,}725 \\
residential pv power & Energy & 233 & 537{,}935 \\
cdc fluview ilinet & Healthcare & 75 & 852 \\
cdc fluview who & Healthcare & 74 & 564 \\
\bottomrule
\end{tabular}
\end{table}

\clearpage

\subsection{Evaluation Data}
\begin{table*}[h!]
\centering
\scriptsize
\caption{
Benchmark datasets dataset summary of GIFT-Eval \cite{aksu2024gift}. 
The benchmark provides various settings for evaluating horizons (H) 
in terms of short/mid/long term forecasts.
Each dataset is evaluated over W windows.
}
\label{tab:dataset_eval_gift_eval}
\resizebox{\textwidth}{!}{%
\begin{tabular}{llc r rrr c rr rr rr}
\toprule
\textbf{Dataset} & \textbf{Domain} & \textbf{Freq.} & \textbf{\# Series} &
\multicolumn{3}{c}{\textbf{Series Length}} & \textbf{\# Target} &
\multicolumn{2}{c}{\textbf{Short-term}} &
\multicolumn{2}{c}{\textbf{Med-term}} &
\multicolumn{2}{c}{\textbf{Long-term}} \\
\cmidrule(lr){5-7}\cmidrule(lr){9-10}\cmidrule(lr){11-12}\cmidrule(lr){13-14}
& & & & \textbf{Avg} & \textbf{Min} & \textbf{Max} & \textbf{Variate} &
\textbf{H} & \textbf{W} &
\textbf{H} & \textbf{Windows} &
\textbf{H} & \textbf{W} \\
\midrule
Jena Weather & Nature & 10T & 1 & 52,704 & 52,704 & 52,704 & 21 & 48 & 20 & 480 & 1 & 720 & 8 \\
Jena Weather & Nature & H & 1 & 8,784 & 8,784 & 8,784 & 21 & 48 & 19 & 480 & 2 & 720 & 2 \\
Jena Weather & Nature & D & 1 & 366 & 366 & 366 & 21 & 30 & 2 &  &  &  &  \\
BizITObs - Application & Web/CloudOps & 10S & 1 & 8,834 & 8,834 & 8,834 & 2 & 60 & 15 & 600 & 2 & 900 & 1 \\
BizITObs - Service & Web/CloudOps & 10S & 21 & 8,835 & 8,835 & 8,835 & 2 & 60 & 15 & 600 & 2 & 900 & 1 \\
BizITObs - L2C & Web/CloudOps & 5T & 1 & 31,968 & 31,968 & 31,968 & 7 & 48 & 20 & 480 & 7 & 720 & 5 \\
BizITObs - L2C & Web/CloudOps & H & 1 & 2,664 & 2,664 & 2,664 & 7 & 48 & 6 & 480 & 1 & 720 & 1 \\
Bitbrains - Fast Storage & Web/CloudOps & 5T & 1,250 & 8,640 & 8,640 & 8,640 & 2 & 48 & 18 & 480 & 2 & 720 & 2 \\
Bitbrains - Fast Storage & Web/CloudOps & H & 1,250 & 721 & 721 & 721 & 2 & 48 & 2 &  &  &  &  \\
Bitbrains - rnd & Web/CloudOps & 5T & 500 & 8,640 & 8,640 & 8,640 & 2 & 48 & 18 & 480 & 2 & 720 & 2 \\
Bitbrains - rnd & Web/CloudOps & H & 500 & 720 & 720 & 720 & 2 & 48 & 2 &  &  &  &  \\
Restaurant & Sales & D & 807 & 358 & 67 & 478 & 1 & 30 & 1 &  &  &  &  \\
ETT1 & Energy & 15T & 1 & 69,680 & 69,680 & 69,680 & 7 & 48 & 20 & 480 & 15 & 720 & 10 \\
ETT1 & Energy & H & 1 & 17,420 & 17,420 & 17,420 & 7 & 48 & 20 & 480 & 4 & 720 & 3 \\
ETT1 & Energy & D & 1 & 725 & 725 & 725 & 7 & 30 & 3 &  &  &  &  \\
ETT1 & Energy & W-THU & 1 & 103 & 103 & 103 & 7 & 8 & 2 &  &  &  &  \\
ETT2 & Energy & 15T & 1 & 69,680 & 69,680 & 69,680 & 7 & 48 & 20 & 480 & 15 & 720 & 10 \\
ETT2 & Energy & H & 1 & 17,420 & 17,420 & 17,420 & 7 & 48 & 20 & 480 & 4 & 720 & 3 \\
ETT2 & Energy & D & 1 & 725 & 725 & 725 & 7 & 30 & 3 &  &  &  &  \\
ETT2 & Energy & W-THU & 1 & 103 & 103 & 103 & 7 & 8 & 2 &  &  &  &  \\
Loop Seattle & Transport & 5T & 323 & 105,120 & 105,120 & 105,120 & 1 & 48 & 20 & 480 & 20 & 720 & 15 \\
Loop Seattle & Transport & H & 323 & 8,760 & 8,760 & 8,760 & 1 & 48 & 19 & 480 & 2 & 720 & 2 \\
Loop Seattle & Transport & D & 323 & 365 & 365 & 365 & 1 & 30 & 2 &  &  &  &  \\
SZ-Taxi & Transport & 15T & 156 & 2,976 & 2,976 & 2,976 & 1 & 48 & 7 & 480 & 1 & 720 & 1 \\
SZ-Taxi & Transport & H & 156 & 744 & 744 & 744 & 1 & 48 & 2 &  &  &  &  \\
M\_DENSE & Transport & H & 30 & 17,520 & 17,520 & 17,520 & 1 & 48 & 20 & 480 & 4 & 720 & 3 \\
M\_DENSE & Transport & D & 30 & 730 & 730 & 730 & 1 & 30 & 3 &  &  &  &  \\
Solar & Energy & 10T & 137 & 52,560 & 52,560 & 52,560 & 1 & 48 & 20 & 480 & 11 & 720 & 8 \\
Solar & Energy & H & 137 & 8,760 & 8,760 & 8,760 & 1 & 48 & 19 & 480 & 2 & 720 & 2 \\
Solar & Energy & D & 137 & 365 & 365 & 365 & 1 & 30 & 2 &  &  &  &  \\
Solar & Energy & W-FRI & 137 & 52 & 52 & 52 & 1 & 8 & 1 &  &  &  &  \\
Hierarchical Sales & Sales & D & 118 & 1,825 & 1,825 & 1,825 & 1 & 30 & 7 &  &  &  &  \\
Hierarchical Sales & Sales & W-WED & 118 & 260 & 260 & 260 & 1 & 8 & 4 &  &  &  &  \\
M4 Yearly & Econ/Fin & A-DEC & 22,974 & 37 & 19 & 284 & 1 & 6 & 1 &  &  &  &  \\
M4 Quarterly & Econ/Fin & Q-DEC & 24,000 & 100 & 24 & 874 & 1 & 8 & 1 &  &  &  &  \\
M4 Monthly & Econ/Fin & M & 48,000 & 234 & 60 & 2,812 & 1 & 18 & 1 &  &  &  &  \\
M4 Weekly & Econ/Fin & W-SUN & 359 & 1,035 & 93 & 2,610 & 1 & 13 & 1 &  &  &  &  \\
M4 Daily & Econ/Fin & D & 4,227 & 2,371 & 107 & 9,933 & 1 & 14 & 1 &  &  &  &  \\
M4 Hourly & Econ/Fin & H & 414 & 902 & 748 & 1,008 & 1 & 48 & 2 &  &  &  &  \\
Hospital & Healthcare & M & 767 & 84 & 84 & 84 & 1 & 12 & 1 &  &  &  &  \\
COVID Deaths & Healthcare & D & 266 & 212 & 212 & 212 & 1 & 30 & 1 &  &  &  &  \\
US Births & Healthcare & D & 1 & 7,305 & 7,305 & 7,305 & 1 & 30 & 20 &  &  &  &  \\
US Births & Healthcare & D & 1 & 7,305 & 7,305 & 7,305 & 1 & 30 & 20 &  &  &  &  \\
US Births & Healthcare & W-TUE & 1 & 1,043 & 1,043 & 1,043 & 1 & 8 & 14 &  &  &  &  \\
US Births & Healthcare & M & 1 & 240 & 240 & 240 & 1 & 12 & 2 &  &  &  &  \\
Saugeen & Nature & D & 1 & 23,741 & 23,741 & 23,741 & 1 & 30 & 20 &  &  &  &  \\
Saugeen & Nature & W-THU & 1 & 3,391 & 3,391 & 3,391 & 1 & 8 & 20 &  &  &  &  \\
Saugeen & Nature & M & 1 & 780 & 780 & 780 & 1 & 12 & 7 &  &  &  &  \\
Temperature Rain & Nature & D & 32,072 & 725 & 725 & 725 & 1 & 30 & 3 &  &  &  &  \\
KDD Cup 2018 & Nature & H & 270 & 10,898 & 9,504 & 10,920 & 1 & 48 & 20 & 480 & 2 & 720 & 2 \\
KDD Cup 2018 & Nature & D & 270 & 455 & 396 & 455 & 1 & 30 & 2 &  &  &  &  \\
Car Parts & Sales & M & 2,674 & 51 & 51 & 51 & 1 & 12 & 1 &  &  &  &  \\
Electricity & Energy & 15T & 370 & 140,256 & 140,256 & 140,256 & 1 & 48 & 20 & 480 & 20 & 720 & 20 \\
Electricity & Energy & H & 370 & 35,064 & 35,064 & 35,064 & 1 & 48 & 20 & 480 & 8 & 720 & 5 \\
Electricity & Energy & D & 370 & 1,461 & 1,461 & 1,461 & 1 & 30 & 5 &  &  &  &  \\
Electricity & Energy & W-FRI & 370 & 208 & 208 & 208 & 1 & 8 & 3 &  &  &  &  \\
\bottomrule
\end{tabular}%
}
\end{table*}

\begin{table*}[t]
\centering
\scriptsize
\caption{
Benchmark datasets dataset summary of fev-bench \cite{shchur2025fev}.
This benchmark contains 100 tasks including the ones listed here and overlapping tasks from GIFT-Eval on the 
BizITObs - L2C, 
ETT, 
Hierarchical Sales, 
Hospital, 
Jena Weather, 
Loop Seattle, 
M-DENSE, 
SZ Taxi, 
Solar
tasks.
}
\label{tab:dataset_eval_fev}
\setlength{\tabcolsep}{3pt}
\begin{tabular}{@{\extracolsep{1pt}}l l c r r r r r r r r@{}}
\toprule
\textbf{Task} & \textbf{Domain} & \textbf{Freq.} & $\boldsymbol{H}$ & $\boldsymbol{W}$ &
\textbf{\shortstack{Median \\ length}} & \textbf{\shortstack{Num \\ series}} & \textbf{\shortstack{Num \\ targets}} &
\textbf{\shortstack{Num \\ past cov.}} & \textbf{\shortstack{Num \\ known cov.}} & \textbf{\shortstack{Num \\ static cov.}} \\
\midrule
Australian Tourism & econ & Q & 8 & 2 & 36 & 89 & 1 & 0 & 0 & 0 \\
FRED-MD - CEE & econ & M & 12 & 20 & 798 & 1 & 3 & 4 & 0 & 0 \\
FRED-MD - Macro & econ & M & 12 & 20 & 798 & 1 & 51 & 0 & 0 & 0 \\
FRED-QD - CEE & econ & Q & 8 & 20 & 266 & 1 & 3 & 4 & 0 & 0 \\
FRED-QD - Macro & econ & Q & 8 & 20 & 266 & 1 & 51 & 0 & 0 & 0 \\
GVAR & econ & Q & 8 & 10 & 178 & 33 & 6 & 3 & 0 & 0 \\
US Consumption & econ & M & 12 & 10 & 792 & 31 & 1 & 0 & 0 & 0 \\
US Consumption & econ & Q & 8 & 10 & 262 & 31 & 1 & 0 & 0 & 0 \\
US Consumption & econ & Y & 5 & 10 & 64 & 31 & 1 & 0 & 0 & 0 \\
World CO2 Emissions & econ & Y & 5 & 9 & 60 & 191 & 1 & 0 & 0 & 0 \\
World Life Expectancy & econ & Y & 5 & 10 & 74 & 237 & 1 & 0 & 0 & 0 \\
World Tourism & econ & Y & 5 & 2 & 21 & 178 & 1 & 0 & 0 & 0 \\
ENTSO-e Load & energy & 15T & 96 & 20 & 175292 & 6 & 1 & 0 & 3 & 0 \\
ENTSO-e Load & energy & 30T & 96 & 20 & 87645 & 6 & 1 & 0 & 3 & 0 \\
ENTSO-e Load & energy & H & 168 & 20 & 43822 & 6 & 1 & 0 & 3 & 0 \\
EPF-BE & energy & H & 24 & 20 & 52416 & 1 & 1 & 0 & 2 & 0 \\
EPF-DE & energy & H & 24 & 20 & 52416 & 1 & 1 & 0 & 2 & 0 \\
EPF-FR & energy & H & 24 & 20 & 52416 & 1 & 1 & 0 & 2 & 0 \\
EPF-NP & energy & H & 24 & 20 & 52416 & 1 & 1 & 0 & 2 & 0 \\
EPF-PJM & energy & H & 24 & 20 & 52416 & 1 & 1 & 0 & 2 & 0 \\
ERCOT & energy & D & 28 & 20 & 6452 & 8 & 1 & 0 & 0 & 0 \\
ERCOT & energy & H & 168 & 20 & 154872 & 8 & 1 & 0 & 0 & 0 \\
ERCOT & energy & M & 12 & 15 & 211 & 8 & 1 & 0 & 0 & 0 \\
ERCOT & energy & W & 13 & 20 & 921 & 8 & 1 & 0 & 0 & 0 \\
GFC12 & energy & H & 168 & 10 & 39414 & 11 & 1 & 0 & 1 & 0 \\
GFC14 & energy & H & 168 & 20 & 17520 & 1 & 1 & 0 & 1 & 0 \\
GFC17 & energy & H & 168 & 20 & 17544 & 8 & 1 & 0 & 1 & 0 \\
Solar with Weather & energy & 15T & 96 & 20 & 198600 & 1 & 1 & 2 & 7 & 0 \\
Solar with Weather & energy & H & 24 & 20 & 49648 & 1 & 1 & 2 & 7 & 0 \\
BOOMLET-1062 & cloud & 5T & 288 & 20 & 16384 & 1 & 21 & 0 & 0 & 0 \\
BOOMLET-1209 & cloud & 5T & 288 & 20 & 16384 & 1 & 53 & 0 & 0 & 0 \\
BOOMLET-1225 & cloud & T & 60 & 20 & 16384 & 1 & 49 & 0 & 0 & 0 \\
BOOMLET-1230 & cloud & 5T & 288 & 20 & 16384 & 1 & 23 & 0 & 0 & 0 \\
BOOMLET-1282 & cloud & T & 60 & 20 & 16384 & 1 & 35 & 0 & 0 & 0 \\
BOOMLET-1487 & cloud & 5T & 288 & 20 & 16384 & 1 & 54 & 0 & 0 & 0 \\
BOOMLET-1631 & cloud & 30T & 96 & 20 & 10463 & 1 & 40 & 0 & 0 & 0 \\
BOOMLET-1676 & cloud & 30T & 96 & 20 & 10463 & 1 & 100 & 0 & 0 & 0 \\
BOOMLET-1855 & cloud & H & 24 & 20 & 5231 & 1 & 52 & 0 & 0 & 0 \\
BOOMLET-1975 & cloud & H & 24 & 20 & 5231 & 1 & 75 & 0 & 0 & 0 \\
BOOMLET-2187 & cloud & H & 24 & 20 & 5231 & 1 & 100 & 0 & 0 & 0 \\
Favorita Store Sales & retail & M & 12 & 2 & 54 & 1579 & 1 & 1 & 1 & 6 \\
Favorita Store Sales & retail & W & 13 & 10 & 240 & 1579 & 1 & 1 & 1 & 6 \\
Favorita Store Sales & retail & D & 28 & 10 & 1688 & 1579 & 1 & 1 & 2 & 6 \\
Favorita Transactions & retail & M & 12 & 2 & 54 & 51 & 1 & 1 & 0 & 5 \\
Favorita Transactions & retail & W & 13 & 10 & 240 & 51 & 1 & 1 & 0 & 5 \\
Favorita Transactions & retail & D & 28 & 10 & 1688 & 51 & 1 & 1 & 1 & 5 \\
KDD Cup 2022 & energy & D & 14 & 10 & 243 & 134 & 1 & 9 & 0 & 0 \\
KDD Cup 2022 & energy & 10T & 288 & 10 & 35279 & 134 & 1 & 9 & 0 & 0 \\
KDD Cup 2022 & energy & 30T & 96 & 10 & 11758 & 134 & 1 & 9 & 0 & 0 \\
M5 & retail & M & 12 & 1 & 58 & 30490 & 1 & 0 & 8 & 5 \\
M5 & retail & W & 13 & 1 & 257 & 30490 & 1 & 0 & 8 & 5 \\
M5 & retail & D & 28 & 1 & 1810 & 30490 & 1 & 0 & 8 & 5 \\
Restaurant & retail & D & 28 & 8 & 296 & 817 & 1 & 0 & 0 & 4 \\
Rossmann & retail & W & 13 & 8 & 133 & 1115 & 1 & 1 & 4 & 10 \\
Rossmann & retail & D & 48 & 10 & 942 & 1115 & 1 & 1 & 5 & 10 \\
Walmart & retail & W & 39 & 1 & 143 & 2936 & 1 & 0 & 10 & 4 \\
ECDC ILI & healthcare & W & 13 & 10 & 201 & 25 & 1 & 0 & 0 & 0 \\
Hospital Admissions & healthcare & D & 28 & 20 & 1731 & 8 & 1 & 0 & 0 & 0 \\
Hospital Admissions & healthcare & W & 13 & 16 & 246 & 8 & 1 & 0 & 0 & 0 \\
UK COVID - Nation - Cumulative & healthcare & D & 28 & 20 & 729 & 4 & 3 & 5 & 0 & 0 \\
UK COVID - Nation - New & healthcare & D & 28 & 20 & 729 & 4 & 3 & 5 & 0 & 0 \\
UK COVID - UTLA - Cumulative & healthcare & W & 13 & 5 & 104 & 214 & 1 & 0 & 0 & 0 \\
UK COVID - UTLA - New & healthcare & D & 28 & 10 & 721 & 214 & 1 & 0 & 0 & 0 \\
\bottomrule
\end{tabular}
\end{table*}

\clearpage

\subsection{Classification Data}
\begin{center}
\scriptsize
\begin{longtable}{llrrrrlc}
\caption{%
    Datasets summary of UCR Time Series Classification Archive \cite{dau2019ucr}.
    The benchmark provides various settings for evaluating time series classification tasks.
    Each dataset is evaluated over a predefined train/test split.%
    Test subsets - 1: Univariate Embedding Based,
    2: Multivariate Embedding Based,
    3: Uni-/Multi-variate Task Adaptation Based.
}
\label{tab:ucr_classification_dataset}\\
\toprule
\textbf{Name} & \textbf{Type} & \textbf{Train} & \textbf{Test} & \textbf{Length} & \textbf{Class} & \textbf{Variate} & \textbf{Test Subset} \\
\midrule
\endfirsthead
\multicolumn{8}{l}{\footnotesize\textit{(continued from previous page)}}\\
\toprule
\textbf{Name} & \textbf{Type} & \textbf{Train} & \textbf{Test} & \textbf{Length} & \textbf{Class} & \textbf{Variate} & \textbf{Test Subset} \\
\midrule
\endhead
\midrule
\multicolumn{8}{r}{\footnotesize\textit{(continued on next page)}}\\
\endfoot
\bottomrule
\endlastfoot

ACSF1 & DEVICE & 100 & 100 & 1460 & 10 & univariate & 3 \\
Adiac & IMAGE & 390 & 391 & 176 & 37 & univariate & 1 \\
ArrowHead & IMAGE & 36 & 175 & 251 & 3 & univariate & 3 \\
ArticularyWordRecognition & MOTION & 275 & 300 & 144 & 25 & multivariate & 2 \\
AsphaltObstaclesCoordinates & MOTION & 390 & 391 & 0 & 4 & multivariate &  \\
AsphaltPavementTypeCoordinates & MOTION & 1055 & 1056 & 0 & 3 & multivariate &  \\
AsphaltRegularityCoordinates & MOTION & 751 & 751 & 0 & 2 & multivariate &  \\
AtrialFibrillation & ECG & 15 & 15 & 640 & 3 & multivariate & 3 \\
BasicMotions & HAR & 40 & 40 & 100 & 4 & multivariate & 3 \\
Beef & SPECTRO & 30 & 30 & 470 & 5 & univariate & 3 \\
BeetleFly & IMAGE & 20 & 20 & 512 & 2 & univariate & 3 \\
BirdChicken & IMAGE & 20 & 20 & 512 & 2 & univariate & 3 \\
BME & SIMULATED & 30 & 150 & 128 & 3 & univariate & 3 \\
Car & SENSOR & 60 & 60 & 577 & 4 & univariate & 3 \\
CBF & SIMULATED & 30 & 900 & 128 & 3 & univariate & 3 \\
CharacterTrajectories & MOTION & 1422 & 1436 & 0 & 20 & multivariate &  \\
Chinatown & TRAFFIC & 20 & 345 & 24 & 2 & univariate & 1 \\
ChlorineConcentration & SIMULATED & 467 & 3840 & 166 & 3 & univariate & 3 \\
CinCECGTorso & ECG & 40 & 1380 & 1639 & 4 & univariate & 3 \\
Coffee & SPECTRO & 28 & 28 & 286 & 2 & univariate & 3 \\
Computers & DEVICE & 250 & 250 & 720 & 2 & univariate & 3 \\
Cricket & HAR & 108 & 72 & 1197 & 12 & multivariate & 2 \\
CricketX & HAR & 390 & 390 & 300 & 12 & univariate & 1 \\
CricketY & HAR & 390 & 390 & 300 & 12 & univariate & 1 \\
CricketZ & HAR & 390 & 390 & 300 & 12 & univariate & 1 \\
Crop & IMAGE & 7200 & 16800 & 46 & 24 & univariate & 1 \\
DiatomSizeReduction & IMAGE & 16 & 306 & 345 & 4 & univariate & 3 \\
DistalPhalanxOutlineAgeGroup & IMAGE & 400 & 139 & 80 & 3 & univariate & 3 \\
DistalPhalanxOutlineCorrect & IMAGE & 600 & 276 & 80 & 2 & univariate & 3 \\
DistalPhalanxTW & IMAGE & 400 & 139 & 80 & 6 & univariate & 3 \\
DuckDuckGeese & AUDIO & 60 & 40 & 270 & 5 & multivariate & 3 \\
Earthquakes & SENSOR & 322 & 139 & 512 & 2 & univariate & 3 \\
ECG200 & ECG & 100 & 100 & 96 & 2 & univariate & 3 \\
ECG5000 & ECG & 500 & 4500 & 140 & 5 & univariate & 3 \\
ECGFiveDays & ECG & 23 & 861 & 136 & 2 & univariate & 3 \\
EigenWorms & MOTION & 131 & 128 & 17984 & 5 & multivariate &  \\
ElectricDevices & DEVICE & 8926 & 7711 & 96 & 7 & univariate & 1 \\
EOGHorizontalSignal & EOG & 362 & 362 & 1250 & 12 & univariate & 1 \\
EOGVerticalSignal & EOG & 362 & 362 & 1250 & 12 & univariate & 1 \\
Epilepsy & HAR & 137 & 138 & 207 & 4 & multivariate & 3 \\
ERing & HAR & 30 & 270 & 65 & 6 & multivariate & 3 \\
EthanolConcentration & SPECTRO & 261 & 263 & 1751 & 4 & multivariate & 3 \\
EthanolLevel & SPECTRO & 504 & 500 & 1751 & 4 & univariate & 3 \\
FaceAll & IMAGE & 560 & 1690 & 131 & 14 & univariate &  \\
FaceDetection & EEG & 5890 & 3524 & 62 & 2 & multivariate & 3 \\
FaceFour & IMAGE & 24 & 88 & 350 & 4 & univariate & 3 \\
FacesUCR & IMAGE & 200 & 2050 & 131 & 14 & univariate &  \\
FiftyWords & IMAGE & 450 & 455 & 270 & 50 & univariate &  \\
FingerMovements & EEG & 316 & 100 & 50 & 2 & multivariate & 3 \\
Fish & IMAGE & 175 & 175 & 463 & 7 & univariate & 3 \\
FordA & SENSOR & 3601 & 1320 & 500 & 2 & univariate & 3 \\
FordB & SENSOR & 3636 & 810 & 500 & 2 & univariate & 3 \\
FreezerRegularTrain & DEVICE & 150 & 2850 & 301 & 2 & univariate & 3 \\
FreezerSmallTrain & DEVICE & 28 & 2850 & 301 & 2 & univariate & 3 \\
GunPoint & HAR & 50 & 150 & 150 & 2 & univariate & 3 \\
GunPointAgeSpan & HAR & 135 & 316 & 150 & 2 & univariate & 3 \\
GunPointMaleVersusFemale & HAR & 135 & 316 & 150 & 2 & univariate & 3 \\
GunPointOldVersusYoung & HAR & 135 & 316 & 150 & 2 & univariate & 3 \\
Ham & SPECTRO & 109 & 105 & 431 & 2 & univariate & 3 \\
HandMovementDirection & EEG & 160 & 74 & 400 & 4 & multivariate & 3 \\
HandOutlines & IMAGE & 1000 & 370 & 2709 & 2 & univariate & 3 \\
Handwriting & HAR & 150 & 850 & 152 & 26 & multivariate & 2 \\
Haptics & MOTION & 155 & 308 & 1092 & 5 & univariate & 3 \\
Heartbeat & AUDIO & 204 & 205 & 405 & 2 & multivariate & 3 \\
Herring & IMAGE & 64 & 64 & 512 & 2 & univariate & 3 \\
HouseTwenty & DEVICE & 34 & 101 & 3000 & 2 & univariate & 3 \\
InlineSkate & MOTION & 100 & 550 & 1882 & 7 & univariate & 3 \\
InsectEPGRegularTrain & EPG & 62 & 249 & 601 & 3 & univariate & 3 \\
InsectEPGSmallTrain & EPG & 17 & 249 & 601 & 3 & univariate & 3 \\
InsectWingbeat & AUDIO & 25000 & 25000 & 0 & 10 & multivariate &  \\
ItalyPowerDemand & SENSOR & 67 & 1029 & 24 & 2 & univariate &  \\
JapaneseVowels & AUDIO & 270 & 370 & 29 & 9 & multivariate &  \\
LargeKitchenAppliances & DEVICE & 375 & 375 & 720 & 3 & univariate & 3 \\
Libras & HAR & 180 & 180 & 45 & 15 & multivariate & 2 \\
Lightning2 & SENSOR & 60 & 61 & 637 & 2 & univariate & 3 \\
Lightning7 & SENSOR & 70 & 73 & 319 & 7 & univariate & 3 \\
LSST & OTHER & 2459 & 2466 & 36 & 14 & multivariate & 2 \\
Mallat & SIMULATED & 55 & 2345 & 1024 & 8 & univariate & 3 \\
Meat & SPECTRO & 60 & 60 & 448 & 3 & univariate & 3 \\
MedicalImages & IMAGE & 381 & 760 & 99 & 10 & univariate & 3 \\
MiddlePhalanxOutlineAgeGroup & IMAGE & 400 & 154 & 80 & 3 & univariate & 3 \\
MiddlePhalanxOutlineCorrect & IMAGE & 600 & 291 & 80 & 2 & univariate & 3 \\
MiddlePhalanxTW & IMAGE & 399 & 154 & 80 & 6 & univariate & 3 \\
MixedShapesRegularTrain & IMAGE & 500 & 2425 & 1024 & 5 & univariate & 3 \\
MixedShapesSmallTrain & IMAGE & 100 & 2425 & 1024 & 5 & univariate & 3 \\
MoteStrain & SENSOR & 20 & 1252 & 84 & 2 & univariate & 3 \\
MotorImagery & EEG & 278 & 100 & 3000 & 2 & multivariate & 3 \\
NATOPS & HAR & 180 & 180 & 51 & 6 & multivariate & 3 \\
OliveOil & SPECTRO & 30 & 30 & 570 & 4 & univariate & 3 \\
OSULeaf & IMAGE & 200 & 242 & 427 & 6 & univariate & 3 \\
PEMS-SF & OTHER & 267 & 173 & 144 & 7 & multivariate & 3 \\
PenDigits & MOTION & 7494 & 3498 & 8 & 10 & multivariate &  \\
PhalangesOutlinesCorrect & IMAGE & 1800 & 858 & 80 & 2 & univariate & 3 \\
Plane & SENSOR & 105 & 105 & 144 & 7 & univariate & 3 \\
PowerCons & DEVICE & 180 & 180 & 144 & 2 & univariate & 3 \\
ProximalPhalanxOutlineAgeGroup & IMAGE & 400 & 205 & 80 & 3 & univariate & 3 \\
ProximalPhalanxOutlineCorrect & IMAGE & 600 & 291 & 80 & 2 & univariate & 3 \\
ProximalPhalanxTW & IMAGE & 400 & 205 & 80 & 6 & univariate & 3 \\
RacketSports & HAR & 151 & 152 & 30 & 4 & multivariate & 2 \\
RefrigerationDevices & DEVICE & 375 & 375 & 720 & 3 & univariate & 3 \\
Rock & SPECTRO & 20 & 50 & 2844 & 4 & univariate & 3 \\
ScreenType & DEVICE & 375 & 375 & 720 & 3 & univariate & 3 \\
SelfRegulationSCP1 & EEG & 268 & 293 & 896 & 2 & multivariate & 3 \\
SelfRegulationSCP2 & EEG & 200 & 180 & 1152 & 2 & multivariate & 3 \\
SemgHandGenderCh2 & SPECTRO & 300 & 600 & 1500 & 2 & univariate & 3 \\
SemgHandMovementCh2 & SPECTRO & 450 & 450 & 1500 & 6 & univariate & 3 \\
SemgHandSubjectCh2 & SPECTRO & 450 & 450 & 1500 & 5 & univariate & 3 \\
ShapeletSim & SIMULATED & 20 & 180 & 500 & 2 & univariate & 3 \\
ShapesAll & IMAGE & 600 & 600 & 512 & 60 & univariate &  \\
SmallKitchenAppliances & DEVICE & 375 & 375 & 720 & 3 & univariate & 3 \\
SmoothSubspace & SIMULATED & 150 & 150 & 15 & 3 & univariate &  \\
SonyAIBORobotSurface1 & SENSOR & 20 & 601 & 70 & 2 & univariate & 3 \\
SonyAIBORobotSurface2 & SENSOR & 27 & 953 & 65 & 2 & univariate & 3 \\
SpokenArabicDigits & SPEECH & 6599 & 2199 & 93 & 10 & multivariate &  \\
StandWalkJump & ECG & 12 & 15 & 2500 & 3 & multivariate & 3 \\
StarLightCurves & SENSOR & 1000 & 8236 & 1024 & 3 & univariate & 3 \\
Strawberry & SPECTRO & 613 & 370 & 235 & 2 & univariate & 3 \\
Symbols & IMAGE & 25 & 995 & 398 & 6 & univariate & 3 \\
SyntheticControl & SIMULATED & 300 & 300 & 60 & 6 & univariate & 3 \\
ToeSegmentation1 & MOTION & 40 & 228 & 277 & 2 & univariate & 3 \\
ToeSegmentation2 & MOTION & 36 & 130 & 343 & 2 & univariate & 3 \\
Trace & SENSOR & 100 & 100 & 275 & 4 & univariate & 3 \\
TwoLeadECG & ECG & 23 & 1139 & 82 & 2 & univariate & 3 \\
TwoPatterns & SIMULATED & 1000 & 4000 & 128 & 4 & univariate & 3 \\
UMD & SIMULATED & 36 & 144 & 150 & 3 & univariate & 3 \\
UWaveGestureLibrary & HAR & 2238 & 2241 & 315 & 8 & multivariate & 3 \\
UWaveGestureLibraryAll & HAR & 896 & 3582 & 945 & 8 & univariate & 3 \\
UWaveGestureLibraryX & HAR & 896 & 3582 & 315 & 8 & univariate & 3 \\
UWaveGestureLibraryY & HAR & 896 & 3582 & 315 & 8 & univariate & 3 \\
UWaveGestureLibraryZ & HAR & 896 & 3582 & 315 & 8 & univariate & 3 \\
Wafer & SENSOR & 1000 & 6164 & 152 & 2 & univariate & 3 \\
Wine & SPECTRO & 57 & 54 & 234 & 2 & univariate & 3 \\
Worms & MOTION & 181 & 77 & 900 & 5 & univariate & 3 \\
WormsTwoClass & MOTION & 181 & 77 & 900 & 2 & univariate & 3 \\
Yoga & IMAGE & 300 & 3000 & 426 & 2 & univariate & 3 \\

\end{longtable}
\end{center}

\clearpage

\subsection{Anomaly Detection Data}
\begin{table}[h]
\caption{Datasets used for anomaly-detection evaluation following UniTS \citep{gao2024units}.}
\label{tab:anomaly_datasets}
\centering
\small
\setlength{\tabcolsep}{8pt}
\begin{tabular}{lccc}
\toprule
Dataset & Sequence Length & Variables & Domain \\
\midrule
SMD \citep{su2019robust} & 96 & 38 & Machine \\
MSL \citep{hundman2018detecting} & 96 & 55 & Spacecraft \\
SMAP \citep{hundman2018detecting} & 96 & 25 & Spacecraft \\
SWaT \citep{mathur2016swat} & 96 & 51 & Infrastructure \\
PSM \citep{abdulaal2021practical} & 96 & 25 & Machine \\
\bottomrule
\end{tabular}
\end{table}

\subsection{Imputation Data}
\begin{table}[h]
\caption{Datasets for imputation tasks.}
\label{tab:imputation_datasets}
\centering
\small
\setlength{\tabcolsep}{8pt}

\resizebox{\textwidth}{!}{%
\begin{tabular}{lccclc}
\toprule
Name
& Sequence Length
& Variables
& Task
& Mask Ratio
& Class \\
\midrule

ETTm1 \citep{zhou2021informer}
& 96
& 7
& Imputation
& 12.5\%, 25\%, 37.5\%, 50\%
& Electricity \\

ETTh1 \citep{zhou2021informer}
& 96
& 7
& Imputation
& 12.5\%, 25\%, 37.5\%, 50\%
& Electricity \\

ECL \citep{trindade2015electricityloaddiagrams20112014}
& 96
& 321
& Imputation
& 12.5\%, 25\%, 37.5\%, 50\%
& Electricity \\

Weather \citep{wetstationweather}
& 96
& 21
& Imputation
& 12.5\%, 25\%, 37.5\%, 50\%
& Weather \\

\bottomrule
\end{tabular}%
}
\end{table}



\end{document}